%% file: main.tex
\documentclass{article}

\usepackage{PRIMEarxiv}

\usepackage[utf8]{inputenc} 
\usepackage[T1]{fontenc}    
\usepackage[backref=page]{hyperref}
\usepackage{url}            
\usepackage{booktabs}       
\usepackage{tabularx}       
\newcolumntype{Y}{>{\centering\arraybackslash}X}
\newcolumntype{Z}{>{\raggedleft\arraybackslash}X}
\usepackage{amsfonts}       
\usepackage{nicefrac}       
\usepackage{microtype}      
\usepackage{pifont}
\newcommand{\cmark}{\ding{51}}%
\usepackage{lipsum}
\usepackage{fancyhdr}       
\usepackage{graphicx}       
\usepackage{amsmath}
\usepackage{caption}
\graphicspath{{media/}}     
\usepackage[normalem]{ulem}
\usepackage{lscape} 
\usepackage{adjustbox}
\usepackage{float}
\usepackage{subcaption}
\usepackage{multirow}
\usepackage{enumitem}
\usepackage[capitalize]{cleveref}
\usepackage[dvipsnames]{xcolor}
\usepackage{xspace}
\newcommand{\eg}{e.g.\xspace}
\newcommand{\ie}{i.e.\xspace}

\title{PANGAEA: A Global and Inclusive Benchmark for Geospatial Foundation Models
}

\author{
  \textbf{Valerio Marsocci}$^{*,1,2}$, \textbf{Yuru Jia}$^{*,2,3}$, \textbf{Georges Le Bellier}$^{4}$, \textbf{David Kerekes}$^{3}$, \\
  \textbf{Liang Zeng}$^{2}$, \textbf{Sebastian Hafner}$^{3,5}$, \textbf{Sebastian Gerard}$^{3}$, \\
  \textbf{Eric Brune}$^{3}$,  \textbf{Ritu Yadav}$^{3}$, \textbf{Ali Shibli}$^{3}$, \textbf{Heng Fang}$^{3}$, \\
  \textbf{Yifang Ban}$^{3}$, \textbf{Maarten Vergauwen}$^{2}$, \textbf{Nicolas Audebert}$^{6,4}$, \textbf{Andrea Nascetti}$^{3}$ \\
   \small$^1$ ESA $\Phi$-lab, Frascati, Italy\\ \small$^2$Geomatics Section, Department of Civil Engineering, Faculty of Engineering Technology, KU Leuven, Ghent, Belgium\\ \small$^3$KTH - Royal Institute of Technology Stockholm, Sweden \\ \small$^4$ Conservatoire national des arts et métiers, CEDRIC, F-75141 Paris, France, \\ \small$^5$University of Glasgow, Glasgow, Scotland \\ \small$^6$ Univ. Gustave Eiffel, ENSG, IGN, LASTIG, F-94160 Saint-Mandé, France\\
  \small$^*$ Equal contribution
}

\begin{document}
\maketitle

\begin{abstract}
Geospatial Foundation Models (GFMs) have emerged as powerful tools for extracting representations from Earth observation data, but their evaluation remains inconsistent and narrow. Existing works often evaluate on suboptimal downstream datasets and tasks, that are often too easy or too narrow, limiting the usefulness of the evaluations to assess the real-world applicability of GFMs. Additionally, there is a distinct lack of diversity in current evaluation protocols, which fail to account for the multiplicity of image resolutions, sensor types, and temporalities, which further complicates the assessment of GFM performance.
In particular, most existing benchmarks are geographically biased towards North America and Europe, questioning the global applicability of GFMs. 
To overcome these challenges, we introduce PANGAEA, a standardized evaluation protocol that covers a diverse set of datasets, tasks, resolutions, sensor modalities, and temporalities.
It establishes a robust and widely applicable benchmark for GFMs. We evaluate the most popular GFMs openly available on this benchmark and analyze their performance across several domains.
In particular, we compare these models to supervised baselines (\eg UNet and vanilla ViT), and assess their effectiveness when faced with limited labeled data. Our findings highlight the limitations of GFMs, under different scenarios, showing that they do not consistently outperform supervised models.
PANGAEA is designed to be highly extensible, allowing for the seamless inclusion of new datasets, models, and tasks in future research. By releasing the evaluation code and benchmark, we aim to enable other researchers to replicate our experiments and build upon our work, fostering a more principled evaluation protocol for large pre-trained geospatial models. The code is available at \url{https://github.com/VMarsocci/pangaea-bench}.
\end{abstract}


\input{figs_tex/teaser}

\section{Introduction}
\label{sec:intro}

Geospatial foundation models (GFMs) have emerged as a powerful tool for extracting representations from Earth observation (EO) data \cite{visionpaper, lacoste2021foundation}. These models are trained on large-scale EO datasets and aim to capture fundamental geospatial features that can be leveraged for downstream tasks \cite{tuia2023artificial, zhang2024geoscience}. However, the rapid development of GFMs has outpaced the establishment of a standardized and robust evaluation methodology.

Current evaluation practices often rely on a limited set of downstream tasks and datasets. For example, EuroSAT \cite{helber2019eurosat} is often used to evaluate downstream performance on land cover classification, despite being virtually solved (MTP \cite{wang2024mtp} reports $>99\%$ accuracy) and despite the community's shift toward segmentation-based tasks for land cover and land use prediction, given the real-world importance.
Reported performance on such datasets is, therefore, not representative of the performance of GFMs across a broader range of applications. Additionally, current evaluation protocols often lack diversity, overlooking differences in image resolution, multiplicity of sensor types and modalities, geographical diversity and other factors crucial to model performance \cite{mai2023opportunities, rolf2024mission}. For instance, the PhilEO Bench \cite{fibaek2024phileo}, though a valuable contribution, evaluates models only through the lens of Sentinel-2 data, limiting the breadth of the conclusions that can be drawn from this benchmark. In comparison, GEO-Bench \cite{lacoste2023geobench} and SustainBench \cite{yeh2021sustainbench}, include data from multiple sensors; however, they cover only land observations, ignore marine environments, and focus on mono-temporal applications. Finally, FoMo-Bench \cite{bountos2023fomo} and Crop Type Bench \cite{Chang2024OnTG} restrict their scope respectively to forest and agriculture environments.  

This lack of a standardized comprehensive evaluation protocol poses a significant challenge for the remote sensing community. It limits the ability to effectively compare models, assess their operational capabilities, and identify areas for improvement. Consequently, researchers and practitioners struggle to make informed decisions about which GFMs to utilize for their specific needs.
To address this critical gap, we propose PANGAEA, a global and inclusive benchmark for GFMs. We aim to establish a robust, standardized and broadly applicable benchmark by curating a diverse and comprehensive collection of downstream datasets and tasks. PANGAEA covers a wide spectrum of tasks, ranging from marine debris segmentation to building damage mapping through change detection. It spans multiple application domains, including urban, agricultural, marine, and forest environments. It also incorporates datasets with diverse temporal configurations (uni-temporal, bi-temporal, and multi-temporal, i.e. satellite image time series), extensive geographical coverage (all over the Earth), various spatial resolutions (from 1.5 m/pixel to 30 m/pixel resolution), and multiple sensor types, including natural colors, multi-spectral and synthetic aperture radar (SAR).

To gain a deeper understanding of the GFM landscape, we select a set of highly representative models and put them to the test on PANGAEA. Specifically, our selection criteria focus on approach, impact, and reproducibility. 
Regarding the different approaches, we include models employing diverse strategies, such as multi-modal contrastive learning (e.g.~\cite{fuller2023croma}), masked image modeling (MIM) (e.g.~\cite{reed2023scalemae}), and supervised training (e.g.~\cite{bastani2022satlas}). 
With impactful contributions, we consider models featured in top-tier computer vision conferences, specialized journals, or recognized within the research community.
For reproducibility, the availability of open-source code and pre-trained models is a pivotal consideration.
This comparative analysis allows us to assess the strengths and weaknesses of different models, providing insights and fostering advancements in the development of GFMs.


Through the development of a standardized evaluation protocol and comprehensive benchmarking effort, we examine a set of key research questions: 
\begin{itemize}[nosep]
    \item \textbf{Generalization across domains:} How effectively do GFMs generalize to diverse downstream task domains? How do different model designs contribute to performance variations across these domains?
    \item \textbf{Comparison to supervised baselines:} Do GFMs consistently beat supervised task-specific baselines?
    \item \textbf{Performance with limited labels:} Do GFMs demonstrate tangible advantages in scenarios with limited labeled data?
\end{itemize}

Our experiments yield new insights into the applicability of GFMs. Pre-training dataset characteristics, with a specific focus on the sensor spectral and spatial resolutions, have a significant impact on the downstream performance of different GFMs. GFMs pre-trained on datasets with richer spectral information or higher spatial resolutions tend to perform better in tasks requiring such features \cref{fig:teaser}. However, while GFMs generally perform well across a wide range of tasks, they do not consistently outperform supervised baselines. Task-specific models, \ie those trained end-to-end on the task at hand, can leverage the spectral and spatial details of the dataset more effectively. This is especially true when there is a notable gap between the characteristics of the pre-training dataset and the downstream dataset (e.g. RGB vs multi-spectral). Similar behavior has been observed in recent works, spanning from time series \cite{xu2024specializedfoundationmodelsstruggle}, to crop monitoring \cite{Chang2024OnTG} and canopy height prediction \cite{rolf2024contrastinglocalglobalmodeling}. 
Similar findings are also in the data scarcity scenario, where some GFMs excel also w.r.t. supervised baselines (\cref{tab:com-10}), but the average performance confirms what is observed when much more labels are available (\cref{fig:teaser}). Finally, we observe no strong downstream performance advantage between a frozen encoder and end-to-end fine-tuning. While fine-tuning does improve performance in some cases, this is not systematic. The impact of fine-tuning depends on the GFM’s architecture and the type of downstream task. Many other factors were tested in several ablations studies, ranging from the impact of normalization to hyperparameter fine-tuning.

In addition to evaluating model performance, PANGAEA emphasizes the importance of results reproducibility. By releasing the full evaluation code alongside our proposed benchmark, we ensure that other researchers can replicate our experiments and compare models transparently thanks to an efficient and robust evaluation protocol. 
This reproducibility fosters greater transparency and collaboration within the research community, promoting trust and consistency in GFM evaluations.
Our benchmark is designed to be extensible and modular, enabling the easy inclusion of additional datasets, tasks, and models in future research. This flexibility ensures that the benchmark remains adaptable to the evolving needs of the remote sensing community, accommodating new sensor types, more complex tasks, and emerging model architectures.

\input{figs_tex/pangaea_fig}
In summary, our contributions are as follows:
\begin{itemize}[nosep]
    \item \textbf{Open-source codebase} for reproducible and transparent benchmarking. We provide an \textbf{easy-to-use and robust evaluation protocol}, to ensure comparability among different GFMs pre-trained under varying conditions.
    \item \textbf{Diverse benchmark} covering a wide range of datasets, tasks, and sensor types, fostering a more comprehensive evaluation of GFMs.
    \item Highlighting the \textbf{gaps in current GFM capabilities} and showcasing the potential for future advancements in model generalization, task specialization, and fine-tuning strategies.
\end{itemize}

By establishing this framework illustrated in \cref{fig:pangaea}, we aim to facilitate a comprehensive and robust benchmark evaluation, making it easier to understand and analyze the performance and generalization capabilities of newly developed GFMs.

\section{Related Work}
\label{sec:rel}
\subsection{Datasets for Earth Observation}

In recent years, the availability of geospatial and EO datasets has expanded significantly, motivating systematic efforts to catalog and organize them \cite{xiong2024earthnets, Schmitt_2023}. These datasets can be classified according to various criteria, including task, modality, geographic region, domain, and temporality.
From a task-based perspective, research has branched into distinct areas, from scene classification to pixel-level regression. 

\textbf{Scene classification}, often focusing on land use or land cover (LULC), is one of the most prevalent tasks due to its alignment with computer vision methodologies. Early examples include UC-Merced \cite{yang2010bag} and WHU-RS19 \cite{Xia2010WHURS19, Dai2011WHURS19}, which consist of high-resolution RGB imagery. MillionAID \cite{Long2022ASP, Long2021DiRS} is among the largest datasets in this category. Beyond RGB images, some datasets like EuroSAT \cite{helber2019eurosat} rely on Sentinel-2 images, while BigEarthNet \cite{sumbul2019bigearthnet, sumbul2021bigearthnet} incorporates both Sentinel-1 and Sentinel-2 data while introducing a multi-label classification task.  
Beyond LULC scene classification, many specialized datasets address specific tasks with domain-specific annotations. For example, METER-ML \cite{zhu2022meterml} is designed for methane source mapping, CV4A Kenya \cite{kluger2021two} for crop type identification, BrazilDAM \cite{ferreira2020brazildam} for dam detection, and ForestNet \cite{irvin2020forestnet} for deforestation driver classification.
We do not include classification tasks in PANGAEA, despite these tasks being widespread in academic datasets, because patch-level classification is often unsuitable for real-world geospatial tasks. Indeed, patch-level predictions lose critical spatial detail and fail to provide the dense information required for real-world applications. This has been the main driver of the community shift towards dense classification.

In addition to classification, \textbf{object detection} in EO imagery has gained considerable attention. This task involves locating and identifying specific objects, such as airplanes or ships, with applications closely tied to the annotated object categories. Notable datasets in this domain include Functional Map of the World (fMoW) \cite{christie2018functional}, a high-resolution global dataset with 62 object categories; DIOR \cite{li2020object}, a large-scale 20-classes multi-resolution object detection benchmark; and DOTA \cite{ding2102object} and iSAID \cite{waqas2019isaid}, which are curated to include many small-size objects and unbalanced classes. Furthermore, specialized datasets cater to specific object categories, e.g. to detect ships \cite{paolo2022xview3sardetectingdarkfishing, huang2017opensarship, li2017ship}, oil storage tanks \cite{rabbi2020small} and vehicles \cite{razakarivony2016vehicle, heitz2008learning}.
To date, object detection in EO doesn't have a universally established protocol for annotation and for robustly using the GFM features; additionally, semantic segmentation often achieves similar results by providing pixel-level information that can be aggregated to detect and delineate objects more flexibly. For this reason, we didn't include object detection in PANGAEA, following the same approach as in other benchmarks \cite{lacoste2023geobench}.

\textbf{Semantic segmentation} is another common task in EO image analysis. Unlike classification task which focuses on image-level labels, segmentation aims at assigning labels at the pixel level. This dense labeling is essential to produce higher-resolution semantic maps.
Several notable semantic segmentation datasets exist for LULC classes. ISPRS 2D Vaihingen and Potsdam \cite{rottensteiner2012isprs} provide an early example of multi-modal data for semantic segmentation. However, the community has moved towards datasets that capture multiple areas of interest, such as LoveDA \cite{wang2022loveda} and Five-Billion Pixels \cite{tong2023enabling}, which consider aerial and Gaofen-2 satellite imagery respectively, covering several Chinese cities. DeepGlobe \cite{demir2018deepglobe} is notable as one of the first global-scale semantic segmentation for LULC. FLAIR-1 \cite{garioud2023flair} integrates temporal and spectral information from optical satellite time series to capture the diverse landscapes of France.
Beyond land cover mapping, semantic segmentation datasets also tackle specialized applications, such as cloud cover segmentation \cite{foga2017cloud, mohajerani2019cloud, li2018deep}, building footprint segmentation \cite{Sirko2023HighResolutionBA,vanetten2019spacenet}, agricultural parcel mapping \cite{m2019semantic,garnot2021mmfusion}, and beyond.

Another important EO analysis task is \textbf{change detection}, which extends single image classification to multiple acquisitions to identify and map alterations in the Earth’s surface over time. This technique is crucial in applications such as monitoring floods \cite{luppino2019unsupervised}, tracking urban development \cite{toker2022dynamicearthnet,Chen2020}, and assessing the impacts of natural disasters \cite{gupta2019xbd, Gupta_2019_CVPR_Workshops}. Change detection is most often addressed as a classification or segmentation problem, taking 2 or more images as inputs. Among change detection datasets, SZTAKI AirChange \cite{benedek2009change} is an early example focused on identifying significant changes in aerial image pairs taken years apart and across seasons. SECOND \cite{yang2021asymmetric} consists of multi-temporal aerial images with pixel-level labels for land-cover variations and semantic change types. OSCD \cite{daudt2018fully} includes multi-spectral Sentinel-2 images of various global cities for urban changes, while Levir-CD \cite{Chen2020} is a very high-resolution (VHR) dataset designed for building change detection.

Beyond the aforementioned tasks, \textbf{regression} tasks are also becoming increasingly prominent in EO analysis, especially in estimating biophysical indicators from remote sensing imagery. Patch-level regression datasets like SatBird \cite{teng2023satbird}, for bird species distribution modeling, and ClimSim \cite{yu2024climsim}, for climate forecasting, showcase the diverse applications in this domain. Similarly, pixel-level regression tasks are explored by datasets such as BioMassters \cite{nascetti2023biomassters} for above-ground forest biomass estimation, and 3DCD \cite{coletta20223dcd, marsocci2023inferring} for elevation change prediction.


Some useful references for a comprehensive list of currently available datasets include \cite{xiong2024earthnets, Schmitt_2023}, along with resources like the Satellite Image Deep Learning Datasets repository\footnote{\url{https://github.com/satellite-image-deep-learning/datasets}}and the IEEE GRSS Earth Observation Dataset Search\footnote{\url{https://eod-grss-ieee.com/dataset-search}}.

\subsection{Geospatial Foundation Models}

\paragraph{About the Naming} 
"Geospatial Foundation Models" refer to large-scale, self-supervised pre-trained deep learning models specifically designed to process and analyze geospatial data \cite{Bommasani2021FoundationModels}. Instead of targeting specific tasks and datasets, GFMs should aim to support a wide array of downstream applications by training on large amounts of data and leveraging their learned representations to generalize across various domains. 

These large pre-trained geospatial models deviate from the standard vision-language foundation models from computer vision and NLP, such as CLIP \cite{clip} and LLaVa \cite{llava}, in several key aspects. First, vision-language models are multi-modal and integrate texts and images across diverse contexts. In comparison, large pre-trained geospatial models mainly rely on satellite imagery as their primary modality and more rarely incorporate other modalities, especially metadata or text. Self-supervised models for geospatial data are mostly limited to a single image modality, and scarcely consider vector data or temporal information. 
Second, vision-based modalities in geospatial are dominated by multi-spectral and SAR imagery. Broader integration of geospatial data remains underdeveloped, with little sensor diversity in the pre-training datasets.
This is in part due to the unique challenges of geospatial data, especially size -- ESA’s Copernicus archives alone contain over 66 petabytes of satellite data as of 2024 -- and non-stationarity, with significant variability across geographic regions, sensors, and time. This combination of incomplete multi-modal integration and non-stationary data distribution further complicates the development of robust, generalizable geospatial models.
Third, most ``geospatial foundation models'' are yet to compete with their classical counterparts, both in parameter count and in pre-training data size.
As shown in \cref{tab:pretraining-datasets,tab:flops}, the largest currently available ``geospatial foundation models'' are CROMA \cite{fuller2023croma} and Scale-MAE \cite{reed2023scalemae}, with $\approx$ 350M parameters, which is smaller than the largest CLIP (ViT-L/14, $\approx$430M parameters). The commonly used LAION-5B dataset \cite{LAION} is comprised of more than 5 billion images with an on-disk size greater than 200TB, compared to the 1TB of Prithvi \cite{jakubik2023foundation} or 30TB of Satlas \cite{bastani2022satlas}.


These limitations are closely tied to issues with generalization. Many geospatial models are tested primarily on narrow tasks and datasets, leaving their ability to handle domain shifts—such as varying geographic regions, sensors, or temporal conditions—largely unexplored.
The term ``geospatial foundation model'' is increasingly used in the remote sensing community, although still inconsistently. Some models still refer to themselves as large self-supervised models, although they are evaluated on downstream tasks for their representation abilities.
In this work, we unify all these models under the same umbrella of ``geospatial foundation models" for simplification.
Nonetheless, it is important to recognize that significant gaps remain to achieve parity with vision-language foundation models, especially regarding multi-modal integration, generalization, and robustness.
These challenges must be addressed for these models to fully realize the potential implied by the ``foundation model '' denomination.

In this section, we review these general-purpose models, including those not explicitly labeled as ``foundation models'' but that have still been developed to address various downstream applications.

\paragraph{Geospatial Foundation Models}
The GFM framework has advanced significantly through both supervised and unsupervised approaches for visual tasks. A prominent example of a supervised model is SatlasNet \cite{bastani2022satlas}, which employs a Swin Transformer backbone trained on a large annotated dataset with seven label types. However, full supervision requires extensive expert annotations, which are time-consuming and costly. Given the vast quantities of EO data and the expense of labels, many GFMs now adopt self-supervised learning (SSL) to learn representative features without labeled supervision. Studies have shown that SSL can enhance downstream performance on diverse tasks \cite{wang2022self, wang2022empirical}.
Among the first attempts, SSL4EO-L \cite{stewart2024ssl4eo} and SSL4EO-S12 \cite{ssl4eo} introduce globally distributed datasets based on Landsat-8 and Sentinel-1/2 imagery, respectively. These datasets have been used to train various off-the-shelf computer vision models, such as masked autoencoders (MAE) \cite{he2022masked}, data2vec \cite{data2vec}, DINO \cite{caron2021emerging}, and MoCo \cite{he2020momentum}. 

Within the SSL framework, several works \cite{xiong2024all, cha2023billion, muhtar2023cmid} focus on the techniques of \textbf{masked image modeling (MIM)}, which first masks a proportion of the image and then learn to reconstruct masked parts. RingMO \cite{sun2022ringmo} falls into this category and was one of the first to mention "foundation models" in the title explicitly. RingMo-Sense \cite{ringmosense} and RingMo-SAM \cite{yan2023ringmo} extend it, respectively making use of spatiotemporal evolution disentangling for spatiotemporal prediction tasks and segment anything model (SAM) \cite{kirillov2023segment} for image segmentation tasks. CtxMIM \cite{zhang2023ctxmim} combines a MIM approach with a Siamese network to give better context to the feature extractor.
GFM \cite{mendieta2023geospatial}\footnote{Which we will refer to as GFM-Swin in this paper to disambiguate from the general term GFM.} proposes a continual pre-training strategy using MIM. msGFM \cite{han2024bridging} extends it, including multi-sensor images through a conditional computation approach (i.e. MoE). 
SatMAE \cite{cong2022satmae} was one of the first frameworks based on masked autoencoders (MAE) \cite{he2022masked} to tackle EO peculiarities, including multi-spectral and spatiotemporal location embeddings. SatMAE++ \cite{noman2024rethinking} enhances it through multi-level feature extraction. Cross-scale MAE \cite{tang2024cross}, also enforces cross-scale consistency through contrastive and generative losses. Scale-MAE \cite{reed2023scalemae} introduces resolution-aware positional encodings to learn features at different scales. USat \cite{irvin2023usat} modifies the patch projection layers by embedding each spectral band separately before feeding them into MAE. RVSA \cite{transf_fm} adapts the original full attention in MAE blocks with rotated varied-size window attention. 
FG-MAE \cite{wang2023feature} learns to reconstruct the image features instead of reconstructing the raw pixels. SatVision-TOA \cite{spradlin2024satvisiontoageospatialfoundationmodel} is an MIM, trained on MODIS data.

Leveraging the MAE approach, some research has been oriented toward directly dealing with \textbf{multi-modality} characteristics of EO data \cite{ssl_fm_igarss, li2024s2mae}. Prithvi \cite{jakubik2023foundation} includes 3D positional embedding and 3D patch embedding modifications to learn from massive Harmonized Landsat-Sentinel (HLS) multi-temporal data. OmniSAT \cite{astruc2024omnisat} proposes a multi-modality fusion technique by including spatially aligned VHR images and optical and SAR time series. SpectralGPT \cite{Hong2024spectralgpt} extends MAEs to hyperspectral data. DOFA \cite{xiong2024neural} leverages wavelength as a unifying parameter across various EO modalities to achieve a multi-modal representation. 
MMEarth \cite{nedungadi2024mmearth} utilizes 12 modalities including pixel-level and image-level modalities as a pretext task for MAE reconstruction to learn representations for Sentinel-2 images. PRESTO \cite{tseng2024lightweight} operates on pixel time series data and learns to reconstruct data when bands or timesteps are incomplete. 

Another line of work investigates \textbf{contrastive learning} \cite{tarasiou2022embedding}, which captures the relationships within the data by contrasting similar and dissimilar data points. GASSL \cite{ayush2021geography} leverages spatially aligned images over time to construct temporal positive pairs, while SeCo \cite{manas2021seasonal} and CACo \cite{Mall_2023_CVPR} model also the seasonality. MATTERS \cite{Akiva_2022_CVPR} learns invariance to illumination and viewing angle to achieve consistency in material and texture representation. SkySense \cite{guo2024skysense} encodes multi-modal temporal remote sensing imagery and utilizes contrastive learning to learn features at various levels of granularities. CROMA \cite{fuller2022transfer} combines contrastive learning and MAE with aligned radar and optical images to learn both unimodal and multi-modal representations. GeRSP \cite{huang2024generic} combines contrastive learning and supervised learning to learn from unlabeled remote sensing imagery and labeled natural images jointly. Swim-diff \cite{tian2024swimdiff} introduces a scene-wide matching approach that combines contrastive learning with diffusion models.

As part of the SSL learning paradigm, various studies have examined \textbf{self-distillation} based approaches, like in the DINO \cite{caron2021emerging}, where two different views are fed into two encoders, known as the student and the teacher, and the student learns to predict the teacher's output. For example, DINO-MM \cite{wang2022dinomm} takes concatenated SAR-optical images as raw inputs and feeds the transformed augmented views to the student-teacher network. X-STARS \cite{marsocci2024crosssensor} adds a cross-sensor alignment objective based on the DINO framework, for both pre-training from scratch and continual pre-training. 
There are also works \cite{wang2024mtp, Reed_2022_WACV, zhang2022consecutive, zbontar2021barlow, zhao2024rsmamba, chen2024rsmamba} contributing to the importance of different pre-training steps. For instance, MTP \cite{wang2024mtp} conducts a multi-task pre-training pipeline that supports both convolutional neural networks and vision transformer foundation models. DeCUR \cite{wang2023decur} decouples representations, distinguishing inter and intra-modal embeddings based on Barlow Twins \cite{zbontar2021barlow}, while RSMamba dense \cite{zhao2024rsmamba} and RSMamba \cite{chen2024rsmamba} are two approaches based on Mamba \cite{gu2023mamba}, using selective state spaces. \cite{li2024visionlanguage} surveys models and trends in \textbf{vision-language modeling}, where RemoteCLIP \cite{mall2023remote} was one of the first vision-language foundation models to learn aligned text embeddings and visual features for remote sensing. Other works explored the use of large language models (LLMs) for spatio-temporal modeling and prediction tasks \cite{li2024urbangpt, guo2024towards}.
In addition, \textbf{generative} foundation models have also advanced within the EO field, examples include DiffusionSAT \cite{khanna2023diffusionsat}, CRS-Diff \cite{tang2024crsdiff} and MetaEarth \cite{yu2024metaearth}, which generate realistic EO images all over the Earth based on the diffusion models. 

\paragraph{Specialized Geospatial Foundation Models' Applications}
Some models, though oriented to one specific domain, share some common characteristics with general-purpose GFMs. FoMo-Net \cite{bountos2023fomo} is shaped to solve any forest monitoring task based on the MAE framework. 
Hydro-FM \footnote{\url{https://github.com/isaaccorley/hydro-foundation-model}} is a foundation model trained to solve water-related tasks. 
CSMAE \cite{hackstein2024exploring} focuses on sensor-agnostic image retrieval for remote sensing. 
Concerning SAR images, several models were considered precursors for strong GFMs, based on different approaches like MAEs \cite{allen2023large}, self-distillation \cite{martínezferrer2023exploring}, and DINO \cite{gallegomejia2023exploring}.
Other works focus on more narrow aspects of GFMs, such as optimizing high-performance computing (HPC) performance \cite{wang2024orbit, tsaris2024pretraining}. Research is also progressing on parameter-efficient fine-tuning (PEFT) methods \cite{dong2024upetu, chen2023time, li2024new, he2024fm}. Finally, various studies employ pre-trained GFMs for specialized applications, including, transfer learning \cite{fuller2022transfer}, parameter forecasting \cite{smith2024earthpt}, radar image target recognition \cite{L2024ARXIV}, disaster management \cite{arnaudo2024fmars},
image classification \cite{lin2023ssmaespatialspectralmaskedautoencoder}, domain generalization \cite{gong2024crossearthgeospatialvisionfoundation}, \cite{arnaudo2024fmars}, and building segmentation \cite{wang2024rsbuilding}.

\subsection{Benchmarks in Geospatial Tasks}

The emergence of GFMs has further driven the development of benchmarks designed to evaluate model performance across various tasks; while several valuable benchmarks have been proposed in recent years, they still exhibit certain limitations. A key issue is that many of these benchmarks focus on highly specific \textbf{domains}, making them valuable for targeting specific performance metrics but limiting their applicability from a more generalist perspective.

For example, PhilEO Bench \cite{fibaek2024phileo} proposed a benchmark dataset comprising Sentinel-2 images, annotated for three downstream semantic segmentation tasks: building density estimation, road segmentation, and land cover classification. 
FoMo-Bench \cite{bountos2024fomobench} targets forest monitoring with datasets covering classification, segmentation, and detection tasks, whereas \cite{Chang2024OnTG} focuses on agriculture, selecting six crop classification datasets sourced from five continents.

Other benchmarks, however, overlook certain critical \textbf{tasks} (e.g. change detection) and fail to account for a pivotal characteristic of EO data: \textbf{multi-temporality}.
For instance, EarthNets \cite{xiong2024earthnets} provides a benchmark after analyzing published datasets, including fMoW \cite{christie2018functional} and BigEarthNet \cite{sumbul2021bigearthnet} for classification; DIOR \cite{li2020object} and fMoW \cite{christie2018functional} for object detection; and GeoNRW \cite{baier2021synthesizing} and SEASONET \cite{kossmann2022seasonet} for semantic segmentation. While these datasets contribute valuable benchmarks for various tasks, they overlook the temporal aspect that is crucial for many geospatial models. 
Also, GEO-Bench \cite{lacoste2023geobench} adapts and curates six classification and six segmentation datasets with diverse modalities and spatial resolutions, yet still does not fully incorporate the temporal dimension of EO data. 
Finally, SustainBench \cite{yeh2021sustainbench}, though one of the most comprehensive benchmarks, focusing on 15 public datasets across 7 Sustainable Development Goals (SDGs) and addressing tasks such as economic development, agriculture, health, education, water and sanitation, climate action, and biodiversity, still lacks other important vision modalities, such as Synthetic Aperture Radar (SAR) in addition to optical imagery. 
These considerations are summarized in \cref{tab:bench-comp}, highlighting differences in application domains, tasks, modalities, and temporalities.
\input{tabs/benchmark_comparison}

Several benchmarks also evaluate vision-language models (VLMs) for EO applications. For example, VRSBench \cite{li2024vrsbench} datasets set a benchmark for image captioning, visual grounding, and visual question answering using remote sensing images with paired captions, object references, and question-answers. UrBench \cite{zhou2024urbench} evaluates multi-modal capabilities in urban environments, with questions spanning geo-localization, scene understanding, and object understanding. VLEO-Bench \cite{zhang2024vleobench} presents 11 real-world scenarios to test VLMs' abilities in scene understanding, localization and counting, and change detection. However, evaluating VLMs falls outside the scope of this work. Our benchmark focuses on dense-label tasks crucial for fine-grained spatial understanding, whereas vision-language tasks often center around image captioning, high-level scene interpretation, or object recognition, which differ from the dense prediction tasks we emphasize.

In the next section, we outline the parameters we have selected to evaluate the significance of a benchmark and describe the methodology we employed to identify the most effective one.

\section{The PANGAEA Benchmark}
\label{sec:meth}

\subsection{Datasets Selection}
\label{sec:meth-data}

\input{tabs/datasets_composition}

To establish a new-generation benchmark aligned with real-world applications across diverse scenarios, we select datasets based on five key parameters: i) \textbf{application domain}, ii)  \textbf{type of task}, iii) \textbf{multi-modality}, iv) \textbf{temporality}, and v) \textbf{geographical diversity}.

Evaluating the effectiveness of vision GFMs requires assessing their performance across diverse \textbf{application domains} to understand their adaptability and generalization capabilities. Each domain presents unique challenges: models pre-trained on agricultural datasets may perform well in crop classification but struggle with marine tasks like pollutant detection, while those optimized for urban environments may underperform in rural or forested areas. To identify a model's generalization ability and versatility, it is essential to consider datasets tailored to various domains.
For instance, datasets such as SpaceNet 7 \cite{vanetten2019spacenet} and DynamicEarthNet \cite{toker2022dynamicearthnet} are used for tasks like \textbf{urban} land cover or building footprint segmentation.
Crop Type Mapping-South Sudan \cite{m2019semantic} (referred to as Crop Type Mapping-SS) and PASTIS-R \cite{garnot2021panoptic, garnot2021mmfusion} are employed to identify and map crop types in \textbf{agricultural} landscapes.
Datasets like MADOS \cite{KIKAKI202439} and BioMassters \cite{nascetti2023biomassters} \cite{agrafiotis2024magicbathynet} support the development of models for tasks in \textbf{marine} and \textbf{forest} environments.
For natural \textbf{disaster} damage assessment, datasets such as xView2 \cite{gupta2019xbd} and Sen1Floods11 \cite{Bonafilia_2020_CVPR_Workshops} are invaluable, addressing scenarios from earthquakes to floods. 
\input{figs_tex/dataset_dist}
In addition, it is essential to assess the versatility of models not only across domains but also across \textbf{downstream tasks}. For example, within the urban domain, GFMs optimized for semantic segmentation of land cover may struggle with change detection (e.g. building damage assessment) or regression tasks (e.g. building height estimation). To address this, we include a diverse set of downstream tasks, including semantic segmentation, change detection, and regression tasks, with a focus on dense, pixel-wise problems that hold significant real-world relevance in remote sensing.
As previously stated, we exclude classification tasks, as they are less relevant for real-world application, as also shown by the community shift towards more dense tasks. We also excluded object detection, following other benchmarks \cite{lacoste2023geobench}, for two main reasons: there is no consensus on how to evaluate GFMs for detection, making it out of scope for the moment; dense tasks like semantic segmentation can achieve similar real-world results (\ie segmenting one class).

Another key consideration in shaping the benchmark is incorporating \textbf{multi-modality} (i.e. data from different sensors or sources such as LiDAR, SAR, optical, or text). Different satellites are equipped with sensors of varying resolutions and spectral properties. Incorporating multi-modal datasets ensures the benchmark evaluates a GFM's ability to adapt to diverse sensing conditions, enhancing robustness for real-world applications. Currently, we focus on optical and SAR modalities due to the limited availability of datasets for other modalities, such as hyperspectral data.


EO data is inherently \textbf{multi-temporal} (i.e. data from the same location captured at different timeframes). However, not all models leverage this dimension. Temporal information is crucial for tasks such as biomass tracking, urban expansion monitoring, and seasonal crop assessment. Evaluating GFMs on temporal data provides a more comprehensive assessment of their capabilities to determine whether these models can interpret dynamic processes.

Finally, concerning \textbf{geographic diversity}, \cite{Schmitt_2023} reports that, in their vast sampling of the available EO datasets, only 25\% cover locations all over the world and almost 40\% of the available datasets focus on "first-world countries" (Europe 21\% and North America 18\%). Asia is covered by 10\% of the datasets, while Africa (5\%), South America (4\%), and Australia (1\%) are significantly underrepresented.
Many GFMs perform poorly in underrepresented regions such as Africa or Southeast Asia, due to biases in the pre-training datasets. By building a geographically diverse dataset selection on a global scale, we aim to alleviate the biases in the evaluation datasets, thus enabling more robust assessments of models and their susceptibility to pre-training biases.

To evaluate GFM across all these dimensions, we curate a diverse and comprehensive set of datasets to compose the PANGAEA benchmark, spanning various domains, tasks, locations, temporalities, and satellite sources, detailed in \cref{tab:datasets-selection}.
We include PASTIS-R \cite{garnot2021panoptic, garnot2021mmfusion} and AI4SmallFarms \cite{ai4smallfarms} ensuring that the benchmark evaluates models on the high-impact applications that are crop mapping and field boundary segmentation, across different regions, leveraging different modalities (\ie Sentinel-1 and Sentinel-2) and multi-temporal data. 
The inclusion of MADOS \cite{KIKAKI202439} in the marine domain adds a global perspective on marine pollutant mapping, addressing worldwide environmental challenges.
Disaster mapping datasets like xView2 \cite{gupta2019xbd}, Sen1Floods11 \cite{rambour2020flood}, and HLS Burn Scars \cite{jakubik2023foundation} test GFMs' abilities to detect and respond to sudden changes, such as natural disasters, across various global locations and temporalities. 
Datasets such as SpaceNet 7 \cite{vanetten2019spacenet} and DynamicEarthNet \cite{toker2022dynamicearthnet} challenge models with tasks like urban change detection and multi-temporal land cover analysis, ensuring robustness in dynamic urban and land cover scenarios.
In the forest domain, BioMassters \cite{nascetti2023biomassters} provides a multi-modal, multi-temporal dataset for forest biomass estimation, a critical task for monitoring forest health and carbon sequestration. By incorporating BioMassters, we also add robust and long-overdue evaluation of GFMs on regression tasks. 
Geographic diversity is further enhanced by datasets such as Five Billion Pixels \cite{tong2023enabling} and AI4SmallFarms \cite{ai4smallfarms} (Asia), as well as Crop Type Mapping-SS \cite{yeh2021sustainbench, rustowicz2019semantic} (Africa), ensuring that the benchmark evaluates GFMs in underrepresented regions with diverse satellite data, including Gaofen and Sentinel-1/2. The spatial distribution of the PANGAEA benchmark datasets across domains is shown in \cref{fig:datasets-dist}.

This extensive coverage of PANGAEA across domains, tasks, modalities, temporalities, and locations ensures that the benchmark is robust, enabling a thorough evaluation of model generalizability, adaptability, and performance in real-world remote sensing applications.

\subsection{Models Selection}
\label{sec:meth-model}

To efficiently benchmark the performance of different GFMs, PANGAEA includes nine models selected based on three criteria: \textbf{reproducibility}, \textbf{approach} and \textbf{impact}. The models are listed alphabetically in~\cref{tab:pretraining-datasets}, along with their corresponding image sources and the data volume used for pre-training.

To ensure reproducibility, we require each model to have open-source code and publicly available weights. With this condition met, we select a diverse set of models, each representing a unique approach to highlighting their strengths in remote sensing and geospatial analysis.

SSL4EO-S12 \cite{stewart2024ssl4eo} is selected as a GFM baseline. This includes four vision-transformer (ViT)-based models, each pre-trained with different self-supervised learning approaches, such as masked image modeling (\textbf{S12-MAE}), contrastive learning (\textbf{S12-MoCo}) and knowledge distillation (\textbf{S12-DINO}, \textbf{S12-Data2Vec}). Since these models employ original computer vision (CV) algorithms without EO-specific modifications but are pre-trained on large-scale remote sensing imagery, they serve as a baseline to provide insights into the 
performance of CV models without adjustments for the peculiarities of EO data.
\input{tabs/model_pretraining_dataset}

\input{tabs/model_lists}
Several models are selected based on the widespread masked image modeling framework, each with a different focus.
\textbf{Scale-MAE} \cite{reed2023scalemae} is included for its ability to handle input images under varying resolutions, demonstrating strong performance and widespread applicability.
\textbf{SpectralGPT} \cite{Hong2024spectralgpt} is selected due to its explicit design for spectral satellite images, addressing the unique characteristics of spectral data. Similarly, to effectively deal with spectral bands, \textbf{DOFA} \cite{xiong2024neural} employs a dynamic network weights generator based on the central wavelengths of each spectral band, allowing generalization to different modalities
\textbf{Prithvi} \cite{jakubik2023foundation} has made a significant impact by training an MAE on NASA’s massive Harmonized Landsat Sentinel-2 (HLS) data. It is also one of the few models to incorporate the temporal dimension in its pre-training strategy, expanding the 2D MAE to a 3D version to utilize 3D spatiotemporal data.
Among other MIM models, we select \textbf{GFM-Swin} \cite{mendieta2023gfm}, which leverages multi-modality data for its pre-training to also learn the cross-sensor reconstruction.

In addition to MIM-based approaches, we also include models that employ contrastive learning strategies. While the former methods are faster to train, several studies \cite{marsocci2024crosssensor, mendieta2023gfm} demonstrate the advantages of contrastive learning which relies less on large quantities of high-resolution images and more on auxiliary data, such as text or additional modalities. In this context, \textbf{RemoteCLIP} \cite{liu2024remoteclip} and \textbf{CROMA} \cite{fuller2023croma} are included. \textbf{RemoteCLIP} is trained to align vision-language representations and learns robust visual features with rich semantics of satellite imagery visual concepts. \textbf{CROMA}, on the other hand, aligns geographically and temporally matched radar and optical image pairs while also combining unimodality MIM training objectives, making it a valuable inclusion.

Furthermore, \textbf{SatlasNet} \cite{bastani2022satlas}, one of the few GFMs based on fully supervised learning, is included to demonstrate the potential and limitations of this approach in geospatial tasks.
 
By including these models in PANGAEA, we aim to investigate how different model designs contribute to variations in performance across various tasks and conditions and determine which strategies are best suited for advancing the field of geospatial analysis. 
The pre-training data sources leveraged by GFMs play a key role in achieving robust representations for geospatial applications. This factor is considered in  \cref{tab:pretraining-datasets}, which showcases the data sources and scales utilized by the selected models in PANGAEA, serving as a critical aspect in evaluating the capabilities of these GFMs.

\section{Experimental Protocol}
\label{sec:exps}

PANGAEA aims to extensively evaluate GFMs more robustly by introducing a standardized evaluation protocol on the 11 benchmark datasets.
To do so, we integrate all selected models and datasets, alongside task-specific fine-tuning and evaluation code into a unified framework.
The protocol incorporates standardized approaches to ensure compatibility between downstream tasks and models, such as consistent band adaptation alongside multi-modal and multi-temporal evaluation strategies.

\paragraph{GFM Evaluation on Downstream Task} 
As commonly done to evaluate self-supervised models \cite{reed2023scalemae, ssl4eo}, we evaluate the representation abilities of the models using frozen encoders.
We treat the encoders of various foundation models as feature extractors by freezing their pre-trained weights.
Because PANGAEA relies on dense pixel-level tasks and not all GFM encoders produce dense features, we then plug a trainable decoder to upsample the features to match the input image resolution.
This decoder performs either regression or pixel-level classification depending on the task.
Following existing works that evaluate GFM on dense tasks \cite{reed2023scalemae, Hong2024spectralgpt}, we implement a UPerNet~\cite{xiao2018unified} decoder, that takes as input four intermediate feature levels from the foundation model encoders, as shown in \cref{fig:uni-temporal}.
We select features from evenly spaced layers depending on the specific GFM architecture.
The decoder is then trained on the datasets under various scenarios, including fine-tuning with limited labeled data (10\%, 50\%, and 100\% of each dataset).
We compare these fine-tuned GFMs against two common fully supervised baselines: a UNet~\cite{ronneberger2015u}, representing a convolutional neural network backbone, and a Vision Transformer~\cite{dosovitskiy2020image}, representing a transformer backbone (ViT-B/16). Both these baselines are trained from scratch using all available bands from the downstream dataset.

Based on this evaluation protocol, we compute the complexity and parameter count for each model, as shown in \cref{tab:flops}. These metrics provide insights into model size and computational efficiency, helping to assess the trade-offs between performance and resource requirements.

\paragraph{Data Preprocessing}
All models are trained using the same preprocessed dataset.
Because different foundation models expect different image sizes, we either crop (for larger images) or resize (for smaller images) the input images as needed.
We normalize the images by removing the mean and dividing by the standard deviation, computed for each band across every dataset.
This ensures consistent scaling across all images from every dataset.
In data-scarce scenarios, we implement stratified sampling to preserve the overall distribution of the dataset. For segmentation tasks, we first calculate the category distributions for the dataset, organizing samples into bins that represent unique combinations of category proportions. By grouping samples into these bins, we can then draw a proportionate number (e.g. 10\%, 50\%) of samples from each bin, ensuring that the original category distribution is maintained. For regression, the mean value of each image is computed and binned into intervals, enabling proportional sampling from each bin. 

\paragraph{Bands Matching and Adaptation}
Foundation models are pre-trained on diverse sensors and spectral bands, which often do not align precisely with those available in downstream datasets (e.g. a GFM trained on RGB images cannot directly process multi-spectral Sentinel-2 imagery).
To address this, we adopt a common band adaptation strategy for all models.
If the bands used to pre-train the GFM are included in bands from the downstream dataset, then we match those. For example, GFMs trained on RGB imagery will only use the red, green and blue bands from datasets like PASTIS-R, which is comprised of Sentinel-2 images.
If certain bands are unavailable in the dataset, we apply zero-padding for those missing bands, as proposed by  \cite{hsu2024geospatial}. 
For example, GFMs trained on Sentinel-2 data are still evaluated on RGB-only datasets, such as xView2 (which uses color imagery from Maxar), however the missing bands are filled with zeros. 

This strategy also enables the models to address multi-modal data by ensuring band matching across all sensor modalities. Among the selected models, only CROMA and DOFA can jointly handle optical and SAR data. We experiment with these two models using either unimodal data or both modalities for datasets containing both optical and SAR images.

\paragraph{Multi-temporal Strategy}
Although we have selected multi-temporal datasets as downstream tasks, not all considered GFMs have the capability to handle image time series.
In our selection, only SatlasNet and Prithvi natively handle temporal information.
SatlasNet employs a dedicated aggregation network that processes individual images through a shared encoder and aggregates features via a pooling strategy. Prithvi incorporates the temporal dimension directly within its patch embedding process.
For these models, we directly feed the decoder with the multi-temporal representations extracted from the encoder.
As illustrated in \cref{fig:bi-temporal}, for bi-temporal datasets such as change detection, we pass both images to the same GFM encoder and then concatenate the features before passing them to the UperNet decoder.
To enable models that only handle single images to be trained on multi-temporal datasets, we introduce and compare two feature aggregation strategies: a naive linear mapping and a lightweight spatial-temporal encoder (L-TAE) \cite{saintefaregarnot_lightweight_2020}. For models supporting only single-timeframe input, each image in the time series from the multi-temporal dataset is passed independently through the encoder. The linear mapping then applies a layer to compress features across the temporal dimension, while L-TAE uses a temporal attention module to capture relationships between frames over time.
The temporally aggregated features are then fed through the UPerNet decoder as before, as shown in \cref{fig:multi-temporal}.

\input{figs_tex/temporal_strategy}

\paragraph{Hyperparameters}
For a fair comparison across models, the decoder is fine-tuned with an identical set of hyperparameters.
For every model and dataset combination, the decoder is fine-tuned for 80 epochs using the Adam optimizer with an initial learning rate of 1e-4, a batch size of 8, and a weight decay of 0.05.
The learning is decreased by a factor of 0.1 using a multi-step schedule at 60\% and 90\% of the training.
Some specific edge cases are detailed in \cref{appendix} for runs that necessitate adjustments due to memory constraints.
The best-performing checkpoint on the validation set is kept and used to evaluate the model on the test set.
For evaluation, to process larger images, we employ a sliding window inference strategy, dividing the input image into evenly distributed smaller crops. A stride is used to ensure the crops collectively cover the entire image with the fewest patches, and the overlapping areas are averaged.

\section{Experimental Results}
\label{sec:expr}

The code to reproduce the results is available at \url{https://github.com/VMarsocci/pangaea-bench}.

\subsection{Main Comparison}
\label{sec:comp}

\cref{tab:comp_100} presents the experimental results of GFM evaluation using the full dataset across all 11 benchmark datasets, listed alphabetically. For multi-temporal datasets, the reported results reflect the use of the L-TAE feature aggregation, except for a few models that inherently support multi-temporal data. We include two fully supervised baselines for comparison, UNet and ViT, trained from scratch in an end-to-end manner without any pre-trained weights. These results provide valuable insights from multiple perspectives. Extended results are presented in \cref{appendix}.

\textbf{Impact of Resolution on Model Performance.} GFMs pre-trained on lower-resolution data tend to perform worse on high-resolution datasets with limited spectral information, such as Five Billion Pixels (FBP) and SpaceNet 7 (see \cref{fig:teaser}). GFMs like Scale-MAE are often among the best-performing models, reaching e.g. 67.19\% mIoU on FBP and 62.96\% mIoU on SpaceNet 7. In contrast, models pre-trained on lower resolutions, such as S12-DINO (51.15\% on FBP) and S12-Data2Vec (48.82\% on FBP), show lower performance. This trend is especially evident in tasks without extensive spectral information, where spatial detail becomes essential. Observe, for example, MADOS, where GFM-Swin achieves a high 64.71\% mIoU, but models like SSL4EO12-MAE (49.90\% mIoU) and SSL4EO12-DINO (49.37\% mIoU) lag behind, showing the advantage of pre-training on high-resolution data. On the other hand, in some cases, when there is a match in the pre-training data and the downstream task data we can observe an average increase in the performance (e.g. Prithvi pre-trained on Harmonized Landsat Sentinel-2 data performs strongly on HLS Burn Scars). When compared to supervised baselines, we can observe that GFMs struggle especially on low-resolution tasks, probably because features useful for these downstream tasks are easier to learn from not specialized architecture (\ie UNet), due to their reduced available dense information. On the other hand, on high-resolution tasks, GFMs, pre-trained on similar resolution data, still offer strong performance (\cref{fig:teaser}).

\input{tabs/main_miou}
\textbf{Role of Spectral Information in Task-Specific Performance.}
Spectral richness significantly impacts tasks that require fine spectral detail (e.g. marine, agricultural, and forest domains). In datasets such as MADOS, Crop Type Mapping-SS, and BioMassters, GFMs trained with a wider spectral range perform best. CROMA reaches 67.55\% mIoU on MADOS and 49.38\% mIoU on Crop Type Mapping-SS. Similarly, SpectralGPT along with CROMA are among the top-performing models with low RMSE on BioMassters, suggesting that models that use diverse spectral data capture essential distinctions, whereas models without this range (e.g. Scale-MAE reaches 47.15 RMSE on BioMassters) underperform. The trend on HLS Burn Scars illustrates this: trained on a wider set of bands reach better performance on average. On simpler tasks where spectral information is less crucial, models with fewer spectral inputs are competitive. For example, GFM-Swin excels in FBP and SpaceNet 7, but struggles on BioMassters, indicating that spatial features can compensate for lower spectral diversity for urban land cover tasks, especially as the downstream datasets use RGB imagery only. Similar observations hold for the supervised baselines that can leverage all of the spectral information and deliver strong performances on low-resolution tasks (\eg UNet reaches 84.51\% and ViT 81.58\% on HLS Burn Scars).

\textbf{Effectiveness of Text-Based Models for Multi-Class Semantic Segmentation Datasets.}
The text-augmented RemoteCLIP model shows advantages in complex, multi-class tasks where class boundaries are subtle, such as MADOS (where it reaches third place with 60.00\% mIoU) and FBP (first place with 69.19\% mIoU). This success likely comes from RemoteCLIP’s ability to leverage text-based information during training to learn features that can differentiate visually similar classes.
However, RemoteCLIP underperforms on simpler tasks, \eg scores only 74.26\% mIoU on Sen1Floods11, compared to S12-MoCo (89.26\% mIoU) and CROMA (90.89\% mIoU). This limitation may be due to its RGB-only training, which restricts its potential in tasks that benefit from a broader spectral range.

\textbf{Strong Convolutional Baseline.}
For simpler datasets with fewer classes, the convolutional UNet baseline excels. UNet achieves the best results on HLS Burn Scars (84.51 \% mIoU), Sen1Floods11 (91.42 \% mIoU), and AI4SmallFarms (46.34 \% mIoU), showing that such a simple convolutional model is well-suited for simpler datasets. Overall, our results indicate that learning the features through convolution is the best approach for one-class datasets. On the other hand, when there is a domain gap between pre-training and downstream task (\ie regression), performance is so poor that even supervised baselines do better than GFMs (35.67 RMSE w.r.t. second best 36.11 RMSE with SpectralGPT). Thus, UNet is the model that reaches top-2 performance (across all models) the biggest number of times (5 times w.r.t to CROMA's 3 and Scale-MAE 3 times).


\textbf{Challenges with Multi-Temporal Tasks.}
Most GFMs do not consider satellite image time series in their design, and the few GFMs that are designed to leverage the temporal dimension of EO data (i.e. Prithvi and SatlasNet) performed comparatively low on multi-temporal tasks. Prithvi scored the lowest mIoU on DynamicEarthNet (27.86\%) and SatlasNet scored the lowest mIoU on PASTIS-R (17.51\%). Both Prithvi and SatlasNet were outperformed by several uni-temporal GFMs when combined with a temporal module (\ie L-TAE). This is stressed by the results on all multi-temporal datasets we evaluated; PASTIS-R, DynamicEarthNet, Crop Type Mapping-SS, and BioMassters. Most notably, the uni-temporal models such as CROMA achieved better scores (32.32\% mIoU on PASTIS-R, 42.12\% mIoU on DynamicEarthNet, and 36.81 RMSE on BioMassters) than both Prithvi and SatlasNet. Therefore, our results highlight the need for dedicated multi-temporal foundation models capable of effectively capturing temporal patterns in satellite image time series.

\textbf{Challenges with Multi-Modal Datasets.}
While some models like S12-MAE, CROMA, and DOFA incorporate SAR data through specialized encoders or dynamic wavelength adjustments, their performance on multi-modal datasets often falls short of expectations. As demonstrated in \cref{tab:pastis-linear-ltae-100,tab:extensive-ctm-100perc,tab:extensive-biomass}, utilizing optical data alone generally yields superior results compared to incorporating SAR information, whether solely or in combination with optical data. This performance gap indicates that effectively leveraging multi-modal data, particularly the fusion of optical and SAR imagery, remains a significant challenge for GFMs.

\subsection{Data Scarcity}
\label{sec:scarc}

\input{tabs/50_percent_miou}

\input{tabs/10_percent_miou}

We further evaluate the performance of the selected GFMs by reducing the percentage of labeled data used for training to 50\% and 10\%, while preserving the original data distribution. The results are summarized in~\cref{tab:com-50,tab:com-10}, respectively, providing insights into the models' performance under limited supervision, simulating real-world scenarios where labeled data is often scarce.

A general trend of performance degradation is observed across various EO tasks. This observation aligns with the well-established principle that models typically benefit from larger training sets, which provide a more comprehensive representation of the data distribution and thereby enable better generalization. 


Nevertheless, with limited training samples, a compelling trend emerges where several GFMs demonstrate notable strengths. We report the number of times each model achieves a top-2 performance across all datasets. While the fully supervised UNet baseline performs best in this score, this metric shifts in favor of GFMs as the amount of labeled data decreases. For instance, UNet achieves a score of 6 for this metric with fully labeled datasets, but its score decreases to 4 and 2 for the 50\% and 10\% labeled data scenarios, respectively. In contrast, some GFMs show improved performance, with CROMA achieving a score of 6 in the 10\% labeled data scenario. This trend is also evident in datasets such as HLS Burn Scars and SpaceNet 7.

This phenomenon can be attributed to the better generalization capabilities of foundation models compared to fully supervised baselines, which stem from their exposure to a vast and diverse range of data during pre-training.  In situations where labeled data is scarce, foundation models can effectively leverage the knowledge acquired during pre-training to compensate for the lack of task-specific labeled data, highlighting the potential of foundation models to address challenges in EO tasks where acquiring labeled data is often costly and time-consuming.

\subsection{Normalization Impact}
\label{sec:norm}

\input{tabs/abl_wo_norm}

In this ablation study, we examine the impact of the normalization strategy during the decoder fine-tuning stage. While our experiments normalized images using the mean and standard deviation of each band (following most of the original papers), this step was omitted in this ablation study to evaluate its effects. The analysis was conducted on both low-resolution datasets (HLS Burn Scars and MADOS) and a high-resolution dataset (SpaceNet 7).

The results in \cref{tab:abl-norm} compare evaluation performance with and without the normalization strategy. Generally, normalization affects low-resolution datasets more significantly than high-resolution ones. In the high-resolution dataset SpaceNet 7, removing normalization has a minimal impact on performance; for example, the largest percentage difference is observed with RemoteCLIP, which shows only a 4\% drop in mIoU. However, removing normalization causes quite large effects on low-resolution datasets. Particularly, models pre-trained on high-resolution data, are more susceptible to the absence of normalization in this setting. For instance, the performance of models like GFM-Swin, Scale-MAE, and RemoteCLIP drops significantly when applied to unnormalized data on both the HLS Burn Scars and MADOS datasets.  

\subsection{Matching Training Resolution of the Model}
In this ablation study, we examine how adjusting the input resolution of test data to more closely match the resolution on which the GFMs were trained during the pre-training stage affects model performance. Specifically, we select models pre-trained on low-resolution data (i.e., S12-DINO and CROMA, pre-trained on 10m Sentinel-2 data, and Prithvi, pre-trained on 30m Harmonized Landsat and Sentinel-2 (HLS) data) and perform ablations on the high-resolution dataset FBP. For the FBP dataset, instead of cropping the images (as in previous experiments), we resized them using bilinear interpolation, effectively reducing the spatial resolution of the input data from 2 m/pixel to match the training resolution of the models.

The results in \cref{tab:abl_inp_size} show that all models experience a significant performance drop when their input resolution is reduced to match the training resolutions. For instance, Prithvi's performance decreases from 46.81\% mIoU to 30.99\% mIoU, and CROMA’s score drops from 51.83\% mIoU to 21.10\% mIoU. This highlights that fine-grained spatial details are essential for complex datasets like FBP, which involve many visually similar classes, even for models trained on lower-resolution data.

\input{tabs/abl_match_input_resolution}


\subsection{Temporal Capabilities and Aggregation Strategies}

We employ two distinct feature aggregation strategies for multi-temporal datasets (PASTIS-R, Crop Type Mapping-SS, BioMassters and DynamicEarthNet): a naive linear mapping and L-TAE. A comparative analysis, presented in \cref{tab:temp},
reveals that L-TAE consistently enhances performance compared to the linear approach in all the datasets but DynamicEarthNet.
Notably, L-TAE yields improvements ranging from 10\% to 20\% for the top-performing models on the PASTIS-R dataset.
On DynamicEarthNet, however, L-TAE may lead to overfitting due to the minor temporal window spanned by the time series (6 days). See \cref{appendix} for more insights.
Generally speaking, these findings underscore the importance of effectively encoding temporal information in remote sensing tasks and highlight the potential of developing more sophisticated temporal aggregation strategies within GFMs.
\input{tabs/multitemp.tex}

\subsection{Domain Adaptation}

\input{tabs/domain_adaptation_fbps}


The cross-region adaptation task on FBP presents a great challenge, primarily due to spatial autocorrelation. The results are reported in \cref{tab:dom_ad_fbps}. Here, the class distribution differs significantly from the in-domain test, with the absence of five minor classes that are present in the training data. This discrepancy in class representation greatly impacts performance. For instance, CROMA, which performs competitively in cross-sensor tests, sees a reduction to 18.38\% in cross-region tests, while other strong GFMs like RemoteCLIP and GFM-Swin score 36.53\% mIoU and 35.39\% mIoU, respectively. 

\subsection{Consideration on the Training Convergence}

\cref{fig:training-losses} illustrates the training loss trajectories of some of the most illustrative GFMs -- Scale-MAE, RemoteCLIP, UNet, CROMA, and DOFA -- across four datasets: FBP, MADOS, HLS Burn Scars, and Sen1Floods11. The trends reveal important insights into the GFMs' behavior depending on dataset difficulty and pre-training.

For complex datasets like FBP and MADOS, the pre-trained GFMs -- with frozen encoders -- tend to begin with lower loss values. This suggests that these models possess beneficial pre-learned features, enabling them to adapt more readily to challenging datasets without requiring significant fine-tuning. By contrast, UNet, which is end-to-end fine-tuned, starts with a higher loss but reduces it steadily, showing that while it initially lags, it progressively learns the data patterns, even if not as efficiently as models with strong initialization from pre-training.

In contrast, on easier datasets like HLS Burn Scars and Sen1Floods11, the training loss for all models, including the pre-trained ones, starts at relatively low values, indicating that the data may be less complex or more straightforward for feature extraction. In these cases, pre-trained models do not appear to retain a distinct advantage in initial loss values over UNet, which starts on nearly the same level. This behavior aligns with the fact that UNet, with end-to-end fine-tuning, performs well on simpler datasets, as it can optimize its parameters fully to match the dataset characteristics. Interestingly, on Sen1Floods11, UNet's stronger performance despite higher training loss values hints at potential overfitting by the foundation models, as they may overly adapt to training data without capturing the generalization needed for evaluation.


\input{figs_tex/training_losses}

\subsection{Impact of Encoder Fine-tuning}

\input{tabs/finetuning_ablation}

In \cref{tab:finetuning-gfm}, we analyzed the effect of end-to-end fine-tuning on the performance of one of the best and worst-performing GFMs across all the datasets. 
From the results, we can observe that the impact of fine-tuning is not consistent across the GFMs and datasets. In some cases, fine-tuning improves the performance, while in others, it results in a decline. For instance, RemoteCLIP shows a noticeable drop in performance on HLS BurnsScars, with the model's performance dropping from 76.59\% (without fine-tuning) to 65.13\% (with fine-tuning) mIoU. This suggests that fine-tuning might have caused the model to lose some of the encoded information from the pre-trained encoder, possibly related to text features. 
On the other hand, GFMs pre-trained on low-resolution data tend to exhibit more positive results, with constant or better performance. For example, CROMA, shows stability in performance after fine-tuning on MADOS, moving from 67.55\% to 66.98\% mIoU. DOFA shows improved results after fine-tuning on FBP, from 43.82\% to 63.74\% mIoU, suggesting that fine-tuning may help refine models trained on lower-resolution data to adapt more effectively to task-specific features.
Overall, the lack of a clear, consistent pattern highlights that fine-tuning does not always guarantee an improvement and that the success of fine-tuning depends on the GFM's pre-training, dataset characteristics, and the specific task at hand. Moreover, fine-tuning requires more resources, given the need for backpropagating the gradient on all the parameters.

\subsection{Sensitivity to Hyperparameters}

For PANGAEA, we aim to adopt a standardized protocol that avoids the need for extensive hyperparameter fine-tuning for every new GFM tested. To achieve this, we set up an environment where learning rate tuning should have the least possible impact. The ADAM optimizer used in our experiments is known to be robust to changes in learning rate. In addition, the datasets are normalized, the decoder architecture is the same for each experiment, and it includes batch normalization layers. Because of this, we expect the magnitude of gradients to be fairly even between models.

To verify this assumption, we optimize the learning rate on the MADOS and HLS Burn Scars datasets by evaluating seven possible learning rates over a log scale, from $10^{-2}$ down to $10^{-5}$. \cref{tab:lr-tuning} shows the results with the default learning rate of $10^{-4}$, versus the best result of the optimization for the specific model-dataset combination, and their relative difference.

The results indicate that in general our assumption stands, and the experiments are fairly stable under learning rate tuning. In the specific case of MADOS and the S12 family, the results do improve by tuning the learning rate, but even with this improvement, the overall ranking of models doesn't change significantly. The S12 family of models is still in close competition with Prithvi for the worst-performing GFM on this dataset.

This learning rate agnosticism is even more pronounced on the HLS Burn Scars experiment, where the results change even less. One interesting detail is that some models lose performance in the tuning experiments, meaning that the stochasticity of running multiple training runs is in the same order of magnitude as the gains achievable by learning rate tuning.

\input{tabs/lr_tuning}

\section{Limitations}
Despite its contributions, this work has some limitations. First, the focus is predominantly on computer vision-based EO tasks, leaving out non-vision applications such as climate modeling or weather forecasting. Second, the computational resources required for training and fine-tuning GFMs may limit accessibility for resource-constrained environments. Additionally, while Pangaea includes diverse datasets, there is room for expanding geographical and application domain coverage. Future work should also explore multi-sensor data integration beyond satellite imagery, incorporating aerial and drone datasets to further enhance GFM applicability. Finally, from a protocol standardization point of view, we foresee releasing new features that can help the fast and consistent reproducibility of our results and the benchmark of newly added GFMs. These features include the release of a fixed stratified subsample for each dataset, and an arena-mode benchmarking, with fixed command lines and randomness to launch experiments and obtain final scores.

\section{Conclusions}
The Pangaea Benchmark addresses critical gaps in the evaluation of GFMs. By incorporating diverse downstream tasks (i.e. semantic segmentation, change detection, and regression) and employing datasets that encompass various resolutions, sensor types, and application domains (e.g. urban, agriculture, forestry), it provides a robust framework for assessing GFM generalization and performance under real-world conditions. We advocate for surveying and testing a representative set of GFM approaches, including vision language-based models and fully-supervised learning paradigms, to gain a deeper understanding of their strengths and weaknesses. Our analysis demonstrates the strengths and weaknesses of GFMs under different scenarios (e.g. label scarcity, domain shift, etc...), especially w.r.t. specialized baseline, that most of the time can still be a strong counterpart. Furthermore, the open-source nature of this benchmark fosters transparency and collaboration, driving progress in the development of adaptable and effective GFMs.

\section*{Contributors} \textbf{Conceptualization}. VM AN NA \textbf{Methodology}. VM AN NA \textbf{Codebase prototype}. VM YJ GL DK LZ \textbf{Semantic Segmentation}. YJ VM DK GL LZ EB AS \textbf{Change Detection}. SH SG HF \textbf{Regression}. RY EB \textbf{Ablation Studies}. VM HF DK YJ RY GL 
\textbf{Manuscript Revision}. AN MV NA \textbf{Supervision}. AN NA MV YB \textbf{Fundings}. AN NA MV YB

\bibliographystyle{unsrt}  
\bibliography{references}  

\newpage

\renewcommand*{\thesection}{\Alph{section}}
\newcommand{\multiref}[2]{\cref{#1}--\ref{#2}}
\renewcommand{\thetable}{A\arabic{table}}
\renewcommand{\thefigure}{A\arabic{figure}}
\setcounter{section}{0}
\setcounter{figure}{0}
\setcounter{table}{0}

\section*{Appendix}
\appendix{}

\section{Dataset Details and Extensive Results}
\label{appendix}

Below, we provide a detailed description of the benchmark datasets, along with the extensive experimental results including additional metrics.

\subsection{HLS Burn Scars}

HLS Burn Scars \cite{jakubik2023foundation} contains Harmonized Landsat and Sentinel-2 imagery of burn scars and the associated masks for 2018-2021 over the contiguous United States. There are 804 512x512 scenes. Scenes contain six bands (Blue - B02, Green - B03, Red  - B04, NIR  - B8A, SW 1  - B11, SW 2  - B12), and masks have one band (with values 1 = Burn scar, 0 = Not burned, -1 = Missing data). For satellite scenes, each band has already been converted to reflectance. The class distribution is 11\% for Burn Scar, 88\% for Not Burned, and 1\% for No Data. The 804 files have been randomly split into training (2/3) and validation (1/3) directories, each containing the masks, scenes, and index files.
To create the dataset, after co-locating the shapefile and HLS scene, the 512x512 chip was formed by taking a window with the burn scar in the center. Burn scars near the edges of HLS tiles are offset from the center.
Images were manually filtered for cloud cover and missing data to provide as clean a scene as possible and burn scar presence was also manually verified. Some example images and labels are displayed in~\cref{fig:hls-examples}.
\input{figs_tex/hlsburnscars_sample}

\subsubsection{Extensive Results}

\input{tabs/extensive_hlsburnscars}

Table \ref{tab:hlsburnscars_extensive} presents a comprehensive comparison of models on the HLS Burn Scars dataset using stratified sampling with 10\%, 50\%, and 100\% labeled data. 

The UNet Baseline achieves the best performance across all metrics for the 100\% label setting, with the highest mIoU (84.51\%), m-f1 (91.11\%), and m-Accuracy (97.19\%). This demonstrates the robustness of the UNet architecture when sufficient labeled data is available. 

The Prithvi model achieves the second-best results in the 100\% label setting, with mIoU (82.67\%), m-f1 (89.90\%), and m-Accuracy (96.55\%). These results closely follow the UNet Baseline, indicating its competitiveness as an alternative model.

At 10\% labels, SpectralGPT excels with the highest mIoU (83.35\%) and m-f1 (90.67\%), outperforming all other models. Its strong results in this low-label regime suggest robustness in low-supervision scenarios. However, its performance decreases at higher label proportions, which could indicate limitations in scalability or overfitting to specific data distributions.

Across all models, in average, performance improves significantly with increased labeled data, reflecting the importance of supervision for these tasks. The most substantial performance gains occur between 10\% and 50\% labels, indicating that moderate amounts of labeled data can significantly boost model effectiveness.

The ViT Baseline, added for comparison, demonstrates competitive performance at 100\% labels with mIoU (81.58\%), m-f1 (89.16\%), and m-Accuracy (96.26\%). While it does not surpass the UNet Baseline, its results highlight the potential of Vision Transformers for remote sensing tasks.

\subsection{MADOS}

The Marine Debris and Oil Spill (MADOS) \cite{KIKAKI202439} dataset contains marine pollutants under a diverse set of locations and sea surface features. The imagery contains multi-spectral (Bands 1-8A, 11 \& 12) Sentinel-2 data from 174 scenes, with approximately 1.5M annotated pixels and a resolution of 10 to 60 meters, depending on the band. After processing, the dataset contains 2803 tiles, each 240x240 pixels in size. The authors define train, validation, and test splits based on location and date, with an approximate ratio of 50\% / 25\% / 25\%.

Four major classes are assigned to different types of backgrounds: Marine Water (34.8\%), Turbid Water (11.7\%), Sediment-Laden Water(20.6\%), and Shallow Water(9.2\%). These and the Oil Spill (18.5\%) category represent around 95\% of all labeled pixels in the data, with the remaining 5\% of labels distributed among the remaining 10 classes. The 15 classes have a sparse distribution over scenes, and examples of multiple classes in a tile are exceedingly rare.

Most pixels in the dataset are unlabeled, causing some batches of randomly selected data to have no valid labels and consequently an invalid loss value. To avoid this problem we employ focus cropping, which only selects crops with at least one valid label. Some examples are shown in \cref{fig:mados-examples}
\input{figs_tex/mados_sample}

\subsubsection{Extensive Results}
\input{tabs/extensive_mados}


\autoref{tab:mados-results} presents a more detailed breakdown of results on the MADOS dataset. The mean accuracy metric places more emphasis on the more abundant categories, and is therefore useful for identifying GFMs that do not capture the rare classes well.

CROMA achieves outstanding results in both m-f1 and mIoU scores while coming in 4th on accuracy, and our evaluation of the per-class results confirms that this is because of its great performance in the rarer classes. We conjecture that the large variety of bands used by the model, together with its modality-focused pre-training helps with differentiating between the high number of visually similar classes.

GFM-Swin performs exceptionally well in the three major water categories while performing adequately in the rarer classes. This results in a high mean accuracy score, but for practical labeling of the target classes RemoteCLIP and SpectralGPT are both better candidates, since they achieve high IoU scores without over-focusing on the background categories.

Despite using all Sentinel-2 bands, SSL4EO models perform among the worst on the mIoU score, while keeping up in the mean accuracy metric. Investigation of the per-class results shows that the models perform well on water categories while performing significantly worse on the much rarer classes of interest. This could mean that the EO-specific adaptations that other models employ are important for detecting the rare classes. Another possible explanation is that the self-supervised methods combined with the mainly land-focused SSL4EO-S12 dataset do not provide good enough features for marine environments.

The best GFMs comfortably outperform both supervised baselines, highlighting the importance of foundation models for sparsely labeled datasets such as MADOS. Despite this, the UNet model achieves a very respectable score that places it in the middle of the pack.


\subsection{PASTIS-R}
\input{figs_tex/pastis_sample}
PASTIS is a comprehensive benchmark dataset designed for panoptic and semantic segmentation of agricultural parcels using satellite time series. It includes 2,433 image patches captured across the French metropolitan area, each with detailed panoptic annotations, providing both instance indices and semantic labels for every pixel. The original dataset is built using Sentinel-2 multi-spectral imagery, offering time series data with variable lengths for each patch. The Sentinel-2 available bands are B2, B3, B4, B5, B6, B7, B8, B8A, B11 and B12.
To test the performance of the model foundations on multi-modal settings, we used the PASTIS-R version of the dataset. This extended dataset introduces aligned Sentinel-1 radar observations. This extension includes approximately 70 radar observations from both ascending and descending orbits for each of the 2,433 patches, complementing the optical data from Sentinel-2. 
In line with the multi-temporal setup of this paper, we have kept 6 captures evenly distributed over time for each satellite time series.

\subsubsection{Extensive Results}
In the next paragraph, we present the results obtained on the PASTIS-R dataset (~\autoref{tab:pastis-linear-ltae-100}), before considering the results obtained by reducing the size of the dataset, i.e. by reducing the number of samples seen during training (~\autoref{tab:pastis-linear} and ~\autoref{tab:pastis-ltae}). 
Initially, we observe that two models stand out and achieve the best performance on the PASTIS-R dataset: CROMA on the optical modality and S12-DINO. Indeed, for all the metrics used in this benchmark, these two models produce the best results. The S12-DINO model is the strongest model when the temporal fusion strategy is linear, with a mIoU of 32.45. For the L-TAE strategy, the CROMA optical model performs best, with a mIoU of 36.18.

In general, L-TAE fusion produces better results than linear fusion, e.g. +6.46\% mIoU points for CROMA, +7.38\% mIoU points for s12-Data2Vec and +8.30\% mIoU points for s12-MoCo.

Interestingly, we can see that in the case of our setup and the PASTIS-R dataset, multi-modality doesn't seem to bring any particular gain: the performances obtained by CROMA joint, i.e. on the optical and SAR modalities are worse than those obtained only on the optical modality (32.32\% vs 38.51\% mIoU). This underperformance seems to be partly explained by the poor results obtained by the various models solely on SAR images: 21.61\% for CROMA + L-TAE and 14.20\% for S12-MAE.

By decreasing the proportion of the dataset seen during training (respectively ~\autoref{tab:pastis-linear} and ~\autoref{tab:pastis-ltae} for linear multi-temporal fusion and L-TAE) we observe that performance varies little, e.g. 34.32\% mIoU for s12-Data2Vec on the full dataset, 33.42\% mIoU on 50\% of the dataset and 33.09 on only 10\% (for L-TAE strategy). Nevertheless, a performance drop of 2.78 mIoU points (24.74 mIoU on 10\% and 27.53 mIoU on 100\%) is observed for SpectralGPT.

\input{tabs/extensive_pastis}

\subsection{Sen1Floods11}
Sen1Floods11\cite{Bonafilia_2020_CVPR_Workshops} offers global flood mapping, featuring two bands (VV and VH) from Sentinel-1, all spectral bands from Sentinel-2, and manually labeled surface water, which includes both permanent and flood water. The dataset comprises 4,831 chips, each with dimensions of 512 x 512 pixels at a 10-meter resolution, encompassing 11 distinct flood events and covering a total area of 120,406 square kilometers. The hand-labeled data is split into training, validation, and testing data using a random 60-20-20 split. The cloudy pixels are labeled and not considered during the experiments. An example sample is shown in~\autoref{fig:sen1floodss11-examples}.
\input{figs_tex/sen1floods11_sample}

\subsubsection{Extensive Results}
In the following, we present the comprehensive experimental results for the Sen1Floods11 flood mapping task. From~\autoref{tab:extensive-floods-scarcity}, we observe that, as a relatively simple task with only two classes, the fully supervised UNet outperforms all other models. The benefits of RemoteCLIP are less pronounced in this limited-class setting, and Scale-MAE demonstrates suboptimal performance with the 10-meter resolution data.

~\autoref{tab:extensive-floods-scarcity} illustrates the impact of label reduction. Even when labels are reduced by 50\% and 10\%, the fully supervised UNet continues to deliver superior performance. Additionally, both CROMA and Prithvi consistently achieve results comparable to those of the fully supervised setting. Given the simplicity of this task, the benefit of using pre-trained models is not as evident when compared to a lightweight fully supervised approach.

\input{tabs/extensive_sen1floods11_scarcity}

\subsection{xView2}
\label{sec:extensive-xview2}
The xBD dataset~\cite{gupta2019xbd} was used in the xView2 challenge and is thus often referred to as the xView2 dataset. It is a high-resolution, bi-temporal dataset focusing on detecting building damage, caused by various disasters, including floods, wildfires, earthquakes, hurricanes, and more. The images are sourced from Maxar satellites and offer RGB bands. Due to the difficulty of comparing damages caused by different events, the dataset authors use a four-level damage scale that is harmonized across disaster types. The damage is assessed per building, such that all pixels of a building have the same damage class. As a result, the dataset represents the task of change detection with one class representing 'no building' and four classes representing various levels of building damage caused by the respective disaster. All images have a size of $1024\times 1024$. The dataset is highly imbalanced, with the majority of pixels not containing buildings. Furthermore, the distribution of damage classes varies between disaster types, making this a rather challenging dataset. \autoref{fig:xview2-examples} shows example data.

\input{figs_tex/xview2_5bp_samples}

\subsubsection{Extensive Results}
\textbf{Details on experimental setup:} To counteract the strong class imbalance, we follow a simple oversampling approach from the winning solution~\cite{durnov_xview2_2020} of the xView2 competition: Images without buildings are seen once per training epoch, images with undamaged buildings are seen twice, and images with damaged buildings are seen thrice. Furthermore, when taking random crops in training, we consider multiple candidates and prefer those with more pixels of the underrepresented classes.
As the decoder model, we use a UPerNet that concatenates the two embeddings computed for the pre- and post-disaster images by the encoder. For the UNet baseline, we use an input side length of 256 pixels. The ViT uses an input side length of 224 pixels. 
Due to optimization issues that occurred when using Dice loss in initial experiments, we deviate from the default training protocol when training models on xView2: Instead of evaluating the model on the validation set every five epochs, we validate after every single epoch. Furthermore, we increase the batch size from the default 8 to 32, to decrease the likelihood of batches that contain no or very few buildings, and thus lead the model into the undesired local minimum. We use the cross-entropy loss for all experiments on xView2. For Table \ref{tab:finetuning-gfm}, we use the second-worst model Prithvi instead of the worst model SpectralGPT, due to the large computational requirements of training the whole SpectralGPT encoder.

\input{tabs/extensive_xview2}

The full results in \autoref{tab:ext-results-xview2} for 100\% of the data show that especially the minor and major damage classes are very hard to classify, which is also why we are oversampling them during training. No model achieves $\geq$30\% IoU on minor damage, which is likely due to the dataset imbalance and the difficulty of distinguishing this class from no damage or major damage. Scale-MAE performs consistently strongly, similar to DOFA, GFM-Swin, and RemoteCLIP. All of them are pre-trained on high-resolution data, which makes them well-equipped to extract features from the high-resolution xView2 images. On the other hand, SpectralGPT, the S12 models and Prithvi perform less well, likely due to them being pre-trained on lower-resolution data. This reinforces the intuition that there should be a match between the GFM and the target dataset to achieve a good performance. 

Compared to the UPerNets that are trained on top of the frozen GFM encoders, the two fully trained baselines perform surprisingly well. The UNet performs roughly as well as the GFMs trained on high-resolution images, which is surprising, given that it has fewer parameters than a UperNet decoder alone. The ViT model performs similarly well, while using ten times as many parameters as the UNet.

\subsection{Five Billion Pixels}

The Five Billion Pixels dataset is a large-scale, high-resolution land cover classification dataset based on Gaofen-2 satellite imagery, originally covering over 50,000 square kilometers of China. It contains more than 5 billion labeled pixels with a 4-meter spatial resolution, providing detailed annotations across 24 categories, including artificial, agricultural, and natural land covers. An image and label paired example is shown in~\cref{fig:fbps_example}. 

To include it in PANGAEA, we made key modifications to enhance its usability. We cropped the images into $520 \times 520$ patches, making them more manageable for model input. We then split it into train, validation, and test, making stratified sampling of 70\%, 10\%, and 20\%. Before performing this split, we created an out-of-domain test set (which wasn’t available in the original release), carefully selecting 22 patches from the northern regions of China to avoid spatial autocorrelation. 



\subsubsection{Extensive Results}
The results from \cref{tab:extensive-fbps} reveal interesting key insights regarding GFM performance, especially in light of pre-training data characteristics. Models pre-trained on high-resolution images, such as RemoteCLIP, GFM-Swin, and Scale-MAE, consistently achieve the highest scores across most metrics. This superiority reflects the importance of high-resolution imagery in tasks involving a large number of classes, where greater image detail likely aids in capturing clear boundaries and distinguishing textures essential for accurate segmentation.

The effectiveness of RemoteCLIP -with the best mIOU (66.22\%), m-f1 (76.41\%), m-precision (74.55\%), and m-accuracy (94.37\%)- also underscores the value of integrating textual information, as it appears particularly beneficial for tasks requiring nuanced class differentiation (see MADOS as well). Models like RemoteCLIP that leverage textual embeddings perform exceptionally well, suggesting that text-based features provide enhanced semantic context. This capability likely helps the model to generalize more effectively across the large number of classes, where textual information augments visual input and reinforces finer distinctions that may not be easily identifiable in pixel-based imagery alone. This is true especially, when the distinctions among classes cannot be based on spectral information.

In contrast, GFMs pre-trained primarily on lower-resolution data perform poorly, as evidenced by models like SpectralGPT and Prithvi, which score significantly lower in nearly all metrics. SpectralGPT, with an mIOU of 32.08\% and m-f1 of 40.47\%, demonstrates the limitations of a low-resolution multi-spectral pre-training approach when the dataset has restricted spectral diversity, such as the FBPs dataset with only four bands. 

UNet still demonstrates respectable results with an mIOU of 58.25\% and an accuracy of 93.38\%. This suggests that, although specialized models with high-resolution or textual pre-training perform best, the baseline UNet architecture can reasonably capture fundamental spatial structures. However, its performance does not match models benefitting from additional pre-training on high-resolution imagery and textual data, which provide a distinct advantage for complex classification tasks. Overall, these results reinforce that models trained on high-resolution imagery and enhanced by textual information are particularly suited to tasks involving numerous classes and poor spectral information, whereas low-resolution pre-training limits segmentation effectiveness due to insufficient detail and limited spectral diversity.

\input{tabs/extensive_fbps}

\subsection{DynamicEarthNet}
DynamicEarthNet dataset is known for its daily observations compared to most datasets with irregular and long revisit intervals. The sequences are acquired from January 2018 to December 2019 with imagery from Planet Labs, covering 75 areas of interest distributed over the globe. Four bands (blue, green, red, and near-infrared) are provided with a ground sample distance of 3 meters and a resolution of $1024 \times 1024$. Pixel-wise semantic segmentation labels are available for the first day of every month, which annotate 6 land use and land cover classes including impervious surface, agriculture, forest, other vegetation, wetlands, soil, water (snow \& ice are annotated but not used due to its extreme scarcity). To reduce the computational requirements for our benchmark, we only sample the images from the first 6 dates in every month as the input image sequences. Example images are shown in ~\autoref{fig:dyn-examples}. Note that, DynamicEarthNet is a challenging dataset with a large distribution shift from train set to validation set and test set.
\input{figs_tex/dyn_sample}

\subsubsection{Extensive Results}
\input{tabs/extensive_dynamicen} 
\cref{tab:extensive-dyn-100perc} demonstrates the extensive results on the DynamicEarthNet Dataset using either linear or L-TAE temporal feature fusion. In general, fusing temporal features with a simple linear layer results in better performance than L-TAE on DynamicEarthNet. Given daily observations with extremely small changes, a simple linear fusion can be more straightforward to capture useful information. Another hypothesis is that linear fusion can better preserve the generalizable representations from GFMs while a heavier L-TAE time strategy may cost more severe overfitting since its additional parameters require more data to be trained properly.

\cref{tab:ext-dyn-linear,tab:ext-dyn-ltae} present the performance of foundation models given a limited portion of training data. When using the full dataset, the best-performing foundation models include CROMA, DOFA and SpectralGPT. These outstanding models are pre-trained on multi-modal data, which is likely the origin of the robustness against the large distribution shift from training to testing. However, Prithvi is an exception as it fails to cope with missing bands during testing time. UNet and ViT Baselines are also competitive on DynamicEarthNet. When using less data, the performance of fully-supervised UNet drops significantly while other models with pre-trained weights still mostly maintain the performance. We notice that using less training data might even lead to increasing performance, such as Prithvi, RemoteCLIP and ViT baseline. For these models, shorter training outweighs using more training data in DynamicEarthNet because of less overfitting. 

\subsection{Crop Type Mapping-South Sudan} 
We include a crop-type semantic segmentation dataset in Africa, specifically South Sudan, to expand the benchmark dataset for underrepresented regions in agricultural areas. The Crop Type Mapping \cite{rustowicz2019semantic} dataset for South Sudan consists of Sentinel-1, Sentinel-2, and Planet satellite imagery in 2017, covering 837 fields and 4 types of crops. The temporal resolution of each scene is variable, with frames that are not temporally aligned. The area of interest is subdivided into $64 \times 64$ pixel grids. For our experiments, we use data from Sentinel-1 and Sentinel-2, which provide 10 spectral bands (B2-B8, B8A, B11, B12) and 2 polarimetric bands (VV, VH), respectively. An example sample is shown in~\autoref{fig:ctm-examples}.
\input{figs_tex/ctm_sample}

\subsubsection{Extensive Results}
\input{tabs/extensive_croptypemap}

\cref{tab:extensive-ctm-100perc} presents the extensive results of the Crop Type Mapping-SS Dataset, with both linear and L-TAE mapping strategies used. Notably, the S12 models, pre-trained on the full spectral bands of Sentinel-2 data, demonstrate top performance. This highlights the importance of leveraging the rich spectral information captured by Sentinel-2, which is particularly crucial for discerning subtle differences between crop types. Alongside the S12 models, RemoteCLIP and DOFA also achieve strong performance. In general, the L-TAE strategy generally yields superior results than linear temporal mapping, except that SpectralGPT exhibits a significant performance drop with this strategy. Regarding the multi-modality experiments, incorporating SAR data does not improve performance and even leads to a decrease compared to using optical data alone. In addition, Scale-MAE, pre-trained on high-resolution imagery, shows the lowest performance on this dataset.

\cref{tab:ext-ctm-linear,tab:ext-ctm-ltae} showcase the data scarcity scenarios for linear and L-TAE strategies, respectively, utilizing only optical information. In these cases, the performance of the fully supervised baseline drops considerably compared to foundation models.  Specifically, CROMA, SatlasNet, and the S12 models significantly outperform the baseline, demonstrating their robustness in limited data settings. This highlights the advantage of foundation models in leveraging learned representations to achieve strong performance even with reduced training data. For the linear mapping strategy, SpectralGPT consistently performs well across different label scarcity settings, further emphasizing the importance of spectral information for accurate crop type mapping.

\subsection{SpaceNet 7}

The SpaceNet 7 dataset features satellite image time series acquired by the PlanetScope constellation between 2017 and 2020 for 80 sites spread across the globe  \cite{van2021multi}. Each time series consists of 18 to 26 monthly image mosaics with a size of $1024 \times 1024$ pixels and a spatial resolution of approximately 4 m. In addition, the SpaceNet 7 dataset provides manually annotated building footprints for the 60 training sites and corresponding masks denoting areas that are obscured by clouds. Although the task of the original SpaceNet 7 challenge was to track these building footprints (i.e. vector data), the SpaceNet 7 dataset is commonly used to evaluate urban change detection methods. A data sample is shown in~\cref{fig:sn7-examples}.

The labeled SpaceNet 7 sites ($n=60$) were first assigned to the training ($n=30$), validation ($n=15$), or test ($n=15$) set, following the split in \cite{hafner2022urban}.
Then, sites were tiled into crops of size $256 \times 256$ pixels. During training, samples for a crop are generated by randomly selecting two timestamps from the image time series with a minimal temporal gap of five timestamps. The rasterized building labels (see \cite{hafner2022urban}) for these timestamps were used to compute the change label. For model evaluation, the first and the last cloud-free images of a time series were selected.
\input{figs_tex/sn7_ai4farms_sample}

\subsubsection{Extensive Results}
\cref{tab:sn7_change_strategy} lists extensive change detection results on the SpaceNet 7 dataset. We compare two strategies, namely concatenation and differencing, to detect changes from the bi-temporal feature pairs extracted by the Siamese encoder (i.e. shared encoders). More specifically, the former strategy uses the concatenated feature pairs as input to the decoder, while the latter strategy uses the subtraction of the feature pairs. Scale-MAE achieved the highest values across all metrics and both change detection strategies, followed by the UNet baseline (trained from scratch) for the concatenation strategy and DOFA for the differencing strategy. Although falling short of the top two performances, SatlasNet and GFM also outperformed the rest of their competitors, achieving IoU values exceeding 20 for both strategies. The change detection strategy only had a minor effect on performance, except for UNet which achieved top-performance using concatenation and lowest performance among the models using differencing. In general, our change detection strategy results indicate that the preferable strategy is model-specific.

\input{tabs/extensive_sn7_change_strategy}

\cref{tab:sn7_limited_label} compares the change detection results on the SpaceNet 7 dataset obtained using the concatenation strategy and 100 \% of the labeled training data to limited label scenarios, using only 50 \% and 10 \% of the labeled training data. In the 50 \% scenario, several foundation models (DOFA, Prithvi, and RemoteCLIP), as well as the UNet baseline, experienced large performance drops (> 5 mIoU) compared to the 100 \% scenario. In comparison, other foundation models successfully retained most of their performance. For example, the performance of GFM, SatlasNet, and Scale-MAE dropped by less than 2 mIoU points. However, none of the models managed to retain their performance from the 100 \% scenario when reducing the labeled training data to 10 \%.

\input{tabs/extensive_sn7_limited_label}

Overall, we attribute the strong performance of Scale-MAE on the SpaceNet 7 change detection task to its robust multi-scale representations, especially since the satellite imagery in SpaceNet 7 (Planet) has a higher spatial resolution (approx. 4 m) than the Sentinel imagery typically used for pertaining foundation models. Furthermore, many other foundation models do not specifically address scale-specific information in satellite images \cite{reed2023scalemae}. It is also noteworthy that most foundation models, including Scale-MAE, were not specifically designed to solve change detection tasks.

\subsection{AI4SmallFarms}
The AI4SmallFarms dataset consists of 439{,}001 manually annotated agricultural field polygons distributed across 62 non-overlapping tiles, each approximately $5 \times 5$ km, in Vietnam and Cambodia. The dataset includes multi-temporal Sentinel-2 image composites, created by aggregating monthly cloud-free images using median pixel values. The Sentinel-2 bands B2, B3, B4 and B8 are included for each tile. Additionally, high-resolution Google Map (GM) images are incorporated to improve boundary accuracy; however, these are not included in training the models, as the goal is to be able to perform farm boundary delineation from Sentinel-2 imagery. The digitization process involved detailed manual annotation from both Sentinel-2 and GM images, followed by a rigorous multi-step quality control procedure that included topological consistency checks and independent verification by multiple quality control teams. ~\cref{fig:ai4sf_example} displays an example sample.

\subsubsection{Extensive Results}

The results on the AI4SmallFarms dataset, presented in \cref{tab:ai4smallfarms}, show that the fully supervised UNet model significantly outperforms all evaluated GFMs. The UNet achieves an mIoU of 46.34\% and a mean F1 score of 62.36\%, while the GFMs exhibit mIoU values ranging from 21.47\% to 27.19\% and m-f1 scores between 35.07\% and 42.73\%.
\input{tabs/extensive_ai4smallfarms}
This performance gap suggests that being able to train the encoder is beneficial for capturing the fine-grained spatial details required for accurate field boundary segmentation in medium-resolution optical imagery. In contrast, the GFMs may lack the necessary spatial resolution or fail to effectively generalize the specific characteristics of smallholder farms in Vietnam and Cambodia. These findings highlight the importance of task-specific model design and suggest that, for certain specialized applications, fully supervised models may offer superior performance over more generalized pre-trained models.





\subsection{BioMassters}
BioMassters is an Above Ground forest Biomass (AGB) estimation dataset at 10m spatial resolution, covering 8.5 million hectares of Finland's boreal forests. The dataset consists of 13,000 reference AGB tiles, each of size $256\times 256$ pixels, along with corresponding time series images from Sentinel-1 and Sentinel-2 satellites. The reference AGB data is derived from Airborne LiDAR campaigns conducted between 2017 and 2021 by the Finland Forest Centre and the National Land Survey (NLS) of Finland. The input data includes 12-month time series of Sentinel-1 and Sentinel-2. The Sentinel-1 input data contains both ascending and descending orbit signals, resulting in four bands in each image: ASC VV, ASC VH, DSC VV, and DSC VH. The Sentinel-2 images contain 10 bands, B2-B8, B8a, B11, and B12. The training and validation set is created using 80-20 split. The test set is created by randomly selecting 30 percent data points from 2018, 2020, and 2021, and 20 percent from 2019. \cref{AGB_sample} shows a data sample. However, it is important to consider that the dataset is localized only in Finland.
\input{figs_tex/biomassters_sample}

\subsubsection{Extensive Results}
The detailed results on BioMassters datasets are presented in \cref{tab:extensive-biomass}. The evaluation is conducted under different temporal strategies, data modalities and limited label strategies. 

\input{tabs/extensive_biomass}

\cref{tab:extensive-biomass} contains evaluation results on all GFMs with pre-trained encoder on representation learning and decoder fine-tuned to estimate forest biomass as a regression task. The results show that the biomass estimation is more accurate with multi-temporal data than uni-temporal data. The estimation is further improved with the L-TAE multi-temporal feature learning strategy. This observation is consistent for all listed GFMs. In terms of modality, we see that the task is better learned through optical modality. In fact, in most of the cases, models solely trained with optical data achieved better results than combined SAR and optical data. We tested GFMs' with limited labels to replicate real world application scenarios, where label scarcity is often an issue. The obtained results show that training with 100\% labels provides better results than training with 50\% which is better than training with 10\% dataset, reflecting the general trend of getting better results with more training labels. However, the results also indicate that using 50\% labels might be sufficient as training with 50\% labels provides almost similar estimations as training with 100\% labels. The best results are obtained from SpectralGPT and CROMA (optical) GFM whereas the worst results are from RemoteCLIP.

We compared the GFM results with two basic supervised baselines; U-Net and ViT. Although not by a big margin but the supervised U-Net baselines outperformed all GFMs, raising questions on the generalizability of GFM for biomass estimation tasks. 


\section{Comparison with existing protocols}

\cref{tab:datasets_per_model} lists all the datasets used to evaluate the surveyed GFMs in their original publication.
Every GFM uses its own protocol on a specific set of tasks, with few overlaps.
Change detection and regression are barely evaluated.

\input{tabs/datasets_per_model}

\section{Additional results}

In this section of the Appendix, we present additional results that explore patterns between model performance and various characteristics of both pre-training and downstream datasets. Specifically, we investigate how performance varies across domains, geographic regions, and sensor resolutions, providing further insight into the factors that influence the generalization ability of geospatial foundation models.

\input{figs_tex/viz1}

\input{figs_tex/viz2}

\input{figs_tex/viz3}

\end{document}

%% file: figs_tex/teaser.tex
\begin{figure*}[!ht]
    \centering
    \includegraphics[width=1.0\linewidth]{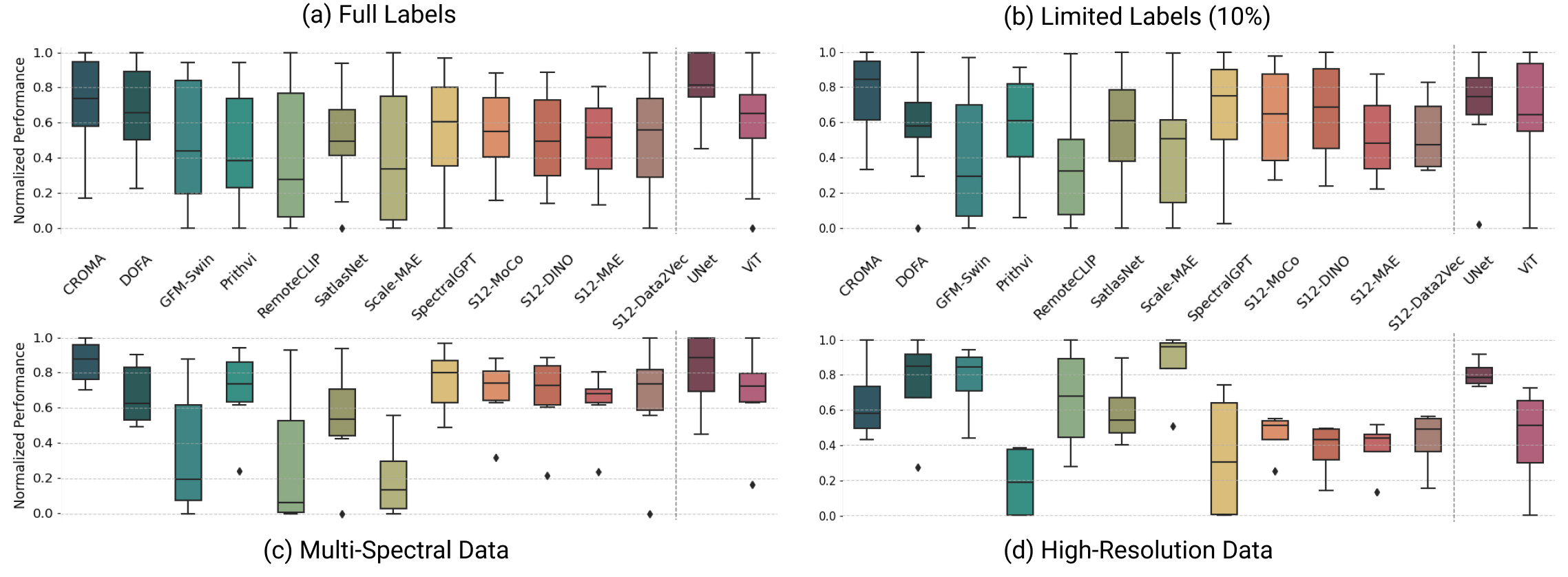}
    \caption{\textbf{Normalized performance comparison of different models across various datasets and training conditions.} The y-axis represents the {normalized performance} across the 11 PANGAEA's datasets, where the best-performing model for each dataset is assigned a value of 1 and the worst-performing model is assigned a value of 0. 
    (a) \textbf{Full Labels}: Models trained on downstream tasks with access to the full labeled dataset. We observe that supervised baselines -- especially UNet -- outperform most of the models; 
    (b) \textbf{Limited Labels (10\%)}: Models trained on downstream tasks with only 10\% of labeled data. In this scenario, some GFMs -- \eg CROMA -- excel and outperform both other GFMs and supervised baselines, although still competitive; 
    (c) \textbf{Multi-Spectral Data}: Results for datasets made exclusively of multi-spectral images, \ie HLS Burn Scars, MADOS, PASTIS-R, Sen1Floods11, Crop Type Mapping-SS, and BioMassters. Multi-spectral datasets tend to use low-resolution images, on which UNet excels. GFMs trained on high-resolution images, \eg Scale-MAE, underperform in this setting that requires both spatial and spectral information;
    (d) \textbf{High-Resolution Data}: Results for high-resolution datasets, \ie xView2, FiveBillionPixels, DynamicEarthNet, and SpaceNet 7. They are made of either RGB or RGB-NIR imagery. In this setting, most GFMs pre-trained on lower-resolution imagery fall behind, except CROMA and DOFA. On the other hand, GFMs pre-trained on high-resolution images and UNet perform well.}
    \label{fig:teaser}
\end{figure*}

%% file: figs_tex/pangaea_fig.tex
\begin{figure*}[htbp]
    \vspace{-3pt}
    \centering
    \includegraphics[width=1.0\linewidth]{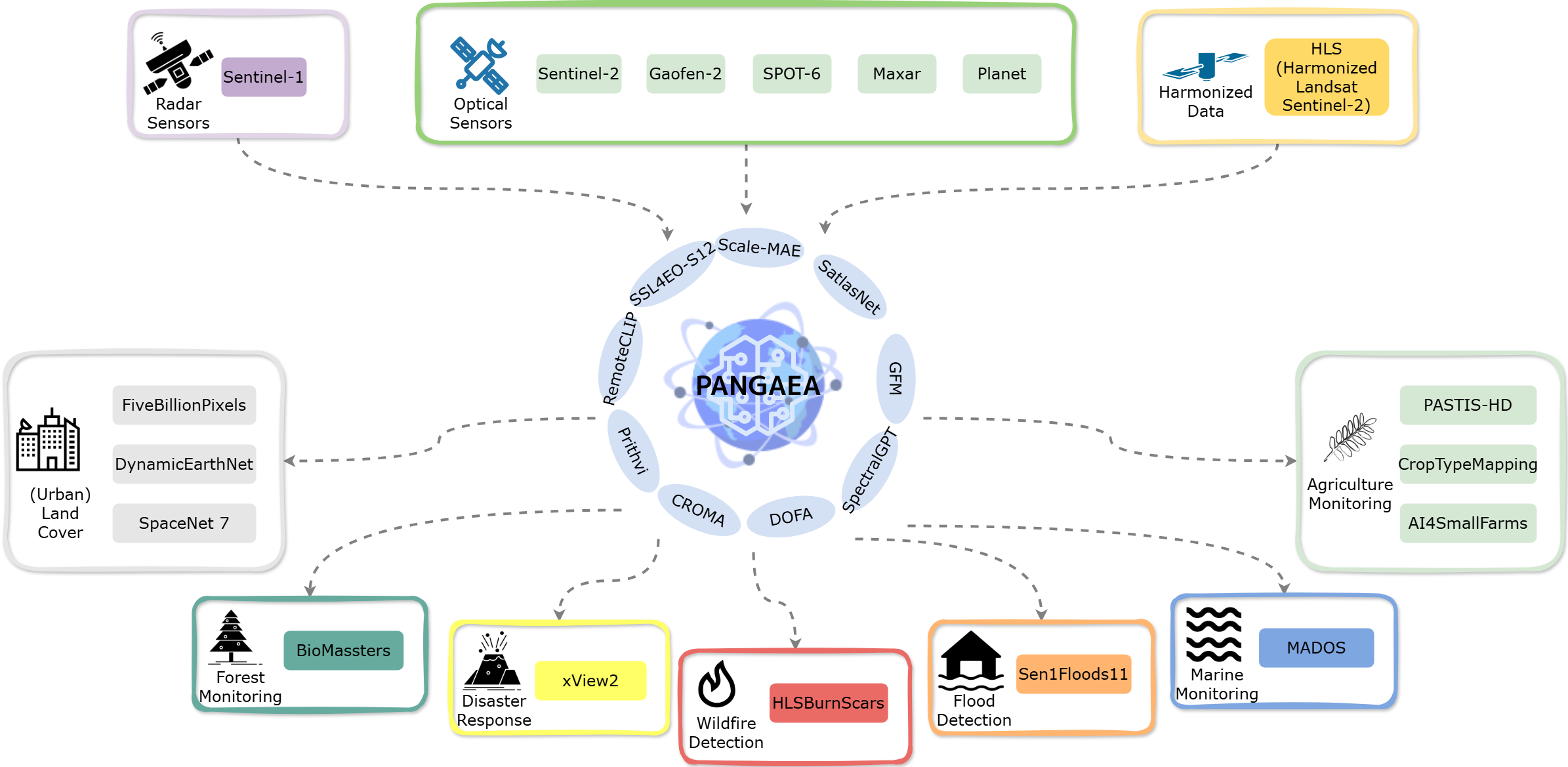}
    \caption{PANGAEA aims for robust evaluation across diverse downstream datasets and applications.}
    \vspace{-8pt}
    \label{fig:pangaea}
\end{figure*}

%% file: tabs/benchmark_comparison.tex

\begin{table*}[htbp]
\centering
\caption{Comparison of the PANGAEA Benchmark with existing geospatial vision benchmarks. We categorized each benchmark by domain (LULC stands for Land Use and Land Cover; SDG for Sustainable Development Goals; Agri for Agricultural), tasks (classification, semantic segmentation, object detection, and regression), modality and temporality.}
\adjustbox{width=0.95\textwidth}{
\begin{tabularx}{\textwidth}{XXlXX} 
\toprule
Benchmark    & Domain             & Tasks               & Modality        & Temporality \\ \midrule
EarthNets    & LULC               & Class, SemSeg, OD   & MSI             & Single      \\
SustainBench & SDG                & Class, SemSeg, Regr & MSI, StreetView & Multi       \\
GEO-Bench    & Urban, Agri, Forest & Class, SemSeg       & MSI, SAR, DSM             & Single      \\
PhilEO Bench & Urban              & SemSeg              & MSI             & Single      \\
FoMO Bench   & Forest             & Class, SemSeg, OD   & MSI             & Multi       \\ \midrule
\textbf{PANGAEA}      & \begin{tabular}[c]{@{}l@{}}Urban, Agri, Forest\\Marine, Disaster\end{tabular} & SemSeg, CD, Regr    & MSI, SAR        & Multi       \\ \bottomrule
\end{tabularx}
}
\label{tab:bench-comp}
\end{table*}

%% file: tabs/datasets_composition.tex
\begin{table*}[htbp]
\centering
\caption{Datasets included in the PANGAEA benchmark, highlighting their domains, tasks, sensor types, geographic coverage and temporality. This selection ensures diverse evaluation of GFMs across a wide range of applications.}
\label{tab:datasets-selection}
\setlength\tabcolsep{4pt}
\begin{tabularx}{\textwidth}{llXXXl}

\toprule
Dataset & Domain & Task & Sensors & Location & Dates\\
\midrule
HLS Burn Scars \cite{jakubik2023foundation} &
  Wildfire &
  Segmentation &
  HLS &
  USA &
  Mono \\
MADOS    \cite{KIKAKI202439}      & Marine             & Segmentation & S2             & Global          & Mono \\
PASTIS-R     \cite{garnot2021mmfusion, garnot2021panoptic}     & Agriculture        & Segmentation & S1, S2, SPOT-6 & France    & Multi       \\
Sen1Floods11    \cite{rambour2020flood}   & Flood              & Segmentation & S1, S2         & Global       & Mono    \\
xView2      \cite{gupta2019xbd, Gupta_2019_CVPR_Workshops}       & HADR               & Change det.      & Maxar          & Global  & Bi         \\
Five Billion Pixels \cite{5bpixels} & (Urban) Land Cover & Segmentation & Gaofen-2       & China     & Mono       \\
DynamicEarthNet   \cite{toker2022dynamicearthnet}  & (Urban) Land Cover & Segmentation & PlanetFusion   & Global    & Mono       \\
Crop Type Mapping-SS  \cite{yeh2021sustainbench, rustowicz2019semantic}   & Agriculture        & Segmentation & S1, S2, Planet & South Sudan   & Multi   \\
SpaceNet 7 \cite{vanetten2019spacenet} &
  Urban &
  Change det. &
  Planet &
  Global & 
  Bi \\
AI4SmallFarms    \cite{ai4smallfarms}   & Agriculture        & Segmentation & S2             & Southeast Asia & Mono\\
BioMassters     \cite{nascetti2023biomassters}    & Forest             & Regression            & S1, S2         & Finland  & Multi        \\
    \bottomrule
\end{tabularx}
\end{table*}

%% file: figs_tex/dataset_dist.tex
\begin{figure*}[htbp]
    \centering
    \includegraphics[width=1.0\linewidth]{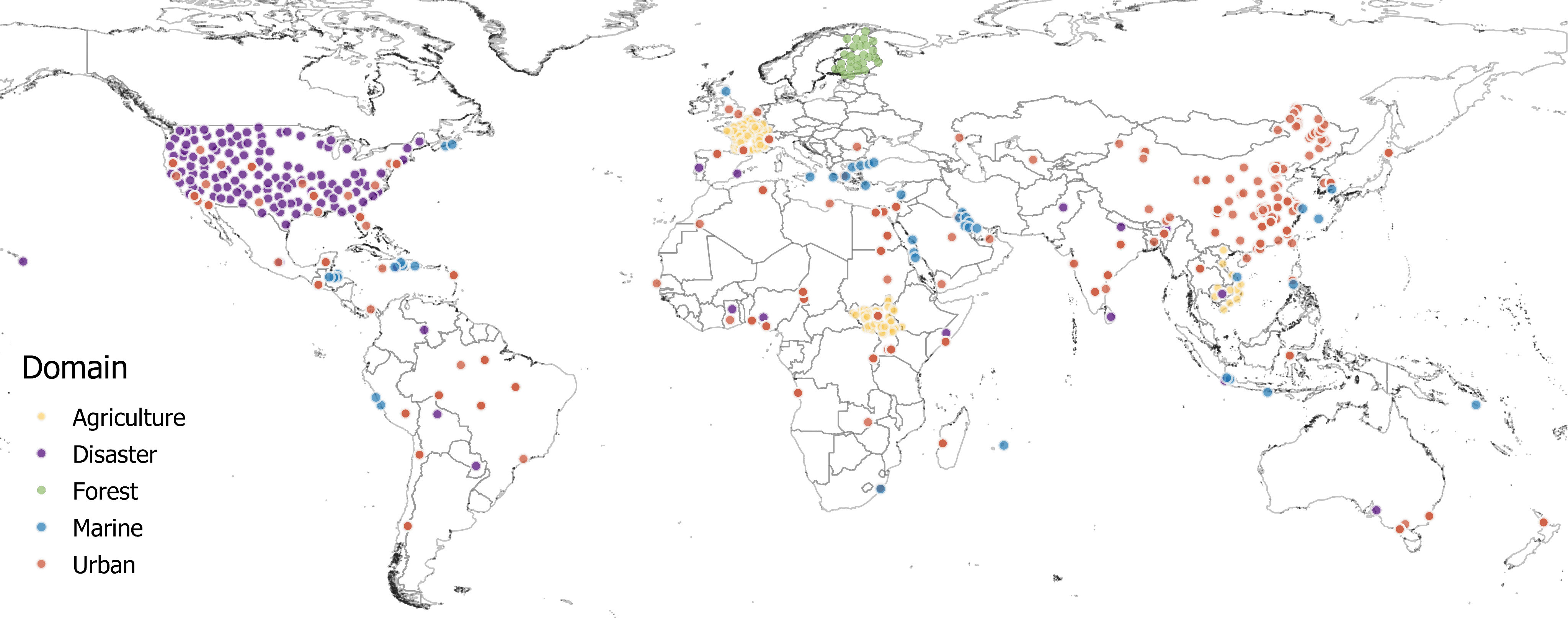}
    \caption{Geographical distribution of PANGAEA benchmark dataset across different domains.}
    \label{fig:datasets-dist}
\end{figure*}

%% file: tabs/model_pretraining_dataset.tex
\begin{table*}[ht]
\centering
\caption{Overview of the pretraining datasets and the number of patches used by the selected GFMs. For Prithvi, the data volume is reported.}
\label{tab:pretraining-datasets}
\setlength\tabcolsep{16pt}
\begin{tabularx}{\textwidth}{lXl}
\toprule
Model &
  Pretraining Images &
  Patches/Volume 
  \\
\midrule
CROMA            & Sentinel-1, Sentinel-2   & 3M  \\ 
DOFA            & Sentinel-1, Sentinel2, Gaofen-2, NAIP, EnMAP  & 8.08M \\
GFM-Swin        & NAIP, RSD46-WHU, MLRSNet, RESISC45, PatternNet & 600K \\
Prithvi         & Harmonized Landsat Sentinel-2 (HLS) & 1TB \\
RemoteCLIP      & SEG-4, DET-10, RET-3 & 165K \\
SatlasNet       & Sentinel-2, NAIP &  856K \\ 
Scale-MAE        & FMoW-RGB & 363.6K \\
SpectralGPT     & fMoW-S2, BigEarthNet  & 1.47M \\
SSL4EO-S12      & Sentinel-1, Sentinel-2   & 3M  \\ 
\bottomrule
\end{tabularx}

\end{table*}

%% file: tabs/model_lists.tex
\begin{table*}[ht]
\centering
\caption{Complexity of the selected GFMs. Please consider that all models include a UPerNet~\cite{xiao2018unified} decoder, except UNet~\cite{ronneberger2015u} which uses its original decoder.}
\label{tab:flops}
\setlength\tabcolsep{8pt}
\begin{tabularx}{\textwidth}{p{0.5cm}lYYY}
\toprule
\multicolumn{2}{c}{Model}            & Complexity (GMacs)   & Trainable Params (M) & Total Params (M) \\
\midrule
\multirow{11}{*}{\rotatebox[origin=c]{90}{Foundation models}} & CROMA            & 139.23  & 46.95     &  350.0   \\
& DOFA             &   65.02      &     39.35            &   150.9  \\
& GFM-Swin         & 19.05   & 33.71           &  120.4   \\
& Prithvi          & 64.85  & 39.34           &  125.7   \\
& RemoteCLIP       & 17.57  & 39.35           &  16.8   \\
& SatlasNet        &  16.70 & 33.25           &  121.2  \\ 
& Scale-MAE         &  103.77   &   46.95       &  350.1   \\
& SpectralGPT      & 380.69 & 164.40           &  249.8   \\
& S12-MoCo     & 46.90  & 30.89            &   53.5  \\
& S12-DINO     & 47.93  & 30.89           &  53.5   \\
& S12-MAE      & 46.90  & 30.89           &  53.5   \\
& S12-Data2Vec & 46.89  & 30.89           &  53.5   \\
\midrule
\multirow{2}{*}{\rotatebox[origin=c]{90}{\tiny Baselines}} & 
UNet             &   27.38      &     14.79            &   14.79  \\
& ViT-B/16            &    64.74     &       125.1    &  125.1   \\
\bottomrule
\end{tabularx}
\end{table*}

%% file: figs_tex/temporal_strategy.tex
\begin{figure*}[htbp]
    \centering
        \begin{subfigure}[b]{0.31\textwidth}
        \centering
        \includegraphics[width=\textwidth]{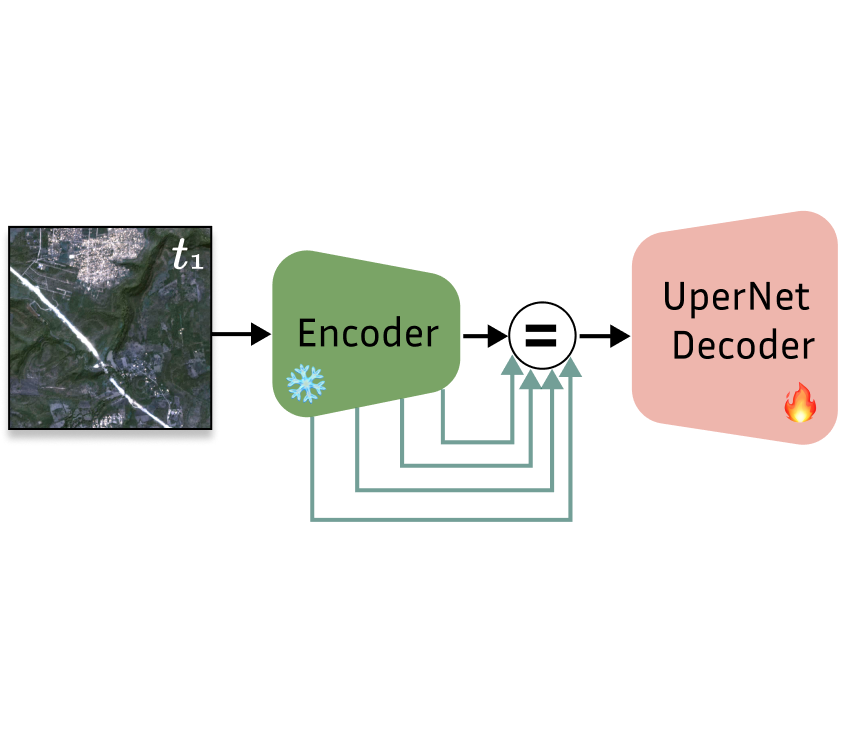} 
        \caption{single temporal}
        \label{fig:uni-temporal}
    \end{subfigure}
    \hfill
    \begin{subfigure}[b]{0.31\textwidth}
        \centering
        \includegraphics[width=\textwidth]{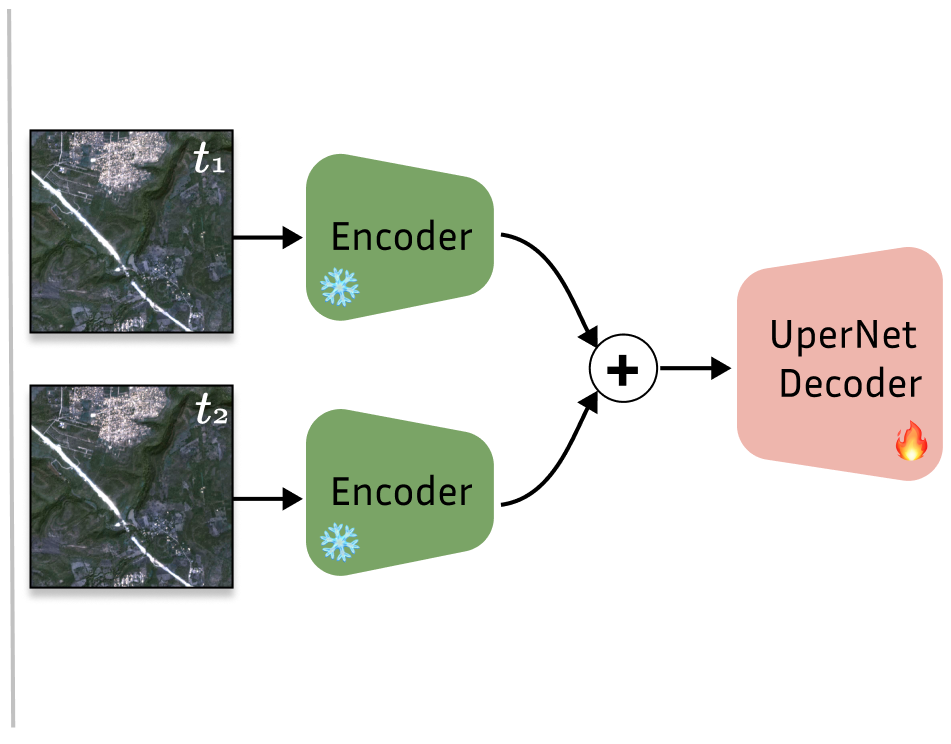} 
        \caption{bi-temporal}
        \label{fig:bi-temporal}
    \end{subfigure}
    \hfill
    \begin{subfigure}[b]{0.36\textwidth}
        \centering
        \includegraphics[width=\textwidth]{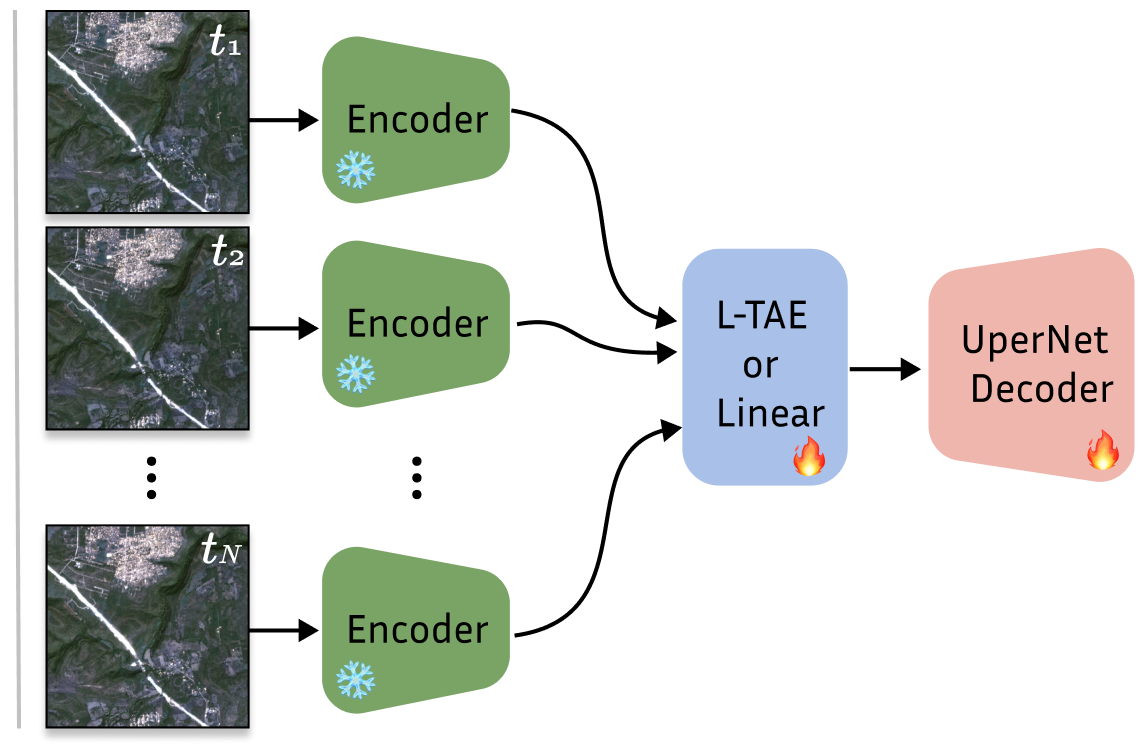} 
        \caption{multi-temporal}
        \label{fig:multi-temporal}
    \end{subfigure}
    \caption{Fine-tuning strategies of the foundation models depending on the input temporality.}
    \label{fig:temporal_strategy}
\end{figure*}

%% file: tabs/main_miou.tex
\begin{table*}[t]
\centering
\caption{Performance evaluation of Geospatial Foundation Models across 11 benchmark datasets using 100\% of the data. For semantic segmentation and change detection tasks, the mIoU $\uparrow$ is reported. For regression task, RMSE $\downarrow$ is reported. \#Top2 indicates the number of datasets where the models achieve top-2 performance across all evaluated datasets.}
\label{tab:comp_100}
\renewcommand{\arraystretch}{1.3} 
\begin{adjustbox}{width=1\textwidth}
\setlength\tabcolsep{2pt}
\begin{tabular}{lccccccccccc | c}
\toprule
Model & HLS Burns & MADOS & PASTIS & Sen1Floods11 & xView2 & FBP & DynEarthNet & CropMap & SN7 & AI4Farms & BioMassters & \#Top2\\
\midrule
CROMA        & 82.42 & \textbf{67.55} & 32.32& \underline{90.89} & 53.27 & 51.83 & 38.29 & 49.38 & 59.28 & 25.65 & 36.81 & 2 \\ 
DOFA         & 80.63 & 59.58 & 30.02& 89.37 & \underline{59.64} & 43.18 & \underline{39.29} & 51.33  & 61.84 & 27.07 & 42.81 & 2 \\ 
GFM-Swin     & 76.90 & \underline{64.71} & 21.24& 72.60 & 59.15 & 67.18 & 34.09 & 46.98 & 60.89 & 27.19 & 46.83 & 1\\ 
Prithvi      & \underline{83.62} & 49.98 & 33.93& 90.37 & 49.35 & 46.81 & 27.86 & 43.07 &  56.54 & 26.86 & 39.99 & 1 \\ 
RemoteCLIP   & 76.59 & 60.00 & 18.23& 74.26 & 57.41 & \textbf{69.19} & 31.78 & \underline{52.05} & 57.76 & 25.12 & 49.79 & 2\\ 
SatlasNet    & 79.96 & 55.86 & 17.51& 90.30 & 52.23 & 50.97 & 36.31 & 46.97  & 61.88 & 25.13 & 41.67 & 0\\ 
Scale-MAE    & 76.68 & 57.32 & 24.55& 74.13 & \textbf{60.72} & \underline{67.19} & 35.11 & 25.42 & \textbf{62.96} & 21.47 & 47.15 & 3\\ 
SpectralGPT  & 80.47 & 57.99 &35.44& 89.07 & 48.40 & 33.42 & 37.85 & 46.95 &  58.86 & 26.75 & \underline{36.11} & 1\\ 
S12-MoCo     & 81.58 & 51.76 & 34.49& 89.26 & 51.59 & 53.02 & 35.44 & 48.58 & 57.64 &  25.38 & 40.21  & 0\\ 
S12-DINO     & 81.72 & 49.37 & \underline{36.18}& 88.61 & 50.56 & 51.15 & 34.81 & 48.66 & 56.47 & 25.62 & 41.23 & 1\\ 
S12-MAE      & 81.91 & 49.90 & 32.03& 87.79 & 50.44 & 51.92 & 34.08 & 45.8 & 57.13 & 24.69 & 41.07 & 0\\ 
S12-Data2Vec & 81.91 & 44.36 & 34.32& 88.15 & 51.36 & 48.82 & 35.90 & \textbf{54.03} & 58.23 & 24.23 & 41.91 & 1\\ 
\midrule 
UNet Baseline  & \textbf{84.51} & 54.79 &  31.60& \textbf{91.42} & 58.68 & 60.47 & \textbf{39.46} & 47.57 &    \underline{62.09}   & \textbf{46.34} & \textbf{35.67}& 6\\
ViT Baseline     &  81.58     & 48.19 &   \textbf{38.53}&87.66   & 57.43  & 59.32  & 36.83  & 44.08  &  52.57 & \underline{38.37} & 38.55& 2\\ 
\bottomrule
\end{tabular}
\end{adjustbox}
\end{table*}

%% file: tabs/50_percent_miou.tex
\begin{table*}[htbp]
\centering
\caption{Performance evaluation of Geospatial Foundation Models across 11 benchmark datasets using 50\% of the data. For semantic segmentation and change detection tasks, the mIoU $\uparrow$ is reported. For regression task, RMSE $\downarrow$ is reported. \#Top2 indicates the number of datasets where the models achieve top-2 performance across all evaluated datasets.}
\label{tab:comp_50}
\renewcommand{\arraystretch}{1.3} 
\begin{adjustbox}{width=1\textwidth}
\setlength\tabcolsep{2pt}
\begin{tabular}{lccccccccccc | c}
\toprule
Model    & HLS Burns & MADOS & PASTIS & Sen1Floods11 & xView2 & FBP & DynEarthNet & CropMap & SN7 & AI4Farms & BioMassters  & \# Top-2\\ \midrule
CROMA              & \underline{81.52} & 57.68 & 32.33& \underline{90.57} & 51.44 & 48.01 & \underline{38.30} & \underline{42.20} & 59.31 & 28.19 & 38.50 &  4\\
DOFA               & 78.02 & 55.21 & 28.60 & 88.39 & \underline{58.91} & 36.90 & \textbf{39.20} &30.93  & 47.06 & 26.69 & 42.81&  2\\
GFM-Swin           & 74.36 & \textbf{63.37} &   20.41& 71.61 & 57.81 & \underline{63.14} & 31.25 & 31.42 & 59.83 & 28.43 & 48.19&  2\\
Prithvi            & 80.89 & 40.79 & 33.13 & 89.69 & 45.79 & 40.27 & 33.43 & \textbf{42.51} & 49.45 & 29.27 & 41.03&  1\\
RemoteCLIP         & 74.28 & 53.26 & 17.46 & 71.67 & 57.43 & \textbf{65.92} & 30.91 & 36.3 & 50.83 & 25.11 & 50.09&  1\\ 
SatlasNet          & 75.97 & 52.24 & 16.78 & 89.45 & 50.74 & 46.04 & 36.34 & 35.29 & \underline{60.74} & 27.08 & 42.23&  1\\
Scale-MAE          & 75.47 & 46.87 & 23.26 & 72.54 & \textbf{59.45} & 62.11 & 32.60 &20.32  & \textbf{61.24} & 26.40 & 46.74&  2\\
SpectralGPT        & 76.40 & \underline{58.00} & 34.61 & 87.52 & 45.94  & 21.71 & 36.52 & 32.09 & 56.28 & 27.46 & \underline{37.34}&  2\\
S12-MoCo           & 79.79 & 42.90 & 32.59 & 89.22 & 49.66 & 46.92 & 34.45 & 41.32  & 56.21 & 28.38 & 41.08&  0\\
S12-DINO           & 80.12 & 40.42 & \textbf{35.71} & 88.93 & 48.46 & 44.85 & 32.76 & 31.13 & 55.14 & 25.68 & 41.47&  1\\
S12-MAE            & 80.13 & 44.29 & 31.15 & 88.43 & 47.09 & 45.63 & 33.29 & 28.07 & 55.55 & 27.50 & 41.66&  0\\
S12-Data2Vec       & 79.82 & 41.22 & 33.42 & 86.58 & 48.84 & 46.73 & 32.61 & 28.53 & 56.94 & 25.84 & 42.82&  0\\
\midrule
UNet Baseline      &   \textbf{82.39}   & 43.87 &  30.25& \textbf{90.91} & 56.58 & 55.42 & 35.14 & 36.30 & 46.82 & \textbf{45.02} & \textbf{36.72}&  4\\
ViT Baseline       &   78.17    & 28.77 &   \textbf{38.71}& 86.08  & 54.82  & 57.32  &  37.33  & 39.53 &  49.21 & \underline{38.37} & 39.56&  2\\ 
\bottomrule
\end{tabular}
\end{adjustbox}
\label{tab:com-50}
\end{table*}

%% file: tabs/10_percent_miou.tex
\begin{table*}[htbp]
\centering
\caption{Performance evaluation of Geospatial Foundation Models across 11 benchmark datasets using 10\% of the data. For semantic segmentation and change detection tasks, the mIoU $\uparrow$ is reported. For regression task, RMSE $\downarrow$ is reported. \#Top2 indicates the number of datasets where the models achieve top-2 performance across all evaluated datasets.}
\label{tab:comp_10}
\renewcommand{\arraystretch}{1.3} 
\begin{adjustbox}{width=1\textwidth}
\setlength{\tabcolsep}{2pt}
\begin{tabular}{lccccccccccc |c}
\toprule
Model & HLS Burns & MADOS & PASTIS & Sen1Floods11 & xView2 & FBP & DynEarthNet & CropMap & SN7 & AI4Farms & BioMassters & \#Top2 \\ 
\midrule 
CROMA              & 76.44 & \textbf{32.44} & 32.80 & \underline{87.22} & 46.54 & 37.39 & \underline{36.08} & \underline{36.77} & 42.15 & \underline{38.48} & \underline{40.25} &6\\
DOFA               & 71.98 & 23.77 & 27.68 & 82.84 & \underline{55.60} & 27.82 & \textbf{39.15} & 29.91 & 46.10 & 27.74 & 46.03  &2\\
GFM-Swin           & 67.23 & 28.19 &   21.47& 62.57 & 53.45 & 55.58 & 28.16 & 27.21 & 39.48 & 32.88 & 49.30 &0\\
Prithvi            & 77.73 & 21.24 & 33.56 & 86.28 & 35.08 & 29.98 & 32.28 & 27.71 & 36.78 & 35.04 & 41.19 &0\\
RemoteCLIP         & 69.40 & 20.57 & 17.19 & 62.22 & 53.75 & \underline{56.23} & 34.43 & 19.86 & 43.11 & 23.85 & 53.32 &1\\
SatlasNet          & 74.79 & \underline{29.87} & 16.76 & 83.92 & 44.07 & 37.86 & 34.64 & 29.08 & \textbf{49.78} & 13.91 & 44.38 &2\\
Scale-MAE          & 75.47 & 21.47 & 22.86 & 64.74 & \textbf{56.06} & 48.75 & 35.27 & 13.44 & \underline{49.68} & 26.66 & 54.16 &2\\
SpectralGPT        & \textbf{83.35} & 20.29 & 34.53 & 83.12 & 35.81 & 39.51 & 35.33 & 31.06 & 36.31 & 37.35 & \textbf{39.44} &2\\
S12-MoCo           & 73.11 & 19.47 & 32.51 & 79.58 & 41.15 & 35.57 & 32.24 & 36.54  & 49.46 & 37.97 & 44.83  &0\\
S12-DINO           & 75.93 & 23.47 & \underline{36.62} & 84.95 & 41.02 & 34.63 & 32.78 & \textbf{38.44} & 41.15 & 37.91 & 42.74  &2\\
S12-MAE            & 76.60 & 18.44 & 31.06 & 84.81 & 39.84 & 35.56 & 30.59 & 35.29 & 40.51 & 23.60 & 43.76 &0 \\
S12-Data2Vec       & 74.38 & 17.86 & 33.09 & 81.91 & 41.60 & 37.27 & 33.63 & 34.11 & 40.66 & 22.85 & 46.52  &0\\
\midrule
UNet Baseline      &   \underline{79.46}    & 24.30 & 29.53& \textbf{88.55} & 46.77 & 52.58 & 35.59 & 13.88 & 46.08 & 34.84 & 40.39 &2\\
ViT Baseline       &   75.92    & 10.18 &   \textbf{38.44}& 81.85  & 44.85  & \textbf{56.53}  & 35.39  &27.76  &  36.01 & \textbf{39.20} & 44.89 &3\\ 
 \bottomrule
\end{tabular}
\end{adjustbox}
\label{tab:com-10}
\end{table*}


%% file: tabs/abl_wo_norm.tex
\begin{table*}[htbp]
    \centering
    \caption{
    Ablation study on the impact of normalization strategies across three benchmark datasets.
    }
    \begin{tabularx}{\textwidth}{lYYrYYrYYr}
\toprule
         & \multicolumn{3}{c}{HLS Burns} & \multicolumn{3}{c}{MADOS} & \multicolumn{3}{c}{SpaceNet 7} \\
         \cmidrule(lr){2-4} \cmidrule(lr){5-7} \cmidrule(lr){8-10}
         Model & w/ & w/o & \% diff & w/ & w/o & \% diff & w/ & w/o & \% diff \\
         \midrule
        CROMA        & 82.42 & 83.76 & \textcolor{ForestGreen}{2\%} & 67.55 & 54.21 & \textcolor{BrickRed}{-20\%} & 59.28 & 58.09 & \textcolor{BrickRed}{-2\%} \\
        DOFA         & 80.63 & 80.47 & 0\% & 59.58 & 51.55 & \textcolor{BrickRed}{-13\%} & 61.84 & 60.41 & \textcolor{BrickRed}{-2\%} \\
        GFM-Swin     & 76.90 & 68.94 & \textcolor{BrickRed}{-10\%} & 64.71 & 50.39 & \textcolor{BrickRed}{-22\%} & 60.89 & 58.87 & \textcolor{BrickRed}{-3\%} \\
        Prithvi      & 83.62 & 81.47 & \textcolor{BrickRed}{-3\%}  & 49.98 & 40.31 & \textcolor{BrickRed}{-19\%} & 56.54 & 56.80 & 0\% \\
        RemoteCLIP   & 76.59 & 66.04 & \textcolor{BrickRed}{-14\%} & 60.00 & 32.82 & \textcolor{BrickRed}{-45\%} & 57.76 & 55.41 & \textcolor{BrickRed}{-4\%} \\
        SatlasNet    & 79.96 & 81.79 & \textcolor{ForestGreen}{2\%}  & 55.86 & 62.66 & \textcolor{ForestGreen}{12\%} & 61.88 & 61.29 & \textcolor{BrickRed}{-1\%} \\
        Scale-MAE    & 76.68 & 61.68 & \textcolor{BrickRed}{-20\%} & 57.32 & 16.81 & \textcolor{BrickRed}{-71\%} & 62.96 & 61.07 & \textcolor{BrickRed}{-3\%} \\
        SpectralGPT  & 80.47 & 80.85 & 0\% & 57.99 & 53.50 & \textcolor{BrickRed}{-8\%}  & 58.86 & 58.06 & \textcolor{BrickRed}{-1\%} \\
        S12-MoCo     & 81.58 & 82.61 & \textcolor{ForestGreen}{1\%}  & 51.76 & 60.18 & \textcolor{ForestGreen}{16\%} & 57.64 & 56.73 & \textcolor{BrickRed}{-2\%} \\
        S12-DINO     & 81.72 & 81.25 & \textcolor{BrickRed}{-1\%}  & 49.37 & 62.31 & \textcolor{ForestGreen}{26\%} & 56.47 & 55.39 & \textcolor{BrickRed}{-2\%} \\
        S12-MAE      & 81.91 & 82.44 & \textcolor{ForestGreen}{1\%}  & 49.90 & 60.48 & \textcolor{ForestGreen}{21\%} & 57.13 & 55.55 & \textcolor{BrickRed}{-3\%} \\
        S12-Data2Vec & 81.91 & 80.96 & \textcolor{BrickRed}{-1\%}  & 44.36 & 51.64 & \textcolor{ForestGreen}{16\%} & 58.23 & 56.53 & \textcolor{BrickRed}{-3\%} \\
        \bottomrule
    \end{tabularx}
    \label{tab:abl-norm}
\end{table*}

%% file: tabs/abl_match_input_resolution.tex
\begin{table*}[htbp]
    \centering
    \caption{
    Ablation study on the effect of downsampling the input resolution to match the model's training resolution on the FiveBillionPixels dataset.
    }
    \begin{tabularx}{0.6\textwidth}{lYYr}
        \toprule
         &  \multicolumn{3}{c}{FiveBillionPixels} \\
         \cmidrule(lr){2-4}
         Model  & w/o matching & w/ matching & \% diff \\
         \midrule
        CROMA & 51.83 & 21.10 & \textcolor{BrickRed}{-59\%} \\
        Prithvi  & 46.81 & 30.99 & \textcolor{BrickRed}{-34\%} \\ 
        S12-DINO & 51.15 & 36.75 & \textcolor{BrickRed}{-28\%} \\ 
        \bottomrule
    \end{tabularx}
    \label{tab:abl_inp_size}
\end{table*}


%% file: tabs/multitemp.tex
\begin{table*}[t]
\centering
\caption{Comparison of different multi-temporal strategies on multi-temporal datasets. L-TAE generally demonstrates an advantage over linear temporal mapping.}
\label{tab:temp}
    \begin{tabularx}{\textwidth}{lYYYYYYYY}
    \toprule
    \multicolumn{1}{c}{}&  \multicolumn{2}{c}{BioMassters}&  \multicolumn{2}{c}{Crop Type Mapping-SS}&  \multicolumn{2}{c}{PASTIS-R}& \multicolumn{2}{c}{DynamicEarthNet}\\
    \cmidrule(lr){2-3} \cmidrule(lr){4-5} \cmidrule(lr){6-7}\cmidrule(lr){8-9}
    Model&  Linear&  L-TAE&  Linear&  L-TAE&  Linear&  L-TAE&  Linear&  L-TAE\\
    \midrule
    CROMA         & 39.99 & 36.81 & 47.02 & 49.38 & 32.06 & 38.51 & 38.25 & 38.29 \\
    DOFA           & 43.25 & 42.81 & 49.81 & 51.33 & 24.49 & 30.02& 40.84 & 39.29 \\
    GFM-Swin     & 47.96 & 46.83 & 39.72 & 46.98 & 15.13 & 21.24 & 35.78 & 34.09  \\
    Prithvi     & 42.90 & 39.99 & 39.92 & 43.07 & 27.40 & 33.93  & 28.42 & 27.86 \\
    RemoteCLIP & 49.60 & 49.79 & 46.50 & 52.05 & 13.31 & 18.23& 34.07 & 31.78 \\
    Scale-MAE    & 52.24 & 47.15 & 21.39 & 25.42 & 11.74 & 24.55& 35.47 & 35.11 \\
    SpectralGPT  & 38.55 & 36.11 & 53.50 & 46.95 & 27.52 & 35.44 & 39.27 & 37.85 \\
    S12-MoCo    & 41.82 & 40.21 & 44.22 & 48.58 & 26.09 & 34.49 & 36.53 & 35.44\\
    S12-DINO  & 41.54 & 41.23 & 46.56 & 48.66 & 32.45 & 36.18& 33.80 & 34.81\\
    S12-MAE  & 42.54 & 41.07 & 46.28 & 45.80 & 27.06 & 32.03& 35.23 & 34.08   \\
    S12-Data2Vec & 42.02 & 41.91 & 54.01 & 54.03 & 26.94 & 34.32 & 36.41 & 35.90 \\ \midrule
    ViT Baseline & 40.94 & 38.55 & 50.24 & 51.78 & 36.57 & 38.53 & 36.83 & 36.38  \\ 
    \bottomrule
    \end{tabularx}
\end{table*}

%% file: tabs/domain_adaptation_fbps.tex
\begin{table*}[htbp]
    \centering
    \caption{
    Domain Adaptation experiments for FiveBillionPixels, comparing model performance on dataset splits without and with regional domain gaps.
    }
    \begin{tabularx}{0.7\textwidth}{lYYr}
        \toprule
         Model & w/o regional domain gap & w/ regional domain gap & \% diff \\
         \midrule
        CROMA & 51.83 & 18.38 & \textcolor{BrickRed}{-65\%} \\
        DOFA & 43.18 & 16.56 & \textcolor{BrickRed}{-62\%} \\
        GFM-Swin & 67.18 & 35.39 & \textcolor{BrickRed}{-47\%} \\
        Prithvi & 46.81 & 16.75 & \textcolor{BrickRed}{-64\%} \\
        RemoteCLIP & 69.19 & 36.53 & \textcolor{BrickRed}{-47\%} \\
        SatlasNet & 50.97 & 21.13 & \textcolor{BrickRed}{-59\%} \\
        Scale-MAE & 67.19 & 34.88 & \textcolor{BrickRed}{-48\%} \\
        SpectralGPT & 33.42 & 18.34 & \textcolor{BrickRed}{-45\%} \\
        S12-MoCo & 53.02 & 17.30 & \textcolor{BrickRed}{-67\%} \\
        S12-DINO & 51.15 & 18.64 & \textcolor{BrickRed}{-64\%} \\
        S12-MAE & 51.92 & 17.78 & \textcolor{BrickRed}{-66\%} \\
        S12-Data2Vec & 48.82 & 18.23 & \textcolor{BrickRed}{-63\%} \\
        \midrule
        UNet Baseline & 60.47 & 30.14 & \textcolor{BrickRed}{-50\%} \\
        ViT Baseline & 59.32 & 22.77 & \textcolor{BrickRed}{-62\%} \\
        \bottomrule
    \end{tabularx}
    \label{tab:dom_ad_fbps}
\end{table*}

%% file: figs_tex/training_losses.tex
\begin{figure*}[htbp]
    \centering
    \begin{subfigure}{0.5\textwidth}
            \centering
            \includegraphics[width=\linewidth]{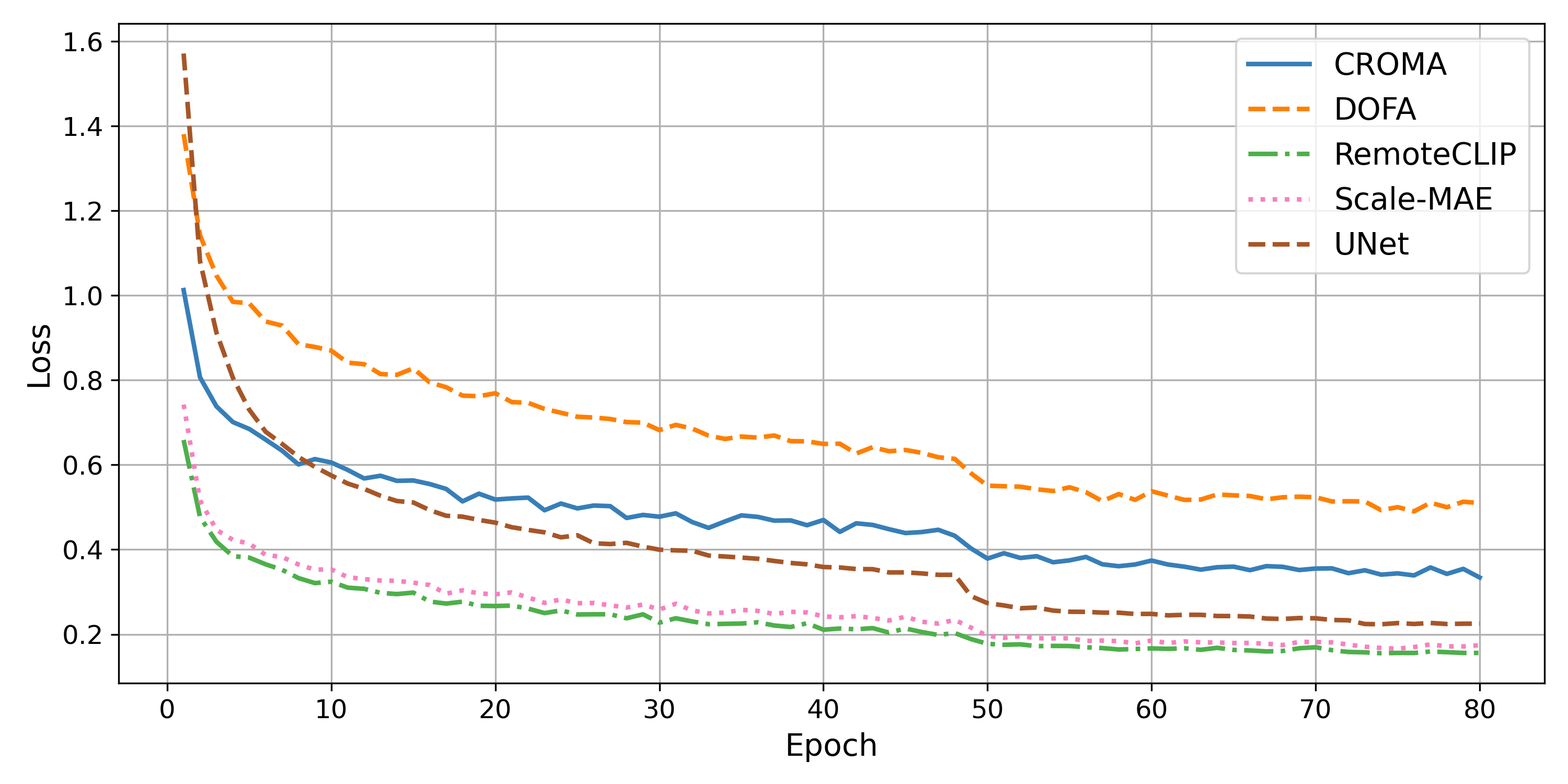}
            \caption*{FiveBillionPixels}
    \end{subfigure}%
    \begin{subfigure}{0.5\textwidth}
            \centering
            \includegraphics[width=\linewidth]{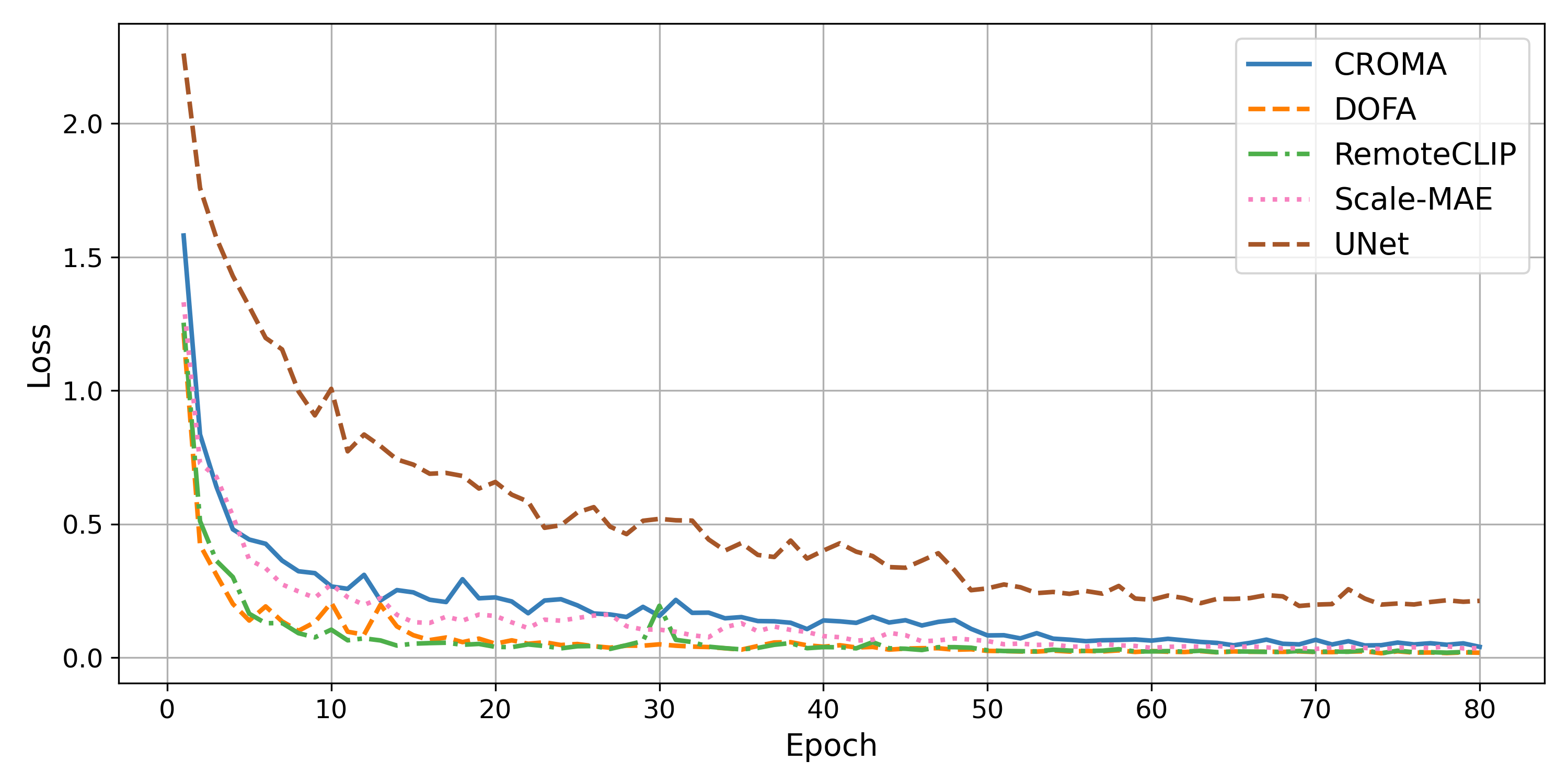}
            \caption*{MADOS}
    \end{subfigure}\\
    \begin{subfigure}{0.5\textwidth}
            \centering
            \includegraphics[width=\linewidth]{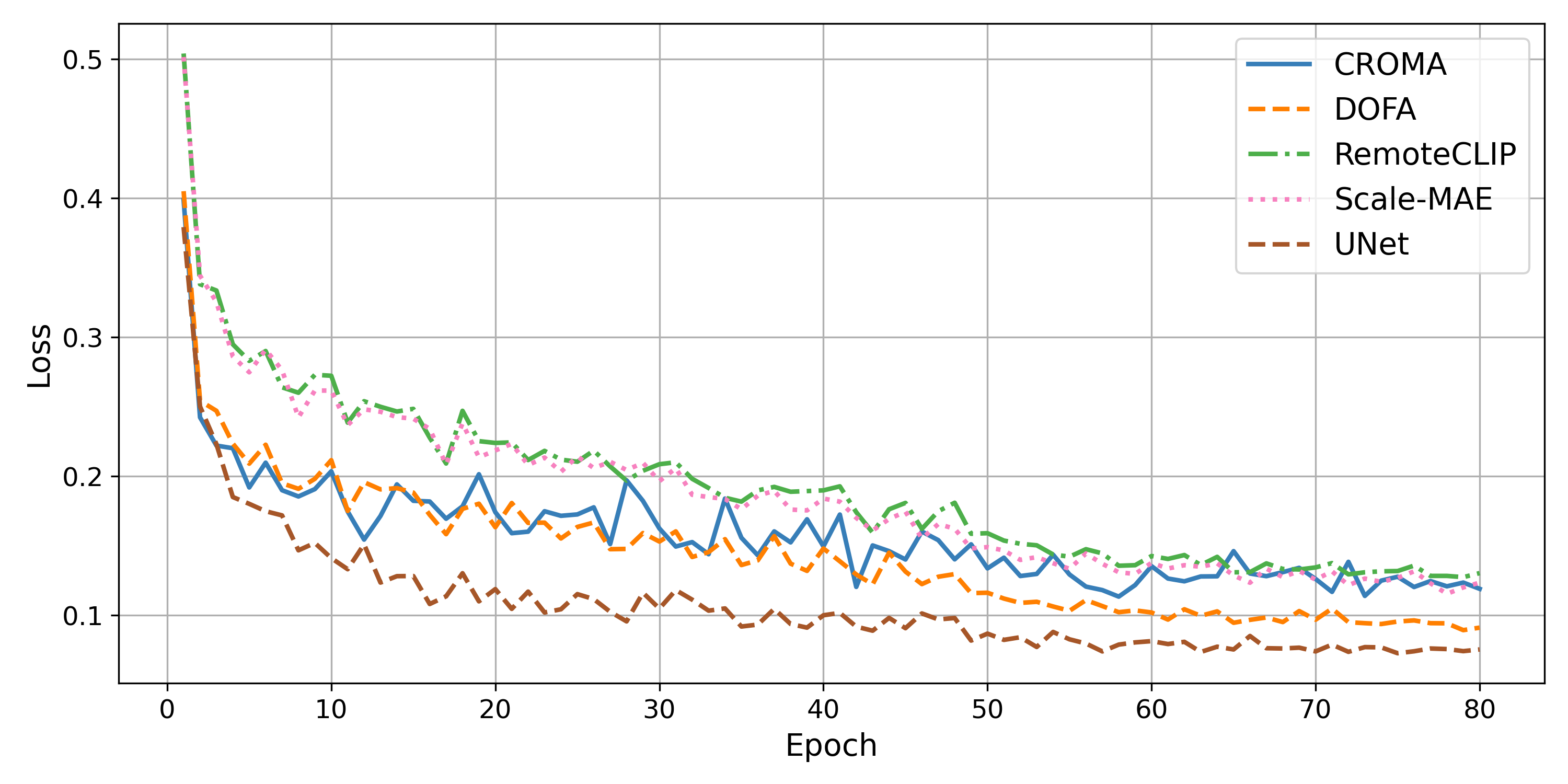}
            \caption*{HLSBurnScars}
    \end{subfigure}%
    \begin{subfigure}{0.5\textwidth}
            \centering
            \includegraphics[width=\linewidth]{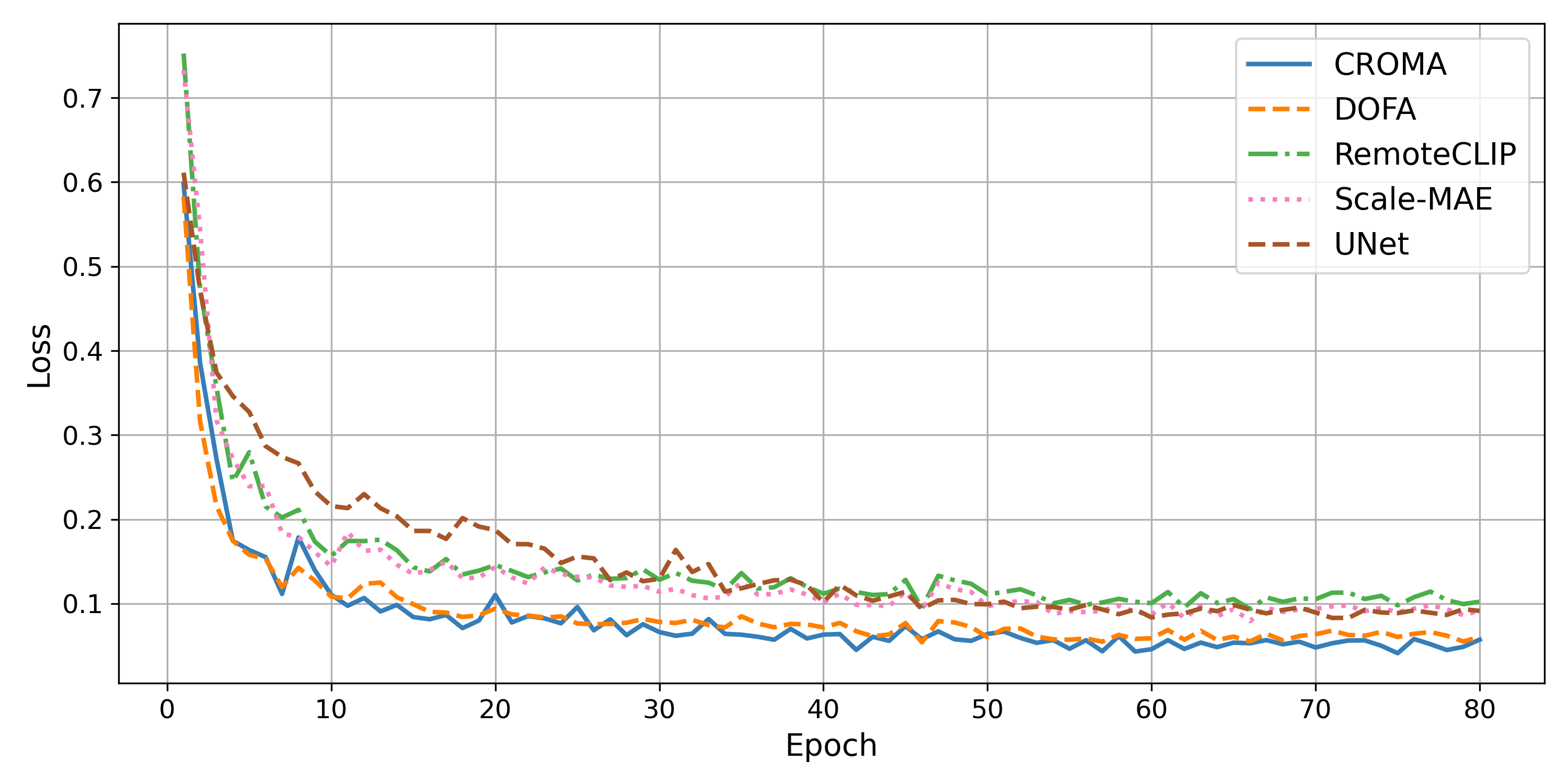}
            \caption*{Sen1Floods11}
    \end{subfigure}%
    \caption{Convergence of training losses for four GFMs -- one pre-trained on high-resolution (Scale-MAE), one on text-image (RemoteCLIP), one on different modalities and resolution (DOFA) and one on low-resolution Sentinel data  --  and a baseline (UNet), on different datasets (\ie different domains and resolutions).}
    \label{fig:training-losses}
\end{figure*}

%% file: tabs/finetuning_ablation.tex
\begin{table*}[htbp]
    \centering
    \caption{Impact of encoder fine-tuning on the performance of the best and worst-performing GFMs for each dataset. The table shows results without (w/o) and with (w/) fine-tuning, along with the relative percentage difference (\% diff). Positive values indicate improvement, while negative values denote performance degradation. Results highlight the variability in fine-tuning effects across different models and datasets.}
    \setlength{\tabcolsep}{8pt}
    \begin{tabularx}{\textwidth}{llYYrlYYr}
    \toprule
        & \multicolumn{4}{c}{Best model} & \multicolumn{4}{c}{Worst model} \\
        \cmidrule(lr){2-5} \cmidrule(lr){6-9}
        Dataset & Model & w/o & w/ & \% diff & Model & w/o & w/ & \% diff \\
        \midrule
        HLS Burns & Prithvi & 83.62 & 85.52 & \textcolor{ForestGreen}{+2\%} & RemoteCLIP & 76.59 & 65.13 & \textcolor{BrickRed}{-15\%} \\
        MADOS & CROMA & 67.55 & 66.98 & \textcolor{BrickRed}{-1\%} & S12-DINO & 49.38 & 56.98 & \textcolor{ForestGreen}{+15\%} \\
        PASTIS & S12-DINO & 36.18 & 44.03 & \textcolor{ForestGreen}{+22\%} & SatlasNet & 17.51 & 29.05 & \textcolor{ForestGreen}{+66\%} \\
        Sen1Floods11 & CROMA & 90.89 & 90.87 & 0\% & GFM-Swin & 72.60 & 75.70 & \textcolor{ForestGreen}{+4\%} \\
        xView2 & ScaleMAE & 60.72 & 62.16 & \textcolor{BrickRed}{+2\%} & Prithvi & 49.35 & 57.33 & \textcolor{ForestGreen}{+16\%} \\
        FBP & RemoteCLIP & 69.19 & 55.58 & \textcolor{BrickRed}{-20\%} & DOFA & 43.82 & 63.74 & \textcolor{ForestGreen}{+45\%} \\
        DynEarthNet & DOFA & 39.29 & 37.76 & \textcolor{BrickRed}{-4\%} & Prithvi & 27.86 & 29.21 & \textcolor{ForestGreen}{+5\%} \\
        CropMap & S12-Data2Vec & 54.03 & 56.27 & \textcolor{ForestGreen}{+4\%} & Scale-MAE & 25.42 & 20.17 & \textcolor{BrickRed}{-21\%} \\
        SN7 & Scale-MAE & 62.96 & 53.98 & \textcolor{BrickRed}{-14\%} & S12-DINO & 56.47 & 56.55 & 0\% \\
        AI4Farms & CROMA & 25.65 & 24.78 & \textcolor{BrickRed}{-3\%} & S12-MAE & 24.69 & 23.73 & \textcolor{BrickRed}{-4\%} \\
        BioMassters & SpectralGPT & 36.11 & 35.86 & \textcolor{BrickRed}{-1\%} & RemoteCLIP & 49.79 & 46.34 & \textcolor{BrickRed}{-7\%} \\
        
 \bottomrule
    \end{tabularx}
    \label{tab:finetuning-gfm}
\end{table*}

%% file: tabs/lr_tuning.tex
\begin{table*}[ht]
    \centering
    \caption{
    Performance comparison of various models on MADOS and HLS Burn Scars with a learning rate tuning. The table reports mIoU scores and the relative percentage difference (\% diff) between untuned (w/o) and tuned (w/) learning rate configurations.
    }
    \begin{tabularx}{\textwidth}{lYYrYYr}
    \toprule
         & \multicolumn{3}{c}{MADOS} & \multicolumn{3}{c}{HLSBurnScars} \\
         \cmidrule(lr){2-4} \cmidrule(lr){5-7}
         Model & w/o & w/ & \% diff & w/o & w/ & \% diff \\
         \midrule
        CROMA         & \textbf{67.55} & \textbf{66.98}  &    \textcolor{BrickRed}{-1\%}  & 82.42 & \underline{83.25} &  \textcolor{ForestGreen}{+1\%}  \\ 
        DOFA          & 59.58 & 61.56  &  \textcolor{ForestGreen}{+3\%}  & 80.63 & 82.04 &  \textcolor{ForestGreen}{+2\%}  \\ 
        GFM-Swin      & \underline{64.71} & \underline{66.48}  &  \textcolor{ForestGreen}{+3\%}  & 76.90 & 76.48 &    \textcolor{BrickRed}{-1\%}  \\ 
        Prithvi       & 49.98 & 53.83  &  \textcolor{ForestGreen}{+8\%}  & \underline{83.62} & 82.83 &    \textcolor{BrickRed}{-1\%}  \\ 
        RemoteCLIP    & 60.00 & 60.93  &  \textcolor{ForestGreen}{+2\%}  & 76.59 & 76.65 &                     0\%  \\ 
        SatlasNet     & 55.86 & 58.31  &  \textcolor{ForestGreen}{+4\%}  & 79.96 & 80.88 &  \textcolor{ForestGreen}{+1\%}  \\ 
        Scale-MAE     & 57.32 & 57.63  &  \textcolor{ForestGreen}{+1\%}  & 76.68 & 76.90 &                     0\%  \\ 
        SpectralGPT   & 57.99 & 58.13  &                     0\%   & 80.47 & 80.65 &                     0\%  \\ 
        S12-Data2Vec  & 44.36 & 54.29  & \textcolor{ForestGreen}{+22\%}  & 81.91 & 81.74 &                     0\%  \\ 
        S12-DINO      & 49.37 & 55.75  & \textcolor{ForestGreen}{+13\%}  & 81.72 & 81.67 &                     0\%  \\ 
        S12-MAE       & 49.90 & 51.66  &  \textcolor{ForestGreen}{+4\%}  & 81.91 & 81.18 &    \textcolor{BrickRed}{-1\%}  \\ 
        S12-MoCo      & 51.76 & 52.15  &  \textcolor{ForestGreen}{+1\%}  & 81.58 & 81.14 &    \textcolor{BrickRed}{-1\%}  \\\midrule
        UNet Baseline & 54.79 & 54.79  &                     0\%   & \textbf{84.51} & \textbf{85.68} &  \textcolor{ForestGreen}{+1\%}  \\ 
        ViT Baseline  & 48.19 & 48.52  &  \textcolor{ForestGreen}{+1\%}  & 81.58 & 80.76 &    \textcolor{BrickRed}{-1\%}\\
        \bottomrule
    \end{tabularx}
    \label{tab:lr-tuning}
\end{table*}

%% file: figs_tex/hlsburnscars_sample.tex
\begin{figure*}[htbp]
    \centering
    \begin{subfigure}[b]{0.16\textwidth}
        \centering
        \includegraphics[width=\textwidth]{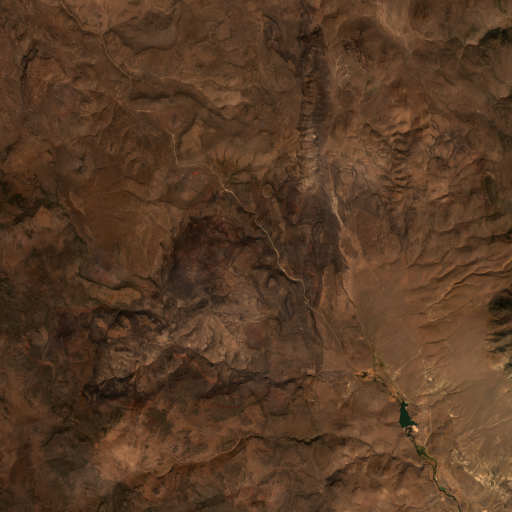} 
        \caption{Image 1}
        \label{fig:burn-sub1}
    \end{subfigure}
    \hfill
    \begin{subfigure}[b]{0.16\textwidth}
        \centering
        \includegraphics[width=\textwidth]{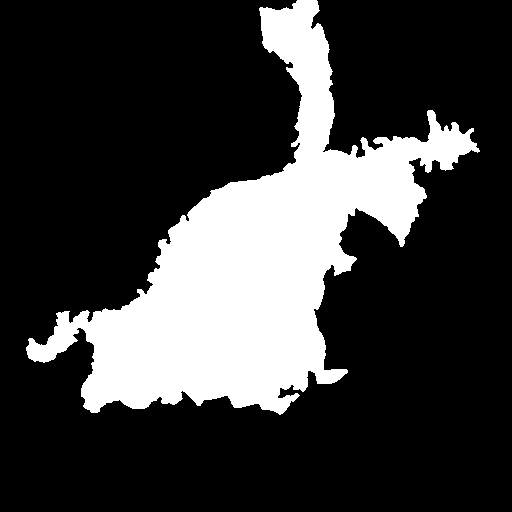} 
        \caption{Label 1}
        \label{fig:burn-sub2}
    \end{subfigure}
    \hfill
    \begin{subfigure}[b]{0.16\textwidth}
        \centering
        \includegraphics[width=\textwidth]{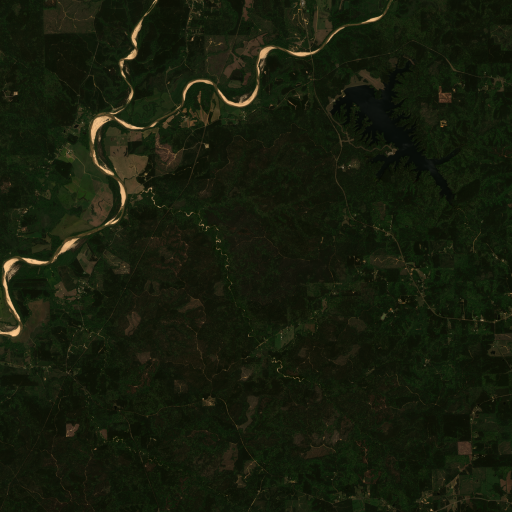} 
        \caption{Image 2}
        \label{fig:burn-sub3}
    \end{subfigure}
    \hfill
    \begin{subfigure}[b]{0.16\textwidth}
        \centering
        \includegraphics[width=\textwidth]{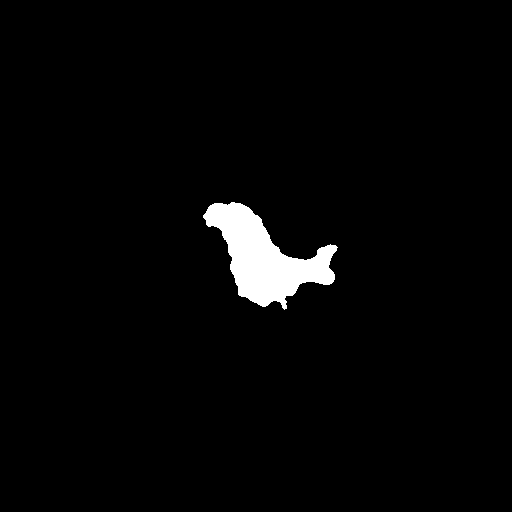} 
        \caption{Label 2}
        \label{fig:burn-sub4}
    \end{subfigure}
    \hfill
    \begin{subfigure}[b]{0.16\textwidth}
        \centering
        \includegraphics[width=\textwidth]{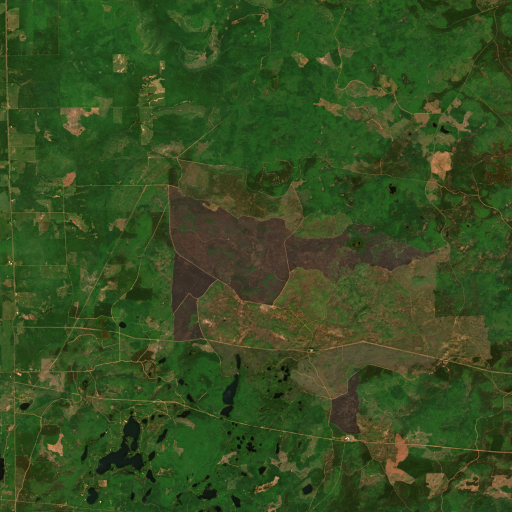} 
        \caption{Image 3}
        \label{fig:burn-sub5}
    \end{subfigure}
    \hfill
    \begin{subfigure}[b]{0.16\textwidth}
        \centering
        \includegraphics[width=\textwidth]{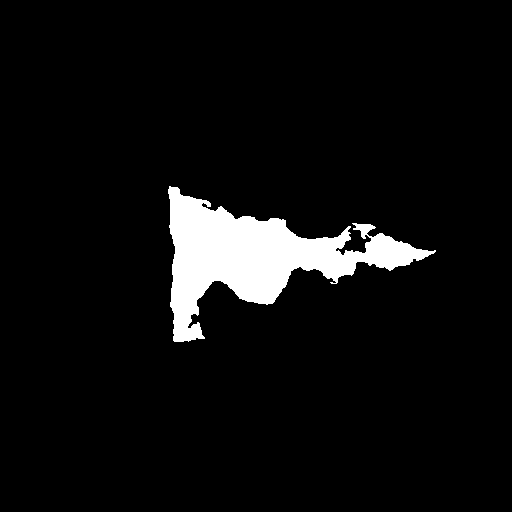} 
        \caption{Label 3}
        \label{fig:burn-sub6}
    \end{subfigure}
    \caption{Example images and labels of \textbf{HLS Burn Scars} (Location: USA. Domain: Wildfire). The images are visualized with B4, B3 and B2 bands.}
    \label{fig:hls-examples}
\end{figure*}

%% file: tabs/extensive_hlsburnscars.tex
\begin{table*}[htbp]
    \centering
    \caption{Results for the HLS Burn Scars comparison using stratified sampling with 10\%, 50\%, and 100\% labels for best and worst models from Table \ref{tab:comp_100}}
    \label{tab:hlsburnscars_extensive}
    \setlength{\tabcolsep}{3pt}
    \resizebox{\textwidth}{!}{%
    \begin{tabular}{lccccccccccccccc}
    \toprule
         &  \multicolumn{3}{c}{mIoU $\uparrow$}&  \multicolumn{3}{c}{m-F1 $\uparrow$}&  \multicolumn{3}{c}{m-prec $\uparrow$}&  \multicolumn{3}{c}{m-rec $\uparrow$}&  \multicolumn{3}{c}{m-acc $\uparrow$}\\
         \cmidrule(lr){2-4} \cmidrule(lr){5-7} \cmidrule(lr){8-10} \cmidrule(lr){11-13} \cmidrule(lr){14-16}
         Model &  10\%&  50\%&100\%&  10\%&  50\%&100\%&  10\%&  50\%&100\%&  10\%&  50\%&100\%& 10\%&50\%&100\%\\
         \midrule
         CROMA& 76.44& \underline{81.52}& \underline{81.95}& 85.42& \underline{89.10}& \underline{89.40}& 85.51& 88.31& \underline{89.13}& 85.32& \underline{89.94}& \underline{89.68}& 95.09& \underline{96.41}& \underline{96.46}\\
         DOFA& 71.98& 78.02& 78.96& 81.77& 86.60& 87.27& 79.45& 86.67& 85.84& 84.63& 86.54& 88.86& 94.33& 95.49& 95.89\\
         GFM-Swin& 67.23& 74.36& 76.17& 78.02& 83.79& 85.24& 83.30& 83.57& 86.92& 74.61& 84.01& 83.76& 91.28& 94.59& 94.81\\
         Prithvi& 77.73& 80.89& \underline{82.67}& 86.34& 88.66& \underline{89.90}& 83.71& 87.44& \underline{90.54}& \underline{89.55}& 89.97& 89.29& \underline{95.75}& 96.31& \underline{96.55}\\
         RemoteCLIP& 69.40& 74.29& 75.55& 79.59& 83.74& 84.77& 78.36& 84.09& 86.54& 80.98& 83.41& 83.21& 93.43& 94.48& 94.63\\
         SatlasNet& 74.79& 75.97& 79.69& 84.08& 85.01& 87.80& 81.78& 83.13& 86.49& 86.85& 87.19& 89.25& 95.02& 95.24& 96.05\\
         Scale-MAE& 75.47& 75.47& 76.71& 84.66& 84.66& 85.64& 84.16& 84.16& 86.96& 85.17& 85.17& 84.44& 94.92& 94.92& 95.00\\
         SpectralGPT& \textbf{83.35}& 76.40& 80.47& \textbf{90.67}& 85.50& 88.36& \textbf{94.49}& 87.87& 87.31& 88.00& 83.51& 89.49& 93.54& 94.33& 96.20\\
         S12-MoCo& 73.11& 79.79& 80.76& 82.85& 87.90& 88.58& 84.66& 88.68& 88.59& 81.27& 87.16& 88.56& 93.92& 95.85& 96.16\\
         S12-DINO& 75.93& 80.12& 81.44& 84.99& 88.13& 89.04& 83.58& 88.63& 88.06& 86.57& 87.66& 90.08& 95.16& 95.96& 96.41\\
         S12-MAE& 76.60& 80.13& 80.86& 85.49& 88.14& 88.66& 83.47& \textbf{88.88}& 89.88& 87.86& 87.44& 87.53& 95.41& 95.94& 96.07\\
         S12-Data2Vec& 74.38& 79.82& 81.14& 83.78& 87.91& 88.83& 82.78& 87.28& 87.73& 84.86& 88.56& 90.02& 94.71& 96.00& 96.35\\
         \midrule
         UNet Baseline& \underline{79.46}& \textbf{82.39}& \textbf{84.51}& \underline{87.60}& \textbf{89.68}& \textbf{91.11}& \underline{83.86}& 86.61& \textbf{88.66}& \textbf{92.57}& \textbf{93.49}& \textbf{94.00}& \textbf{96.25}& \textbf{96.80}& \textbf{97.19}\\
         ViT Baseline& 75.92& 78.17& 81.58& 84.98& 86.70& 89.16& 83.32& 86.12& 90.20& 86.87& 87.30& 88.19& 95.20& 95.60& 96.26\\
    \bottomrule
    \end{tabular}}
\end{table*}

%% file: figs_tex/mados_sample.tex
\begin{figure*}[htbp]
    \centering
    \begin{subfigure}[b]{0.16\textwidth}
        \centering
        \includegraphics[width=\textwidth]{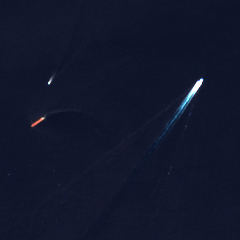} 
        \caption{Tile 1}
        \label{fig:mados-sub1}
    \end{subfigure}
    \hfill
    \begin{subfigure}[b]{0.16\textwidth}
        \centering
        \includegraphics[width=\textwidth]{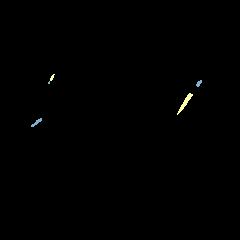} 
        \caption{Label 1}
        \label{fig:mados-sub2}
    \end{subfigure}
    \hfill
    \begin{subfigure}[b]{0.16\textwidth}
        \centering
        \includegraphics[width=\textwidth]{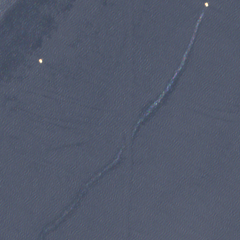} 
        \caption{Tile 2}
        \label{fig:mados-sub3}
    \end{subfigure}
    \hfill
    \begin{subfigure}[b]{0.16\textwidth}
        \centering
        \includegraphics[width=\textwidth]{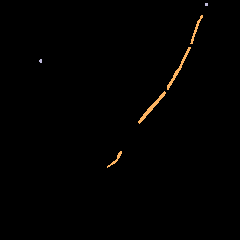} 
        \caption{Label 2}
        \label{fig:mados-sub4}
    \end{subfigure}
    \hfill
    \begin{subfigure}[b]{0.16\textwidth}
        \centering
        \includegraphics[width=\textwidth]{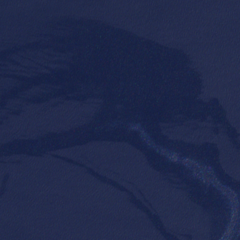} 
        \caption{Tile 3}
        \label{fig:mados-sub5}
    \end{subfigure}
    \hfill
    \begin{subfigure}[b]{0.16\textwidth}
        \centering
        \includegraphics[width=\textwidth]{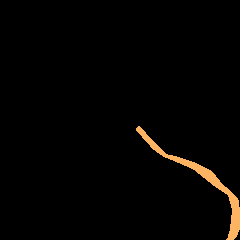} 
        \caption{Label 3}
        \label{fig:mados-sub6}
    \end{subfigure}
    \caption{Example images and labels from \textbf{MADOS}. (Location: Global. Domain: Marine.) The images are visualized with B4, B3, B2 bands. Black labels represent unlabeled pixels.}
    \label{fig:mados-examples}
\end{figure*}

%% file: tabs/extensive_mados.tex

\begin{table*}[htbp]
    \centering
    \caption{Results for MADOS dataset, best results in \textbf{bold}, second best \underline{underlined}.}
    \label{tab:mados-results}
    \begin{tabularx}{\textwidth}{lYYYYY}
    \toprule
        Model & {mIoU $\uparrow$}&  {m-F1 $\uparrow$}& {m-prec $\uparrow$}&  {m-rec $\uparrow$}& {m-acc $\uparrow$}\\
        \midrule
        CROMA&  \textbf{67.55}& \textbf{77.19}& 91.38& 90.85& 91.04\\
        DOFA&  59.58& 68.19& \underline{92.12}& \underline{94.26}& \underline{92.81}\\
        GFM-Swin&  \underline{64.71}& \underline{74.23}& \textbf{94.97}& \textbf{94.77}& \textbf{95.30}\\
        Prithvi& 49.98& 60.26& 83.10& 83.58& 83.92\\
        RemoteCLIP& 60.00& 70.48& 91.63& 90.59& 91.64\\
        SatlasNet& 55.86& 67.03& 86.92& 85.68& 87.56\\
        Scale-MAE& 57.32& 68.50& 91.28& 91.03& 90.92\\
        SpectralGPT& 57.99& 70.18& 84.95& 84.97& 85.27\\
        S12-Data2Vec& 44.36& 55.18& 80.18& 80.58& 82.41\\
        S12-DINO& 49.38& 60.65& 83.25& 82.93& 83.66\\
        S12-MAE& 49.90& 60.79& 84.29& 83.79& 84.70\\
        S12-MoCo& 51.76& 63.02& 84.88& 84.64& 85.21\\
        \midrule
        UNet Baseline& 54.79& 65.97& 85.20& 84.67& 85.61\\
        ViT Baseline& 48.19& 58.78& 58.14& 62.59& 88.00\\
    \bottomrule
    \end{tabularx}
\end{table*}

%% file: figs_tex/pastis_sample.tex
\begin{figure*}[htbp]
    \centering
    \begin{subfigure}[b]{0.18\textwidth}
        \centering
        \includegraphics[width=\textwidth]{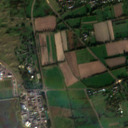} 
        \caption{S2 from 2018-09-29}
        \label{subfig:p1_optical_image}
    \end{subfigure}
    \hfill
    \begin{subfigure}[b]{0.18\textwidth}
        \centering
        \includegraphics[width=\textwidth]{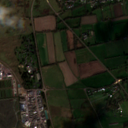} 
        \caption{S2 from 2018-09-29}
        \label{subfig:p2_optical_image}
    \end{subfigure}
    \hfill
    \begin{subfigure}[b]{0.18\textwidth}
        \centering
        \includegraphics[width=\textwidth]{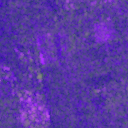} 
        \caption{S1 from 2018-10-04}
        \label{subfig:p1_sar_image}
    \end{subfigure}
    \hfill
    \begin{subfigure}[b]{0.18\textwidth}
        \centering
        \includegraphics[width=\textwidth]{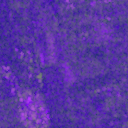} 
        \caption{S1 from 2018-10-04}
        \label{subfig:p2_sar_image}
    \end{subfigure}
    \hfill
    \begin{subfigure}[b]{0.18\textwidth}
        \centering
        \includegraphics[width=\textwidth]{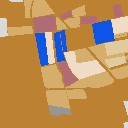} 
        \caption{Segmentation label}
        \label{subfig:labels}
    \end{subfigure}
    
    \caption{Example time series data of \textbf{PASTIS-R} (Location: France. Domain: Agriculture.). The Sentinel-1 images are visualized by taking VV, VH, VV-VH bands
, and the Sentinel-2 images are visualized by taking B4, B3, and B2 bands.}
    \label{fig:pastis-examples}
\end{figure*}

%% file: tabs/extensive_pastis.tex
\begin{table*}[htbp]
    \centering
    \caption{Results for PASTIS-R dataset with 100\% of the data, best results in \textbf{bold}, second best \underline{underlined}.}
    \label{tab:pastis-linear-ltae-100}
    \resizebox{\textwidth}{!}{%
    \begin{tabular}{lccccccccccc}
    \toprule
        \multicolumn{2}{c}{}&  \multicolumn{2}{c}{mIoU $\uparrow$}&  \multicolumn{2}{c}{m-F1 $\uparrow$}&  \multicolumn{2}{c}{m-prec $\uparrow$}&  \multicolumn{2}{c}{ m-rec $\uparrow$}& \multicolumn{2}{c}{ m-acc $\uparrow$}\\
        \cmidrule(lr){3-4} \cmidrule(lr){5-6} \cmidrule(lr){7-8} \cmidrule(lr){9-10} \cmidrule(lr){11-12}
        Model&  Modality&  Linear&  L-TAE&  Linear&  L-TAE&  Linear&  L-TAE&  Linear&  L-TAE& Linear&L-TAE\\
        \midrule
        CROMA&  joint&  27.75&  32.32&  39.91&  45.27&  36.79&  44.09&  49.28&  51.33& 71.93&74.47\\
        CROMA&  optical&  32.06&  \underline{38.51}&  44.92&  \underline{52.53}&  41.97&  \underline{48.76}&  {52.95}&  \textbf{62.66}& \underline{74.58}&\underline{77.76}\\
        CROMA&  sar&  20.68&  21.61&  30.12&  31.93&  28.82&  32.22&  34.29&  39.43& 68.72&66.91\\
        DOFA&  joint&  24.49&  30.02&  35.88&  42.37&  33.28&  39.45&  43.64&  50.54& 70.29&74.09\\
        GFM-Swin&  optical&  15.13&  21.24&  23.45&  32.02&  21.71&  28.71&  30.18&  45.36& 62.11&67.39\\
        Prithvi&  optical&  27.40&  33.93&  39.56&  46.95&  36.69&  43.95&  47.22&  56.46& 71.60&75.75\\
        RemoteCLIP& optical& 13.31& 18.23& 20.83& 27.57& 21.34& 26.57& 23.17& 32.46& 58.39&65.80\\
        SatlasNet& optical& \multicolumn{2}{c}{17.51}& \multicolumn{2}{c}{27.14}& \multicolumn{2}{c}{23.97}& \multicolumn{2}{c}{38.03}& \multicolumn{2}{c}{63.04}\\
        Scale-MAE& optical&  11.74& 24.55& 17.45& 35.75& 17.34& 33.64& 28.00& 43.69& 61.91&71.19\\
        SpectralGPT& optical& 27.52& 35.44& 40.05& 49.19& 38.22& 46.60& 46.59& 56.67& 72.84&76.11\\
        S12-MoCo& optical& 26.09& 34.49& 38.32& 48.07& 35.08& 44.97& 47.05& 57.21& 69.98&75.43\\
        S12-DINO& optical& \underline{32.45}& 36.18& \underline{45.86}& 49.89& \underline{42.34}& 46.17& \underline{54.95}& 60.88& 73.60&76.05\\
        S12-MAE& optical& 27.06& 32.03& 39.22& 45.23& 35.32& 41.09& 50.61& 56.21& 69.62&74.09\\
        S12-MAE& sar& 18.81& 14.20& 28.08& 21.71& 26.41& 21.45& 18.81& 34.13& 66.40&58.48\\
        S12-Data2Vec& optical& 26.94& 34.32& 39.22& 47.61& 36.34& 44.89& 47.01& 59.39& 70.56&75.28\\
 \midrule
 UNet Baseline& optical& \multicolumn{2}{c}{31.60}& \multicolumn{2}{c}{43.52}& \multicolumn{2}{c}{39.37}& \multicolumn{2}{c}{51.57}& \multicolumn{2}{c}{72.12}\\
 ViT Baseline& optical& \textbf{36.57}& \textbf{38.53}& \textbf{50.03}& \textbf{54.89}& \textbf{46.38}& \textbf{51.53}& \textbf{59.46}& \underline{62.61}& \textbf{75.96}&\textbf{78.34}\\
 \bottomrule
    \end{tabular}
    }
\end{table*}

\begin{table*}[htbp]
    \centering
    \caption{Results for PASTIS-R dataset with linear time strategy, best results in \textbf{bold}, second best \underline{underlined}.}
    \label{tab:pastis-linear}
    \setlength{\tabcolsep}{3pt}
    \resizebox{\textwidth}{!}{%
    \begin{tabular}{lcccccccccccccccc}
    \toprule
         \multicolumn{2}{c}{}&  \multicolumn{3}{c}{mIoU $\uparrow$}&  \multicolumn{3}{c}{m-F1 $\uparrow$}&  \multicolumn{3}{c}{m-prec $\uparrow$}&   \multicolumn{3}{c}{m-rec $\uparrow$}&  \multicolumn{3}{c}{m-acc $\uparrow$}\\
         \cmidrule(lr){3-5} \cmidrule(lr){6-8} \cmidrule(lr){9-11} \cmidrule(lr){12-14} \cmidrule(lr){15-17}
         Model&  Modality&  10\%&  50\%&100\%&  10\%&  50\%&100\%&  10\%&  50\%&100\%&  10\%&  50\%&100\%& 10\%&50\%&100\%\\
         \midrule
         CROMA&  joint&  27.56&  27.59&27.75&  39.61&  39.74&39.91&  36.55&  36.80&36.79&  48.64&  48.80&49.28& 71.76&71.63&71.93\\
         CROMA&  optical&  \underline{31.93}&  \underline{31.41}&32.06&  \underline{44.80}&  \underline{44.01}&44.92&  \underline{41.64}&  40.26&41.97&  53.73&  \underline{53.83}&52.95& \underline{74.22}&\underline{74.09}&\underline{74.58}\\
         CROMA&  sar&  19.47&  19.63&20.68&  28.88&  29.02&30.12&  27.24&  27.83&28.82&  33.55&  32.93&34.29& 67.35&67.66&68.72\\
         DOFA&  joint&  23.72&  24.17&24.49&  34.99&  35.45&35.88&  31.83&  33.31&33.28&  43.22&  43.08&43.64& 69.31&69.78&70.29\\
         GFM-Swin&  optical&  15.58&  15.79&15.13&  24.13&  24.72&23.45&  23.63&  23.58&21.71&  28.78&  28.46&30.18& 60.69&60.36&62.11\\
         Prithvi&  optical&  26.33&  26.92&27.40&  38.23&  39.05&39.56&  35.76&  36.54&36.69&  45.02&  46.01&47.22& 70.63&70.75&71.60\\
 RemoteCLIP& optical& 12.25& 12.91&13.31& 19.08& 20.20&20.83& 18.73& 20.33&21.34& 21.84& 21.75&23.17& 58.35&57.69&58.39\\
 SatlasNet& optical& 16.76& 16.78&17.51& 26.11& 26.00&26.98& 23.09& 23.54&23.99& 34.82& 37.40&37.12& 61.58&61.86&62.98\\
 Scale-MAE& optical&  10.40& 10.45&11.74& 15.57& 15.87&17.45& 15.19& 15.56&17.34& 28.12& 29.97&28.00& 59.73&58.61&61.91\\
 SpectralGPT& optical& 24.74& 25.74&27.52& 36.59& 37.83&40.05& 34.29& 35.69&38.22& 43.69& 44.54&46.59& 70.96&71.47&72.84\\
 S12-MoCo& optical& 24.29& 24.79&26.09& 35.97& 36.67&38.32& 33.06& 33.09&35.08& 44.53& 46.53&47.05& 68.97&69.19&69.98\\
 S12-DINO& optical& 30.84& 30.82&\underline{32.45}& 43.93& 43.84 &\underline{45.86}& 40.82& \underline{40.44}&\underline{42.34}& \underline{53.97}& 52.95&\underline{54.95}& 72.43&72.56&73.60\\
 S12-MAE& optical& 25.31& 26.40&27.06& 36.68& 38.40&39.22& 33.48& 35.08&35.32& 45.94& 47.44&50.61& 68.48&68.27&69.62\\
 S12-MAE& sar& 17.29& 17.54&18.81& 25.86& 26.04&28.08& 24.51& 24.75&26.41& 29.58& 30.20&18.81& 65.23&66.12&66.40\\
 S12-Data2Vec& optical& 24.65& 24.52&26.94& 36.11& 35.82&39.22& 33.28& 32.38&36.34& 44.15& 45.97&47.01& 68.72&68.85&70.56\\
 \midrule
 UNet Baseline& optical& 29.52& 30.24 & 31.60& 41.19& 41.87& 43.52& 39.70& 40.40& 39.37& 52.27& 51.53& 51.57& 74.57& 75.11 &72.12\\
 ViT Baseline& optical& \textbf{36.38}& \textbf{36.13}& \textbf{36.57}& \textbf{50.07}& \textbf{49.74}& \textbf{50.03}& \textbf{46.16}& \textbf{46.21}& \textbf{46.38}& \textbf{59.32}& \textbf{59.43}& \textbf{59.46}&\textbf{75.37} &\textbf{75.36} &\textbf{75.96}\\
 \bottomrule
    \end{tabular}}
\end{table*}

\begin{table*}[htbp]
    \centering
    \caption{Results for PASTIS-R dataset with L-TAE time strategy, best results in \textbf{bold}, second best \underline{underlined}.}
    \label{tab:pastis-ltae}
    \setlength{\tabcolsep}{3pt}
    \resizebox{\textwidth}{!}{%
    \begin{tabular}{lcccccccccccccccc}
    \toprule
         \multicolumn{2}{c}{}&   \multicolumn{3}{c}{mIoU $\uparrow$}&   \multicolumn{3}{c}{m-F1 $\uparrow$}&   \multicolumn{3}{c}{m-prec $\uparrow$}&    \multicolumn{3}{c}{m-rec $\uparrow$}&   \multicolumn{3}{c}{m-acc $\uparrow$}\\
         \cmidrule(lr){3-5} \cmidrule(lr){6-8} \cmidrule(lr){9-11} \cmidrule(lr){12-14} \cmidrule(lr){15-17}         
         Model&  Modality&   10\%&50\%&  100\%&   10\%&50\%&  100\%&   10\%&50\%&  100\%&   10\%&50\%&  100\%&  10\%&50\%&100\%\\
         \midrule
         CROMA&  joint&   32.80&32.33&  32.32&   45.73&45.12&  45.27&   42.59&42.21&  44.09&   54.42&53.12&  51.33&  75.37&74.89&74.47\\
         CROMA&  optical&   \underline{37.24}&\underline{37.76}&  \underline{38.51}&   \underline{50.94}&\underline{51.52}&  \underline{52.53}&   46.76&\underline{48.49}&  \underline{48.76}&   \textbf{62.19}&\textbf{61.78}&  \textbf{62.66}&  \textbf{77.17}&\underline{77.47}&\underline{77.76}\\
         CROMA&  sar&   21.62&21.31&  21.61&   31.42&31.21&  31.93&   30.78&29.97&  32.22&   36.80&35.75&  39.43&  68.31&67.01&66.91\\
         DOFA&  joint&   27.68&28.60&  30.02&   39.32&40.43&  42.37&   36.16&37.52&  39.45&   49.52&50.16&  50.54&  73.13&73.69&74.09\\
         GFM-Swin&  optical&   21.47&20.41&  21.24&   32.01&30.85&  32.02&   28.59&27.66&  28.71&   42.51&43.76&  45.36&  65.33&67.98&67.39\\
         Prithvi&  optical&   33.56&33.13&  33.93&   46.96&46.01&  46.95&   43.55&43.23&  43.95&   55.05&54.19&  56.46&  74.58&75.51&75.75\\
 RemoteCLIP& optical&  17.19&17.46& 18.23&  26.52&26.66& 27.57&  25.39&26.15& 26.57&  30.01&32.88& 32.46&  63.14&64.23&65.80\\
 SatlasNet& optical&  16.76&16.78& 17.51&  26.11&26.00& 26.98&  23.09&23.54& 23.99&  34.82&37.40& 37.12&  61.58&61.86&62.98\\
 Scale-MAE& optical&   22.86&23.26& 24.55&  33.44&34.56& 35.75&  33.25&23.02& 33.64&  40.88&38.32& 43.69&  68.75&68.45&71.19\\
 SpectralGPT& optical&  34.53&34.61& 35.44&  48.16&48.25& 49.19&  45.59&45.96& 46.60&  54.33&54.46& 56.67&  75.95&75.64&76.11\\
 S12-MoCo& optical&  32.51&32.59& 34.49&  45.66&45.84& 48.07&  42.38&42.91& 44.97&  55.89&57.71& 57.21&  74.28&74.02&75.43\\
 S12-DINO& optical&  36.62& 35.71& 36.18&  50.68&49.48& 49.89& 47.47&45.48& 46.17&  \underline{60.87}&60.33& 60.88&  75.76&75.28&76.05\\
 S12-MAE& optical&  31.06&31.15& 32.03&  44.05&44.15& 45.23&  40.74&39.38& 41.09&  54.39&56.17& 56.21&  73.47&73.38&74.09\\
 S12-MAE& sar&  13.00&14.33& 14.20&  19.58&21.67& 21.71&  20.34&22.13& 21.45&  24.20&31.92& 34.13&  60.34&59.07&58.48\\
 S12-Data2Vec& optical&  33.09&33.42& 34.32&  45.98&46.63& 47.61&  43.46&43.60& 44.89&  56.51&56.14& 59.39&  74.58&74.31&75.28\\
 \midrule
  UNet Baseline& optical& 29.52& 30.24 & 31.60& 41.19& 41.87& 43.52& 39.70& 40.40& 39.37& 52.27& 51.53& 51.57& 74.57& 75.11 &72.12\\
 
 ViT Baseline& optical& \textbf{38.44}& \textbf{38.77}& \textbf{38.53}& \textbf{52.49}& \textbf{52.59}& \textbf{54.89}& \textbf{49.51}& \textbf{49.25}& \textbf{51.53}& 60.35& \underline{61.27}& \underline{62.61}& \underline{76.93}& \textbf{77.71}&\textbf{78.34}\\
\bottomrule    
\end{tabular}}
\end{table*}

%% file: figs_tex/sen1floods11_sample.tex
\begin{figure*}[htbp]
    \centering
    \begin{subfigure}[b]{0.16\textwidth}
        \centering
        \includegraphics[width=\textwidth]{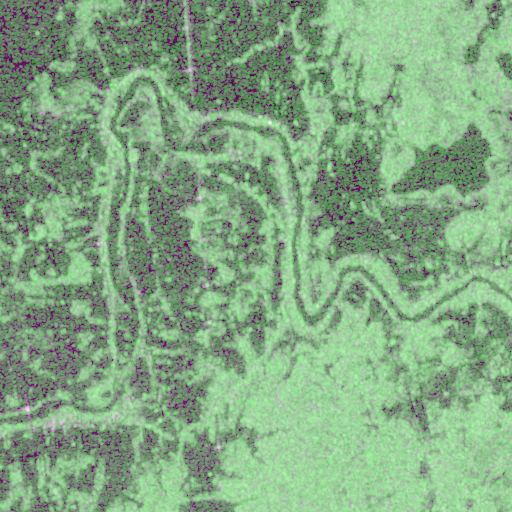} 
        \caption{S1 image 1}
        \label{fig:floods-sub1}
    \end{subfigure}
    \hfill
    \begin{subfigure}[b]{0.16\textwidth}
        \centering
        \includegraphics[width=\textwidth]{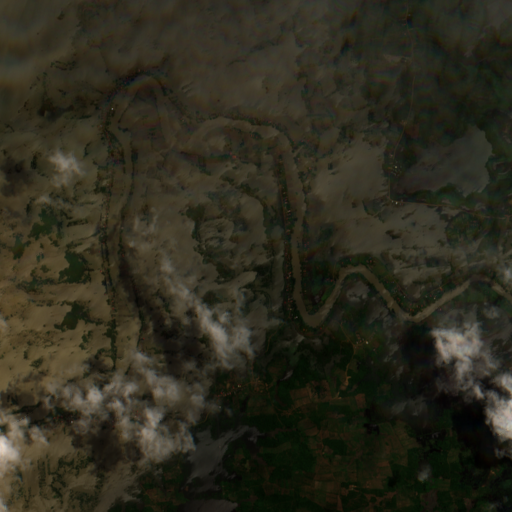} 
        \caption{S2 image 1}
        \label{fig:floods-sub2}
    \end{subfigure}
    \hfill
    \begin{subfigure}[b]{0.16\textwidth}
        \centering
        \includegraphics[width=\textwidth]{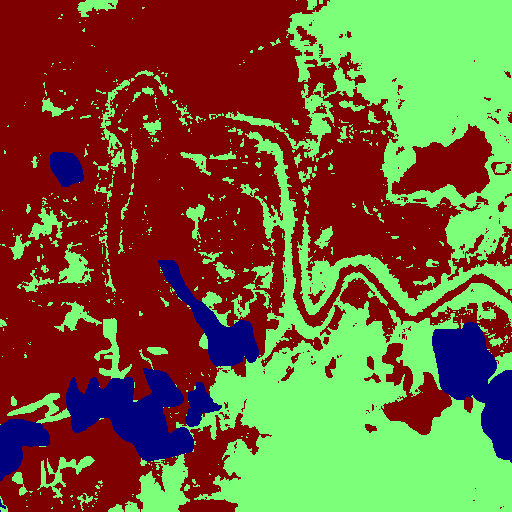} 
        \caption{Label 1}
        \label{fig:floods-sub3}
    \end{subfigure}
    \begin{subfigure}[b]{0.16\textwidth}
        \centering
        \includegraphics[width=\textwidth]{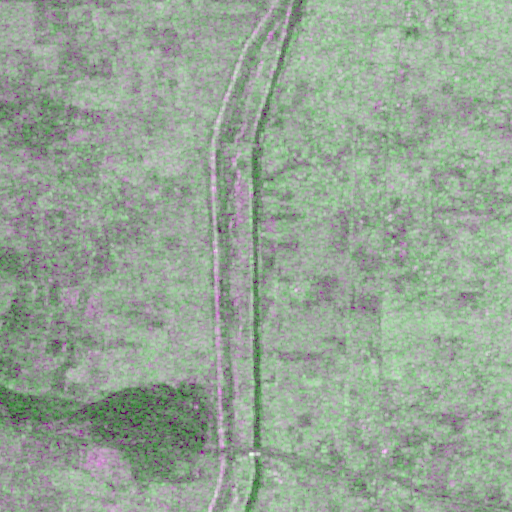} 
        \caption{S1 image 2}
        \label{fig:floods-sub4}
    \end{subfigure}
    \hfill
    \begin{subfigure}[b]{0.16\textwidth}
        \centering
        \includegraphics[width=\textwidth]{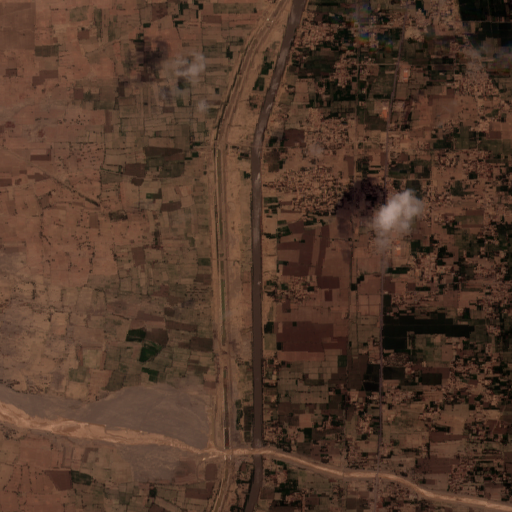} 
        \caption{S2 image 2}
        \label{fig:floods-sub5}
    \end{subfigure}
    \hfill
    \begin{subfigure}[b]{0.16\textwidth}
        \centering
        \includegraphics[width=\textwidth]{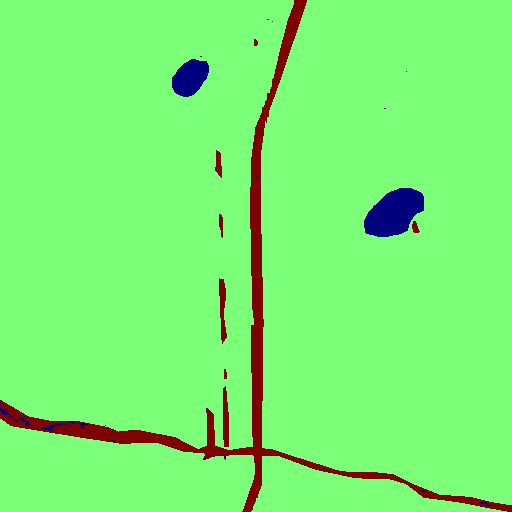} 
        \caption{Label 2}
        \label{fig:floods-sub6}
    \end{subfigure}
    
    \caption{Example data of \textbf{Sen1Floods11} (Location: Global. Domain: Flood.). The Sentinel1 images are visualized by taking the VV, VH, VV bands, and the Sentinel-2 images are visualized by taking the B4, B3, and B2 bands. In the semantic labels, red represents water, green indicates areas without water, and blue marks cloudy pixels that are ignored.}
    \label{fig:sen1floodss11-examples}
\end{figure*}

%% file: tabs/extensive_sen1floods11_scarcity.tex
\begin{table*}[htbp]
\centering
\caption{Sen1Floods11 results using different limited training labels with best results in \textbf{bold}, second best \underline{underlined}.}
\label{tab:extensive-floods-scarcity}
\resizebox{\textwidth}{!}{%
    \setlength{\tabcolsep}{4pt}
    \begin{tabular}{lcccccccccccc}
    \toprule
        &   \multicolumn{3}{c}{mIoU $\uparrow$}&   \multicolumn{3}{c}{m-F1 $\uparrow$}&   \multicolumn{3}{c}{water IoU $\uparrow$}&    \multicolumn{3}{c}{water F1 $\uparrow$} \\
        \cmidrule(lr){2-4}         \cmidrule(lr){5-7}         \cmidrule(lr){8-10}         \cmidrule(lr){11-13}
         Model&   10\% & 50\% & 100\% & 10\% & 50\% & 100\% & 10\% & 50\% & 100\% & 10\% & 50\% & 100\% \\
        
        \midrule 
        CROMA & \underline{87.22} & \underline{90.57} & \underline{90.89} & \underline{92.90} & \underline{94.91} & \underline{95.09} & \underline{77.71} & \underline{83.59} & \underline{84.15} & \underline{87.46} & \underline{91.06} & \underline{91.39} \\  
        DOFA & 82.84 & 88.39 & 89.37 & 90.13 & 93.62 & 94.20 & 70.70 & 79.84 & 81.54 & 82.83 & 88.79 & 89.83 \\ 
        GFM-Swin & 62.57 & 71.61 & 72.60 & 73.11 & 81.70 & 82.51 & 34.21 & 50.68 & 52.36 & 50.98 & 67.26 & 68.73  \\ 
        Prithvi & 86.28 & 89.69 & 90.37 & 92.32 & 94.40 & 94.80 & 76.21 & 82.11 & 83.26 & 86.50 & 90.18 & 90.86 \\ 
        RemoteCLIP & 62.22 & 71.66 & 74.26 & 72.77 & 81.76 & 83.83 & 33.67 & 50.85 & 55.18 & 50.38 & 67.42 & 71.11 \\
        SatlasNet & 83.92 & 89.45 & 90.30 & 90.82 & 94.25 & 87.51 & 72.30 & 81.65 & 83.12 & 83.92 & 89.90 & 90.79   \\ 
        Scale-MAE & 64.74 & 72.54 & 74.13 & 75.51 & 82.43 & 83.75 & 38.81 & 52.09 & 55.04 & 55.92 & 68.50 & 71.00 \\ 
        SpectralGPT & 83.12 & 87.52 & 89.07 & 90.31 & 93.09 & 94.03 & 71.11 & 78.29 & 81.02 & 83.12 & 87.82 & 89.52 \\
        S12-MoCo & 79.58 & 89.22 & 89.26  & 87.92 & 94.12  & 94.14 & 65.24 & 81.29 & 81.37 & 78.96 & 89.68 & 89.73 \\ 
        S12-DINO & 84.95 & 88.93 & 88.61 & 91.48 & 93.94 & 93.75 & 73.96 & 80.76 & 80.25 & 85.03 & 89.35 & 89.04   \\ 
        S12-MAE & 84.81 & 88.43 & 87.79 & 91.39 & 93.64 & 93.25 & 73.78 & 79.93 & 78.79 & 84.91 & 88.85 & 88.13  \\ 
        S12-Data2Vec & 81.91 & 86.58 & 88.15 & 89.48 & 92.51 & 93.47 & 68.80 & 76.71 & 79.44 & 81.52 & 86.82 & 88.54   \\ 
        
        \midrule
        UNet Baseline & \textbf{88.55} & \textbf{90.91} & \textbf{91.42} & \textbf{93.72} & \textbf{95.11} & \textbf{95.40} & \textbf{80.19} & \textbf{84.19} & \textbf{85.05} & \textbf{89.01} & \textbf{91.42} & \textbf{91.92}  \\ 
        ViT Baseline & 81.85 & 86.08 & 87.66 & 89.40 & 92.19 & 93.71 & 68.28 & 75.81 & 78.56 & 81.15 & 86.24 & 87.99\\
        \bottomrule
    \end{tabular}
}
\end{table*}

%% file: figs_tex/xview2_5bp_samples.tex
\begin{figure*}[htbp]
    \centering
    \begin{minipage}{0.6\textwidth}
        \centering
    \begin{subfigure}[b]{0.32\textwidth}
        \centering
        \includegraphics[width=\textwidth]{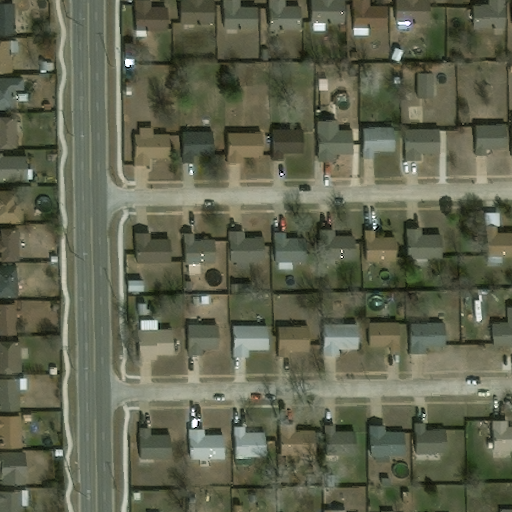} 
        \caption{Pre-disaster image}
        \label{fig:sub1}
    \end{subfigure}
    \hfill
    \begin{subfigure}[b]{0.32\textwidth}
        \centering
        \includegraphics[width=\textwidth]{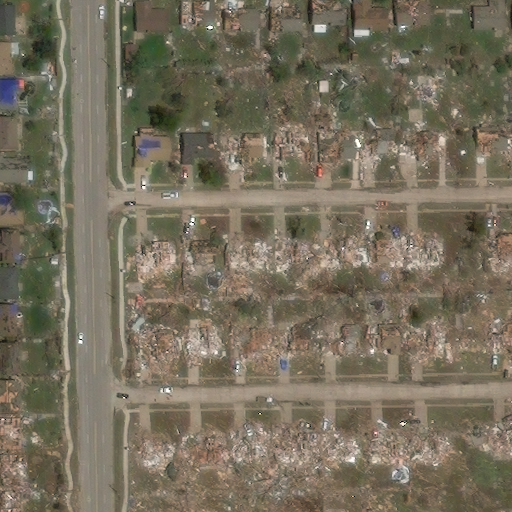} 
        \caption{Post-disaster image}
        \label{fig:sub2}
    \end{subfigure}
    \hfill
    \begin{subfigure}[b]{0.32\textwidth}
        \centering
        \includegraphics[width=\textwidth]{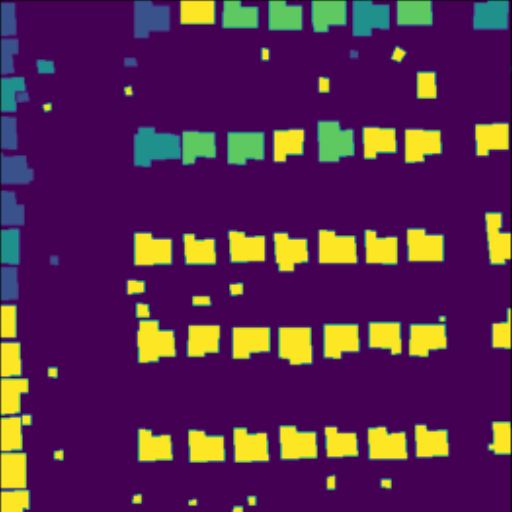} 
        \caption{Damage label}
        \label{fig:sub3}
    \end{subfigure}
    
    \caption{Example data of \textbf{xView2} (Location: Global. Domain: HADR.). The image is taken from the 2013 Moore Tornado. Different colors in the label indicate different building damage classes.}
    \label{fig:xview2-examples}
    \end{minipage}
    \hfill
    \begin{minipage}{0.38\textwidth}
        \centering
        \begin{subfigure}[b]{0.48\textwidth}
        \centering 
        \includegraphics[width=\textwidth]{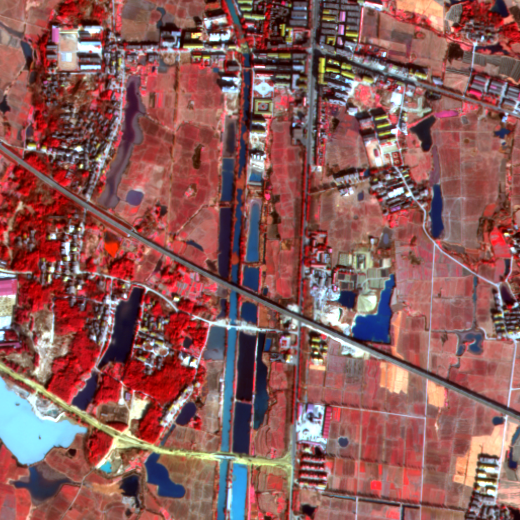} 
        \caption{Image} 
        \label{fig:subfig1}
    \end{subfigure}
    \hfill
    \begin{subfigure}[b]{0.48\textwidth}
        \centering
        \includegraphics[width=\textwidth]{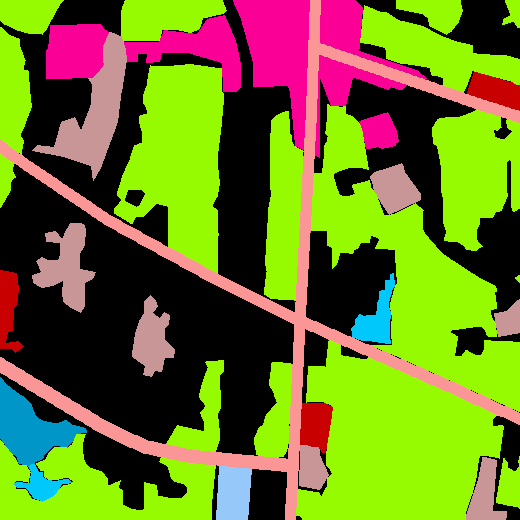} 
        \caption{Land cover label}
        \label{fig:subfig2}
    \end{subfigure}
    \caption{Example data of \textbf{Five Billion Pixels} (Location: China. Domain: Urban Land Cover). The label colors indicate land cover categories.}

    \label{fig:fbps_example}
    \end{minipage}

    \label{fig:twofigs}
\end{figure*}

%% file: tabs/extensive_xview2.tex
\begin{table*}[htbp]
    \centering
    \caption{Bi-temporal change detection results on xView2 using 100\% of the training set, best results in \textbf{bold}, second best \underline{underlined}.}
    \label{tab:ext-results-xview2}
    \setlength{\tabcolsep}{4pt}
    \begin{tabularx}{\textwidth}{lcccYYYYY}
    \toprule
         \multicolumn{1}{c}{} & & & & \multicolumn{5}{c}{Per-class IoU $\uparrow$}\\
         \cmidrule(lr){5-9}
         Model & \multicolumn{1}{c}{mIoU $\uparrow$} & \multicolumn{1}{c}{m-F1 $\uparrow$} & \multicolumn{1}{c}{m-acc $\uparrow$} & No build & No dam & Minor dam & Major dam & Destr. \\
         \midrule
         CROMA & 53.27 & 65.81 & 97.21 & 97.77 & 60.64 & 19.97 & 36.13 & 51.85 \\
         DOFA & \underline{59.64} & \underline{71.73} & \underline{97.88} & 98.33 & \underline{70.22} & 25.27 & 45.45 & \underline{58.91} \\
         GFMSwin & 59.15 & 71.45 & 97.73 & 98.19 & 68.11 & \underline{25.92} & 45.45 & 58.06 \\
         Prithvi & 49.35 & 61.92 & 96.92 & 97.36 & 56.66 & 16.54 & 35.13 & 41.05 \\
         RemoteCLIP & 57.41 & 69.87 & 97.58 & 98.01 & 65.75 & 24.05 & 43.76 & 55.50 \\
         SatlasNet & 52.23 & 64.66 & 97.14 & 97.76 & 59.96 & 19.25 & 32.43 & 51.78 \\
         Scale-MAE & \textbf{60.72} & \textbf{72.70} & \textbf{97.99} & \textbf{98.45} & \textbf{71.43} & \textbf{26.18} & \textbf{47.76} & \textbf{59.78} \\
         SpectralGPT & 48.40 & 60.89 & 96.72 & 97.36 & 54.20 & 17.06 & 28.64 & 44.74 \\
         S12-MoCo & 51.59 & 64.29 & 97.09 & 97.57 & 58.71 & 18.97 & 37.56 & 45.14 \\
         S12-DINO & 50.56 & 63.38 & 96.95 & 97.39 & 57.03 & 19.36 & 36.26 & 42.77 \\
         S12-MAE & 50.44 & 63.05 & 96.99 & 97.44 & 57.95 & 17.44 & 36.15 & 43.24 \\
         S12-Data2Vec & 51.36 & 64.14 & 97.03 & 97.49 & 57.97 & 19.29 & 38.32 & 43.73 \\
         \midrule
         UNet Baseline & 58.68 & 70.80 & 97.82 & \underline{98.34} & 68.97 & 23.62 & 44.49 & 57.96 \\
         ViT Baseline & 57.43 & 70.00 & 97.53 & 97.91 & 65.26 & 24.36 & \underline{46.74} & 52.87 \\
    \bottomrule
    \end{tabularx}
\end{table*}

%% file: tabs/extensive_fbps.tex

\begin{table*}[htbp]
    \centering
    \caption{Results for FiveBillionPixels dataset, best results in \textbf{bold}, second best \underline{underlined}.}
    \label{tab:extensive-fbps}
    \setlength{\tabcolsep}{6pt}
    \begin{tabularx}{\textwidth}{lYYYYY}
    \toprule
         Model& {mIoU $\uparrow$}&  {m-F1 $\uparrow$}& {m-prec $\uparrow$}&  {m-rec $\uparrow$}& {m-acc $\uparrow$}\\
        \midrule
        CROMA& 51.85& 63.50& 59.96& 77.34& 90.14\\
        DOFA& 43.80& 54.14& 51.22& 66.71& 89.00\\
        GFM-Swin& 67.19& 77.67& \underline{76.28}& 80.30& \underline{98.12}\\
        Prithvi& 46.82& 58.15& 55.39& 70.83& 91.20\\
        RemoteCLIP& \textbf{69.19}& \textbf{79.18}& \textbf{77.64}& \underline{82.57}& \textbf{98.31}\\
        SatlasNet& 50.97& 62.89& 59.94& 70.63& 93.18\\
        Scale-MAE& \underline{67.19}& \underline{78.12}& 75.10& \textbf{82.28}& 98.05\\
        SpectralGPT& 33.42& 42.15& 41.76& 48.11& 85.24\\
        S12-Data2Vec& 48.82& 59.53& 56.85& 69.26& 92.73\\
        S12-DINO& 51.15& 62.26& 59.51& 71.20& 93.53\\
        S12-MAE& 51.50& 62.69& 59.79& 69.98& 93.43\\
        S12-MoCo& 53.42& 64.40& 61.80& 69.97& 94.32\\
        \midrule
        UNet Baseline& 60.68& 70.92& 70.14& 72.80& 97.27\\
        ViT Baseline& 59.32& 69.54&68.95& 71.42&93.28 \\
    \bottomrule
    \end{tabularx}
\end{table*}

%% file: figs_tex/dyn_sample.tex
\begin{figure*}[htbp]
    \centering
    \begin{subfigure}[b]{0.16\textwidth}
        \centering
        \includegraphics[width=\textwidth]{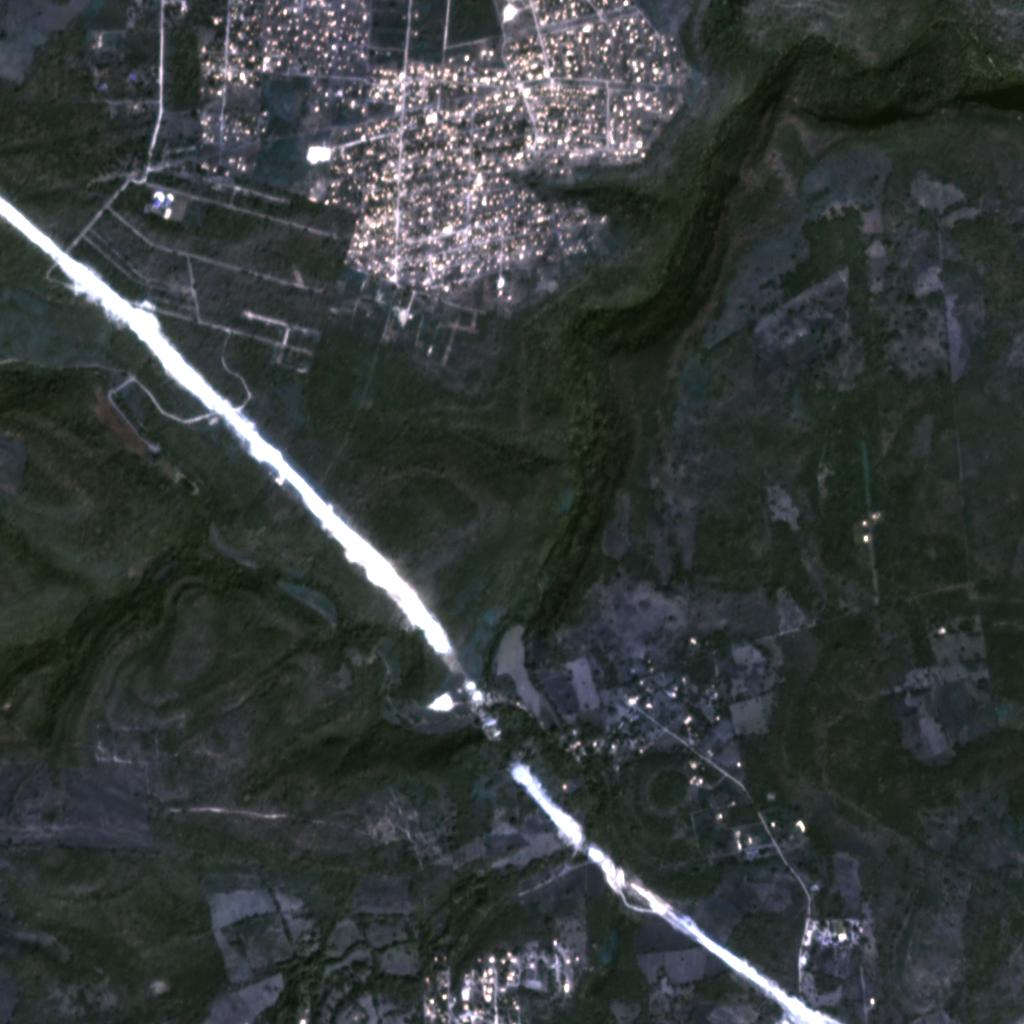} 
        \caption{2018-01-01}
        \label{fig:dyn-sub1}
    \end{subfigure}
    \hfill
    \begin{subfigure}[b]{0.16\textwidth}
        \centering
        \includegraphics[width=\textwidth]{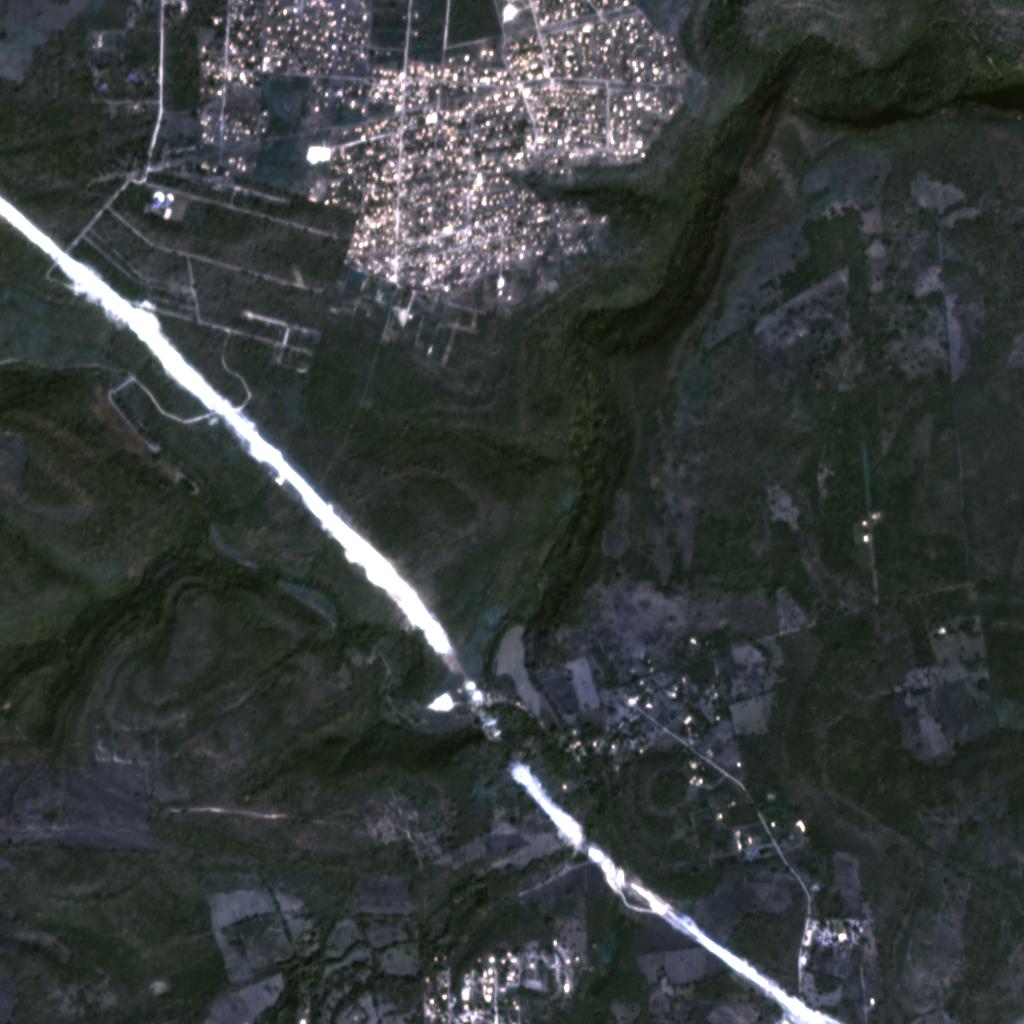} 
        \caption{2018-01-02}
        \label{fig:dyn-sub2}
    \end{subfigure}
    \hfill
    \begin{subfigure}[b]{0.16\textwidth}
        \centering
        \includegraphics[width=\textwidth]{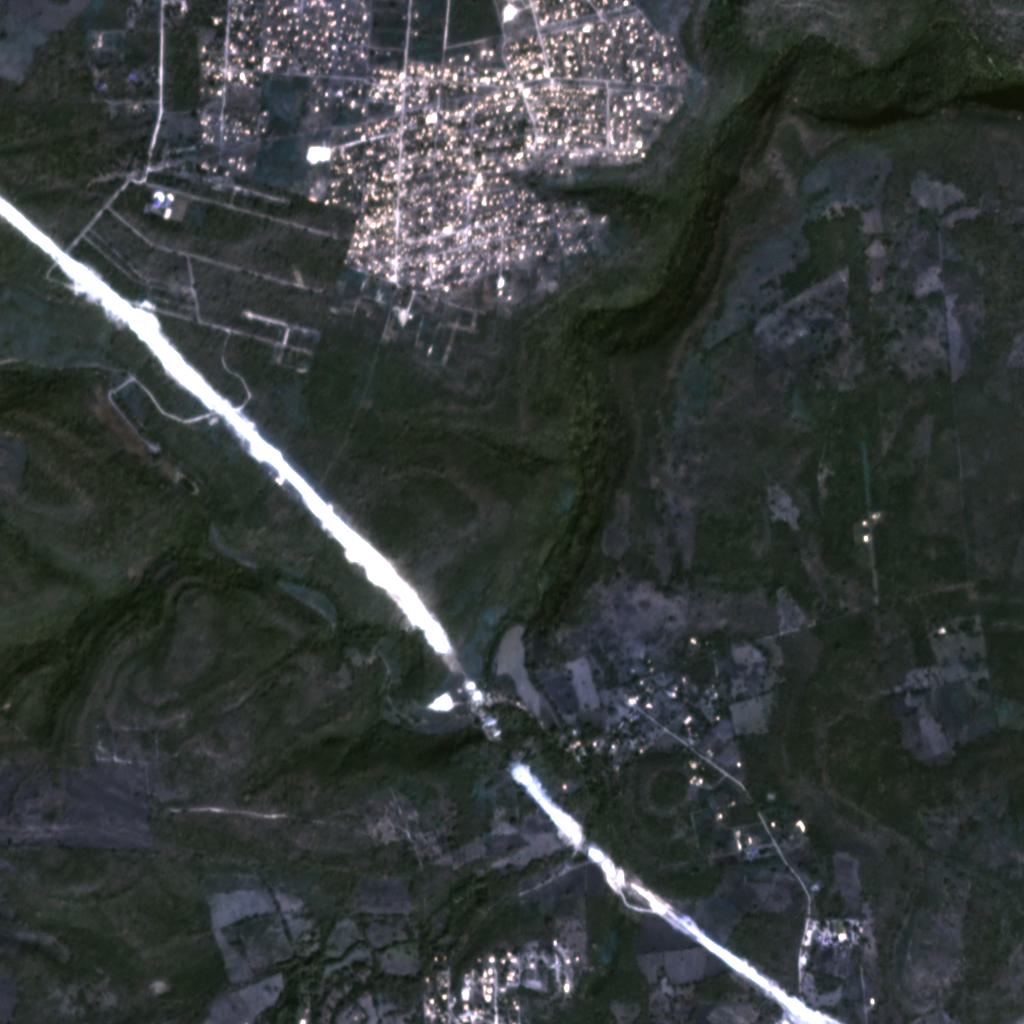} 
        \caption{2018-01-03}
        \label{fig:dyn-sub3}
    \end{subfigure}
    \hfill
    \begin{subfigure}[b]{0.16\textwidth}
        \centering
        \includegraphics[width=\textwidth]{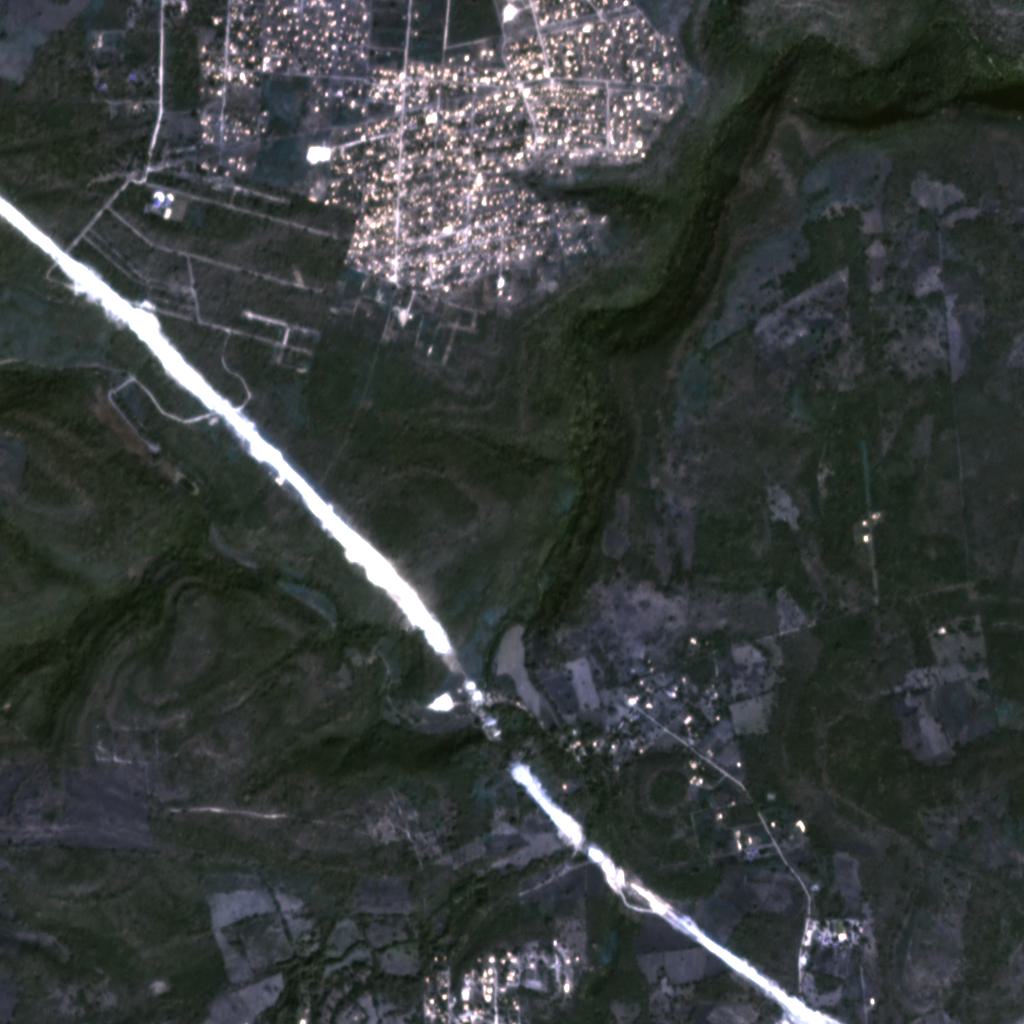} 
        \caption{2018-01-04}
        \label{fig:dyn-sub4}
    \end{subfigure}
    \hfill
    \begin{subfigure}[b]{0.16\textwidth}
        \centering
        \includegraphics[width=\textwidth]{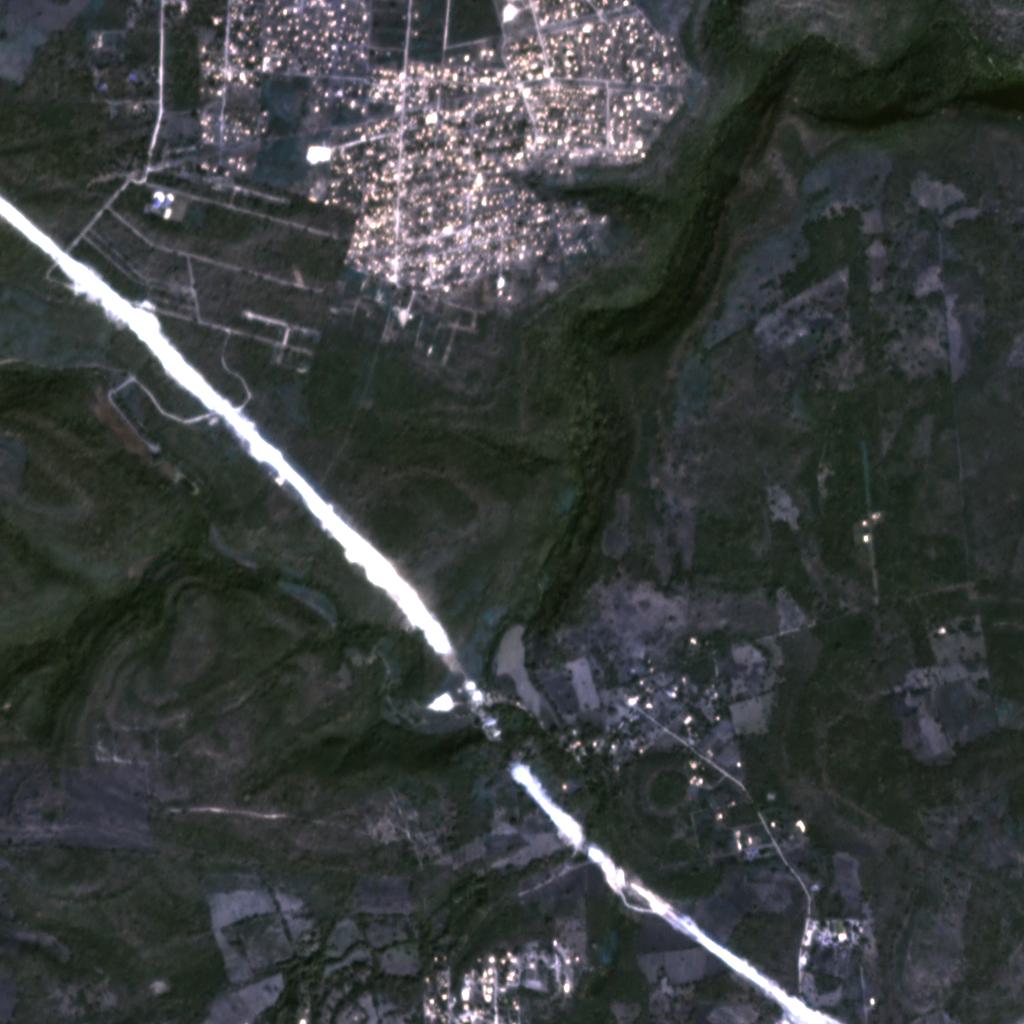} 
        \caption{2018-01-05}
        \label{fig:dyn-sub5}
    \end{subfigure}
    \hfill
    \begin{subfigure}[b]{0.16\textwidth}
        \centering
        \includegraphics[width=\textwidth]{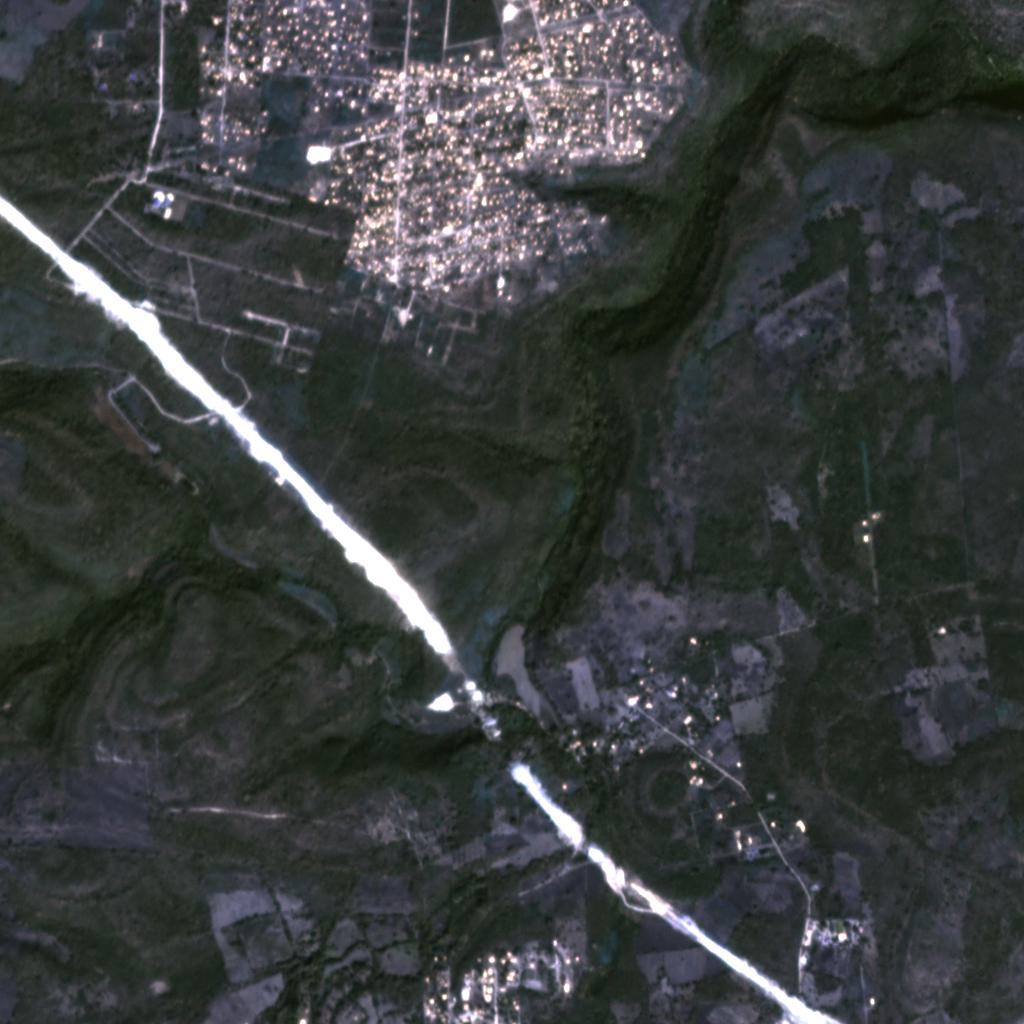} 
        \caption{2018-01-06}
        \label{fig:dyn-sub6}
    \end{subfigure}
    \caption{Example time series data of \textbf{DynamicEarthNet}. (Location: Global. Domain: Urban Land Cover.) The images sequences are visualized with B4, B3, B2 bands and sampled from the first 6 dates of every month to reduce computation requirement. This dataset challenges the models by extremely subtle daily changes.}
    \label{fig:dyn-examples}
\end{figure*}

%% file: tabs/extensive_dynamicen.tex
\begin{table*}[htbp]
    \centering
        \caption{DynamicEarthNet results using 100\% of the dataset, comparing linear and L-TAE temporal mapping strategies. Best results are in \textbf{bold}, second best are \underline{underlined}.}
    \label{tab:extensive-dyn-100perc}
    \resizebox{\textwidth}{!}{%
    \begin{tabular}{lccccccccccc}
    \toprule
        &  \multicolumn{2}{c}{mIoU $\uparrow$}&  \multicolumn{2}{c}{m-F1 $\uparrow$}&  \multicolumn{2}{c}{m-prec $\uparrow$}&  \multicolumn{2}{c}{ m-rec $\uparrow$}& \multicolumn{2}{c}{ m-acc $\uparrow$}\\
        \cmidrule(lr){2-3} \cmidrule(lr){4-5} \cmidrule(lr){6-7} \cmidrule(lr){8-9} \cmidrule(lr){10-11}
        Model &  Linear &  L-TAE &  Linear &  L-TAE &  Linear &  L-TAE &  Linear &  L-TAE & Linear & L-TAE\\
        \midrule
        CROMA         & 38.25 & 38.29 & 47.14 & 47.31 & 46.70 & 47.08 & 49.84 & 51.42 & 69.54 & 69.72 \\
        DOFA          & \textbf{40.84} & 39.29 & \textbf{51.21} & 49.52 & \textbf{50.38} & 48.06 & 53.49 & \underline{56.75} & \underline{70.03} & 69.82 \\
        GFM-Swin      & 35.78 & 34.09 & 46.59 & 46.41 & 45.33 & 43.44 & 49.77 & 54.88 & 66.51 & 66.19 \\
        Prithvi       & 28.42 & 27.86 & 39.33 & 38.74 & 37.62 & 37.95 & 47.36 & 42.81 & 63.18 & 61.49 \\
        RemoteCLIP    & 34.07 & 31.78 & 45.70 & 43.79 & 43.71 & 42.24 & 50.88 & 52.18 & 66.27 & 64.27 \\
        SatlasNet     & \multicolumn{2}{c}{36.31} & \multicolumn{2}{c}{45.45} & \multicolumn{2}{c}{44.69} & \multicolumn{2}{c}{47.39} & \multicolumn{2}{c}{67.22} \\
        Scale-MAE     & 35.47 & 35.11 & 48.86 & 48.47 & 46.59 & 46.21 & 53.65 & \textbf{60.62} & 66.44 & 66.30 \\
        SpectralGPT   & 39.27 & 37.85 & 49.42 & 47.76 & \underline{49.71} & 47.38 & 50.31 & 51.03 & 69.24 & 69.39 \\
        S12-MoCo      & 36.53 & 35.44 & 44.94 & 44.02 & 44.63 & 44.60 & 48.12 & 45.88 & 68.43 & 67.84 \\
        S12-DINO      & 33.80 & 34.81 & 40.69 & 43.89 & 42.79 & 44.15 & 43.53 & 47.19 & 68.90 & 66.00 \\
        S12-MAE       & 35.23 & 34.08 & 43.18 & 41.06 & 43.32 & 42.32 & 45.89 & 46.90 & 67.98 & 67.42 \\
        S12-Data2Vec  & 36.41 & 35.90 & 44.32 & 44.79 & 44.54 & 45.61 & 48.51 & 45.21 & 69.28 & 66.85 \\
        \midrule
        UNet Baseline &  \multicolumn{2}{c}{\underline{39.46}} & \multicolumn{2}{c}{48.81} &  \multicolumn{2}{c}{47.94} &  \multicolumn{2}{c}{52.58} &   \multicolumn{2}{c}{\textbf{70.47}}  \\
        ViT Baseline  & 36.83 & 36.38 & \underline{49.84} & 49.33 & 48.07 &	47.35 & 53.68 & 54.25 & 66.75 & 67.02 \\
 \bottomrule
    \end{tabular}
    }
\end{table*}

\begin{table*}[htbp]
    \centering
    \caption{Results for DynamicEarthNet dataset with linear time strategy, best results in \textbf{bold}, second best \underline{underlined}.}
    \label{tab:ext-dyn-linear}
    \setlength{\tabcolsep}{4pt}
    \resizebox{\textwidth}{!}{%
    \begin{tabular}{lccccccccccccccc}
    \toprule
         \multicolumn{1}{c}{}&  \multicolumn{3}{c}{mIoU $\uparrow$}&  \multicolumn{3}{c}{m-F1 $\uparrow$}&  \multicolumn{3}{c}{m-prec $\uparrow$}&   \multicolumn{3}{c}{m-rec $\uparrow$}&  \multicolumn{3}{c}{m-acc $\uparrow$}\\
        \cmidrule(lr){2-4} \cmidrule(lr){5-7} \cmidrule(lr){8-10} \cmidrule(lr){11-13} \cmidrule(lr){14-16}
    Model  & 10\% & 50\% & 100\%   & 10\% & 50\% & 100\%   & 10\% & 50\% & 100\%   & 10\% & 50\% & 100\%   & 10\% & 50\% & 100\%\\
 \midrule
CROMA     & 36.72 & 38.94 & 38.25 & 45.54 & 48.57 & 47.14 & 45.20 & 47.93 & 46.70 & 49.89 & 51.29 & 49.84 & 68.89 & \underline{69.42} & 69.54 \\
DOFA       & \underline{39.81} & \textbf{40.16} & \textbf{40.84} & \underline{50.82} & \underline{50.33} & \textbf{51.21} & \underline{49.63} & \underline{49.24} & \textbf{50.38} & \underline{53.20} & \underline{55.92} & 53.49 & \underline{69.15} & \textbf{70.38} & \underline{70.03} \\
GFM-Swin   & 33.73 & 33.05 & 35.78 & 44.25 & 44.28 & 46.59 & 43.72 & 43.05 & 45.33 & 46.80 & 47.09 & 49.77 & 64.96 & 64.55 & 66.51 \\
Prithvi    & 29.63 & 29.88 & 28.42 & 39.06 & 39.81 & 39.33 & 38.18 & 38.43 & 37.62 & 44.73 & 46.60 & 47.36 & 63.78 & 64.45 & 63.18 \\
RemoteCLIP & 35.62 & 37.17 & 34.07 & 46.86 & 48.92 & 45.70 & 45.44 & 48.85 & 43.71 & 52.24 & 53.17 & 50.88 & 66.64 & 67.05 & 66.27 \\
SatlasNet   & 34.64 & 36.34 & 36.31 & 44.20 & 45.11 & 45.45 & 44.29 & 44.68 & 44.69 & 46.29 & 47.30 & 47.39 & 67.22 & 67.65 & 67.22 \\
Scale-MAE  & 37.11 & 35.46 & 35.47 & 48.88 & 48.21 & 48.86 & 47.06 & 46.93 & 46.59 & 52.22 & 53.61 & \underline{53.65} & 67.27 & 66.23 & 66.44 \\
SpectralGPT  & 35.91 & 36.73 & 38.28 & 44.43 & 46.22 & 48.14 & 45.41 & 46.44 & 47.07 & 49.86 & 47.37 & 51.01 & \textbf{69.75} & 68.14 & 68.65 \\
S12-MoCo   & 32.38 & 35.59 & 36.53 & 40.47 & 43.57 & 44.94 & 41.15 & 43.79 & 44.63 & 42.98 & 48.47 & 48.12 & 66.07 & 68.43 & 68.43 \\
S12-DINO   & 34.03 & 34.03 & 33.80 & 41.64 & 41.40 & 40.69 & 43.00 & 42.26 & 42.79 & 41.50 & 48.66 & 43.53 & 67.53 & 67.35 & 68.90 \\
S12-MAE     & 33.62 & 35.40 & 35.23 & 42.15 & 43.23 & 41.06 & 42.92 & 43.57 & 43.32 & 43.26 & 48.17 & 45.89 & 66.31 & 68.27 & 67.98 \\
S12-Data2Vec & 34.71 & 35.36 & 36.41 & 41.77 & 42.85 & 44.79 & 42.89 & 43.64 & 44.54 & 47.52 & 49.81 & 48.51 & 68.35 & 68.95 & 69.28 \\
\midrule
UNet Baseline& 35.59 & 35.14 & \underline{39.46} & 44.45 & 44.87 & 48.81 & 45.69 & 44.59 & \underline{47.94} & 45.47 & 48.09 & 52.58 & \underline{68.38} & 67.82 & \textbf{70.47} \\
ViT Baseline& \textbf{39.89} & \underline{39.26} & 36.83 & \textbf{52.77} & \textbf{51.79} & \underline{49.84} & \textbf{50.40} & \textbf{49.81} & 48.07 & \textbf{56.85} & \textbf{56.67} & 53.68 & 68.55 & 69.19 & 66.75 \\
 \bottomrule
    \end{tabular}}
\end{table*}

\begin{table*}[htbp]
    \centering
    \caption{Results for DynamicEarthNet dataset with L-TAE time strategy, best results in \textbf{bold}, second best \underline{underlined}.}
    \label{tab:ext-dyn-ltae}
    \setlength{\tabcolsep}{4pt}
    \resizebox{\textwidth}{!}{%
    \begin{tabular}{lccccccccccccccc}
    \toprule
         \multicolumn{1}{c}{}&  \multicolumn{3}{c}{mIoU $\uparrow$}&  \multicolumn{3}{c}{m-F1 $\uparrow$}&  \multicolumn{3}{c}{m-prec $\uparrow$}&   \multicolumn{3}{c}{m-rec $\uparrow$}&  \multicolumn{3}{c}{m-acc $\uparrow$}\\
        \cmidrule(lr){2-4} \cmidrule(lr){5-7} \cmidrule(lr){8-10} \cmidrule(lr){11-13} \cmidrule(lr){14-16}
    Model  & 10\% & 50\% & 100\% & 10\% & 50\% & 100\% & 10\% & 50\% & 100\% & 10\% & 50\% & 100\% & 10\% & 50\% & 100\% \\
         \midrule
CROMA     & 36.08 & \underline{38.30} & 38.29 & 45.46 & 47.47 & 47.31 & \underline{48.04} & 46.69 & 47.07 & 51.17 & 52.97 & 51.42 & 66.22 & \underline{68.89} & 69.72 \\
DOFA      & \textbf{39.15} & \textbf{39.20} & \underline{39.29} & \textbf{48.75} & \textbf{50.53} & \textbf{49.52} & \textbf{50.21} & \textbf{49.32} & \textbf{48.06} & 53.52 & 55.11 & \underline{56.75} & \textbf{69.64} & \textbf{70.10} & \underline{69.82} \\
GFM-Swin   & 28.16 & 31.25 & 34.09 & 39.72 & 43.18 & 46.41 & 38.55 & 41.57 & 43.44 & 51.04 & 62.93 & 54.88 & 61.85 & 64.95 & 66.19 \\
Prithvi    & 32.28 & 33.42 & 27.86 & 42.80 & 43.35 & 38.74 & 41.19 & 43.35 & 37.95 & 46.95 & 45.22 & 42.81 & 64.15 & 66.04 & 61.49 \\
RemoteCLIP & 34.43 & 30.91 & 31.78 & 46.01 & 42.06 & 43.79 & 43.29 & 39.90 & 42.24 & \textbf{57.38} & 50.05 & 52.18 & 66.72 & 64.30 & 64.27 \\
SatlasNet  & 34.64 & 36.34 & 36.31 & 44.20 & 45.11 & 45.45 & 44.29 & 44.68 & 44.69 & 46.29 & 47.30 & 47.39 & 67.22 & 67.65 & 67.22 \\
Scale-MAE  & 35.27 & 35.57 & 35.11 & 45.95 & 48.33 & 48.47 & 44.71 & \underline{48.41} & 46.21 & 50.35 & \textbf{61.09} & \textbf{60.62} & 67.02 & 65.12 & 66.30 \\
SpectralGPT & 35.33 & 36.52 & 37.85 & 45.17 & 45.51 & 47.76 & 46.06 & 46.12 & 47.38 & 45.07 & 46.07 & 51.03 & 65.41 & 66.84 & 69.39 \\
S12-MoCo   & 32.24 & 34.45 & 35.44 & 40.03 & 42.62 & 44.02 & 41.40 & 43.35 & 44.60 & 45.94 & 46.75 & 45.88 & 64.31 & 66.47 & 67.84 \\
S12-DINO   & 32.78 & 32.76 & 34.81 & 41.99 & 40.99 & 43.89 & 44.04 & 42.62 & 44.15 & 40.67 & 45.52 & 47.19 & 65.46 & 65.82 & 66.00 \\
S12-MAE    & 30.59 & 33.29 & 34.08 & 38.76 & 40.37 & 41.06 & 40.86 & 41.98 & 42.32 & 40.74 & 49.48 & 46.90 & 67.42 & 67.15 & 67.42 \\
S12-Data2Vec & 33.63 & 32.61 & 35.90 & 41.73 & 40.05 & 44.79 & 43.56 & 41.42 & 45.61 & 42.06 & 46.97 & 45.21 & 67.74 & 65.06 & 66.85 \\
\midrule
UNet Baseline & 35.59 & 35.14 & \textbf{39.46} & 44.45 & 44.87 & 48.81 & 45.69 & 44.59 & \underline{47.94} & 45.47 & 48.09 & 52.58 & \underline{68.38} & 67.82 & \textbf{70.47} \\
ViT Baseline  & 35.39 & 37.33 & 36.38 & \underline{47.87} & \underline{49.81} & \underline{49.33} & 45.25 & 47.46 & 47.35 & \underline{56.31} & \underline{55.29} & 54.25 & 66.36 & 67.58 & 67.02 \\
 \bottomrule
    \end{tabular}
    }
\end{table*}

%% file: figs_tex/ctm_sample.tex
\begin{figure*}[htbp]
    \centering
    \begin{subfigure}[b]{0.19\textwidth}
        \centering
        \includegraphics[width=\textwidth]{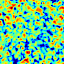} 
        \caption{S1 from 2017-12-21}
        \label{fig:ctm-sub1}
    \end{subfigure}
    \hfill
    \begin{subfigure}[b]{0.19\textwidth}
        \centering
        \includegraphics[width=\textwidth]{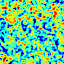}
        \caption{S1 from 2017-12-27}
        \label{fig:ctm-sub2}
    \end{subfigure}
    \hfill
    \begin{subfigure}[b]{0.19\textwidth}
        \centering
        \includegraphics[width=\textwidth]{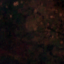}
        \caption{S2 from 2017-12-21}
        \label{fig:ctm-sub3}
    \end{subfigure}
    \hfill
    \begin{subfigure}[b]{0.19\textwidth}
        \centering
        \includegraphics[width=\textwidth]{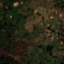}
        \caption{S2 from 2017-12-26}
        \label{fig:ctm-sub4}
    \end{subfigure}
    \hfill
    \begin{subfigure}[b]{0.19\textwidth}
        \centering
        \includegraphics[width=\textwidth]{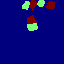}
        \caption{Semantic label}
        \label{fig:ctm-sub5}
    \end{subfigure}
    
    \caption{Example time series data of \textbf{Crop Type Mapping-SS} (Location: South Sudan. Domain: Agriculture). The Sentinel-1 images are visualized by taking VH band and the Sentinel-2 images are visualized by taking B4, B3, B2 bands.The last two frames from 2017 are displayed for this example.}
    \label{fig:ctm-examples}
\end{figure*}

%% file: tabs/extensive_croptypemap.tex
\begin{table*}[htbp]
    \centering
        \caption{Crop Type Mapping-South Sudan results using 100\% of the dataset, comparing linear and L-TAE temporal mapping strategies. Best results are in \textbf{bold}, second best are \underline{underlined}.}
    \label{tab:extensive-ctm-100perc}
    \resizebox{\textwidth}{!}{%
    \begin{tabular}{lccccccccccc}
    \toprule
\multicolumn{2}{c}{}&  \multicolumn{2}{c}{mIoU $\uparrow$}&  \multicolumn{2}{c}{m-F1 $\uparrow$}&  \multicolumn{2}{c}{m-prec $\uparrow$}&  \multicolumn{2}{c}{ m-rec $\uparrow$}& \multicolumn{2}{c}{ m-acc $\uparrow$}\\
\cmidrule(lr){3-4} \cmidrule(lr){5-6} \cmidrule(lr){7-8} \cmidrule(lr){9-10} \cmidrule(lr){11-12}
Model&  Modality&  Linear&  L-TAE&  Linear&  L-TAE&  Linear&  L-TAE&  Linear&  L-TAE& Linear&L-TAE\\
\midrule
CROMA& joint& 38.91	 &43.86	 &48.88	 &55.90	 &51.85	 &58.03	 &46.93 &	58.54 &	73.66 &	80.31\\
CROMA& optical & 47.02 & 49.38 & 57.34 & 61.96 & 56.65 & 70.43 & 50.58 & 62.28 & 72.19 & 75.50 \\
CROMA& sar& 35.39 &	50.44 &	44.68 &	57.95 &	49.90	 &59.81	 &41.78 &	57.93	 &68.78 &	75.80\\
DOFA& optical & 49.81 & 51.33 & 57.53 & 61.37 & 57.24 & 60.81 & 60.38 & 63.82 & 74.65 & 80.61 \\
GFM-Swin& optical & 39.72 & 46.98 & 48.56 & 50.33 & 52.10 & 51.16 & 48.64 & 55.60 & 75.36 & \underline{82.50} \\
Prithvi& optical & 39.92 & 43.07 & 51.04 & 57.92 & 56.81 & 70.00 & 47.31 & 54.59 & 70.98 & 72.48 \\
RemoteCLIP& optical & 46.50 & 52.05 & 54.34 & 61.97 & 51.86 & 63.81 & 59.26 & 63.13 & 79.37 & 80.68 \\
SatlasNet& optical & 46.97 & 46.97 & 53.74 & 53.74 & 53.88 & 53.88 & 53.62 & 53.62 & 71.72 & 71.72 \\
Scale-MAE& optical & 21.39 & 25.42 & 25.84 & 33.92 & 26.18 & 34.00 & 40.79 & 36.23 & 77.10 & 73.43 \\
SpectralGPT& optical & 53.50 & 46.95 & 61.29 & 59.11 & 60.56 & 65.91 & 64.84 & 63.01 & 81.76 & 64.69 \\
S12-Data2Vec& optical & \underline{54.01} & \textbf{54.03} & \underline{66.95} & \textbf{67.55} & \underline{69.01} & \textbf{69.44} & 66.63 & \textbf{71.02} & \underline{81.92} & \textbf{83.57} \\   
S12-DINO& optical& 46.56& 48.66& 59.02& 63.14&61.48& 72.75& 59.73& 62.97& 79.93&76.69\\
S12-MAE& optical& 46.28& 45.8& 57.73& 58.38& 60.10& 63.35& 57.72& 58.77& 74.72&79.03\\
S12-MAE& sar&28.51&	21.06&	37.68&	30.50&	36.01&	52.05&	40.83&	30.79&	76.87&	54.82\\
S12-MoCo& optical& 44.22& 48.58& 51.38& 58.63& 50.76& 60.47& 52.68& 60.1& 70.5&74.99\\
            
\midrule
UNet Baseline& optical &  \multicolumn{2}{c}{ 47.57} & \multicolumn{2}{c}{54.85} &  \multicolumn{2}{c}{55.23} &  \multicolumn{2}{c}{ \underline{68.51}} &   \multicolumn{2}{c}{81.89}  \\
ViT Baseline& optical &50.24	&51.78	&57.38	&60.43	&57.47	&61.95	&57.36	&59.52	&74.58	&74.17\\
 \bottomrule
    \end{tabular}
    }
\end{table*}

\begin{table*}[htbp]
    \centering
    \caption{Results for Crop Type Mapping-South Sudan dataset with linear time strategy, best results in \textbf{bold}, second best \underline{underlined}.}
    \label{tab:ext-ctm-linear}
    \setlength{\tabcolsep}{4pt}
    \resizebox{\textwidth}{!}{%
    \begin{tabular}{lcccccccccccccccc}
    \toprule
         \multicolumn{2}{c}{}&  \multicolumn{3}{c}{mIoU $\uparrow$}&  \multicolumn{3}{c}{m-F1 $\uparrow$}&  \multicolumn{3}{c}{m-prec $\uparrow$}&   \multicolumn{3}{c}{m-rec $\uparrow$}&  \multicolumn{3}{c}{m-acc $\uparrow$}\\
        \cmidrule(lr){3-5} \cmidrule(lr){6-8} \cmidrule(lr){9-11} \cmidrule(lr){12-14} \cmidrule(lr){15-17}
         Model&  Modality&  10\%&  50\%&100\%&  10\%&  50\%&100\%&  10\%&  50\%&100\%&  10\%&  50\%&100\%& 10\%&50\%&100\%\\
 \midrule
 CROMA& optical& 27.67& 43.00& 47.02& 39.63& \textbf{56.98}& 57.34& \textbf{57.85}& \textbf{66.39}& 59.98& 37.79& 53.76& 55.18& 58.02& 76.51& 74.71\\
 DOFA& optical& 28.03& 39.03& 49.81& 40.16& 51.76& 57.53& 48.69& 60.23& 57.24& 37.79& 47.20& 60.38& 58.20& 71.72& 74.65\\
 GFM-Swin& optical& 26.13& 19.73& 42.11& 34.43& 22.05& 45.71& 34.69& 24.45& 49.73& 34.43& 20.08& 42.30& 61.13& \underline{78.90}& \textbf{82.59}\\
 Prithvi& optical& 24.99& 36.14& 39.92& 35.02& 47.00& 51.04& 45.36& 54.49& 56.81& 32.33& 43.44& 47.31& 58.94& 72.29& 70.98\\
 RemoteCLIP& optical& 14.96& 23.53& 46.50& 23.21& 32.36& 54.34& 31.36& 47.13& 51.86& 25.05& 29.99& 59.26& 42.25& 64.73& 79.37\\
 SatlasNet& optical& \underline{38.80}& \underline{46.37}& 46.97& 46.60& 51.60& 53.74& 48.54& 51.27& 53.88& \underline{45.41}& 52.91& 53.62& 65.92& 75.35& 71.72\\
Scale-MAE& optical& 21.89& 21.79& 21.39& 29.83& 27.01& 25.84& 33.71& 27.08& 26.18& 28.02& 28.45& 40.79& \underline{66.12}& 75.14& 77.10\\
SpectralGPT& optical& \textbf{45.36}& \textbf{49.94}& \underline{53.50}& \textbf{52.62}& \underline{56.85}& \underline{61.29}& \underline{57.74}& 55.98& 60.56& \textbf{53.19}& \textbf{58.34}& 64.84& 65.32& 77.45& 81.76\\
S12-Data2Vec& optical& 34.51& 35.05& \textbf{54.01}& 46.66& 48.20& \textbf{66.95}& 57.41& 57.93& \textbf{69.01}& 42.98& 45.29& \underline{66.63}& 62.37& 66.52& \underline{81.92}\\
S12-DINO& optical& 29.30& 34.61& 46.56& 41.53& 46.42& 59.02& 52.03& 57.51& \underline{61.48}& 38.17& 41.66& 59.73& 65.98& 71.11& 79.93\\
S12-MAE& optical& 32.58& 43.21& 46.28& 45.18& 56.17& 57.73& 55.80& \underline{60.79}& 60.10& 41.31& \underline{55.55}& 57.72& 61.20& 78.82& 74.72\\
S12-MoCo& optical& 36.71& 37.41& 44.22& \underline{47.02}& 44.61& 51.38& 53.10& 49.13& 50.76& 43.44& 41.59& 52.68& \textbf{69.73}& 73.77& 70.50\\

\midrule
UNet Baseline & optical& 13.88& 36.30& 47.57& 21.10& 43.44& 54.85& 32.42& 47.14& 55.23& 23.06& 42.10& \textbf{68.51}& 42.51& \textbf{79.24}& 81.89\\
ViT Baseline& optical& 30.10& 34.41& 50.24& 42.11& 46.39& 57.38& 54.21& 56.45& 57.47& 39.28& 42.61& 57.36& 59.31& 71.98& 74.58\\
 \bottomrule
    \end{tabular}
    }
\end{table*}

\begin{table*}[htbp]
    \centering
    \caption{Results for Crop Type Mapping-South Sudan dataset with L-TAE time strategy, best results in \textbf{bold}, second best \underline{underlined}.}
    \label{tab:ext-ctm-ltae}
    \setlength{\tabcolsep}{4pt}
    \resizebox{\textwidth}{!}{%
    \begin{tabular}{lcccccccccccccccc}
    \toprule
         \multicolumn{2}{c}{}&  \multicolumn{3}{c}{mIoU $\uparrow$}&  \multicolumn{3}{c}{m-F1 $\uparrow$}&  \multicolumn{3}{c}{m-prec $\uparrow$}&   \multicolumn{3}{c}{m-rec $\uparrow$}&  \multicolumn{3}{c}{m-acc $\uparrow$}\\
        \cmidrule(lr){3-5} \cmidrule(lr){6-8} \cmidrule(lr){9-11} \cmidrule(lr){12-14} \cmidrule(lr){15-17}
          Model&  Modality&  10\%&  50\%&100\%&  10\%&  50\%&100\%&  10\%&  50\%&100\%&  10\%&  50\%&100\%& 10\%&50\%&100\%\\
         \midrule
 CROMA& optical& 36.77& 42.20& 49.38& \textbf{
 49.76}& \textbf{56.72}& 61.96& \underline{60.99} & \textbf{67.83}& \underline{71.19}& 44.81& \textbf{53.13}& 59.97& 70.30& 70.07& 68.30\\
 DOFA& optical& 29.91& 30.93& 51.33& 43.31& 42.65& 61.37& 59.89& 57.50& 60.81& 39.47& 39.04& \underline{63.82}& 61.53& 69.02& 80.61\\
 GFM-Swin& optical& 13.30& 38.58& 46.98& 18.97& 44.74& 50.33& 25.03& 49.63& 51.16& 41.19& 42.98& 55.60& 44.47& \textbf{81.12}& \underline{82.50} \\
 Prithvi& optical& 27.71& \underline{42.51}& 43.07& 37.89& 54.18& 57.92& 51.44& \underline{60.21} & 70.00& 38.06& 50.98& 54.59& 70.44& 77.62& 72.48\\
 RemoteCLIP& optical& 19.86& 32.09& \underline{52.05} & 30.90& 44.35& 61.97& 43.76& 48.60& 63.81& 32.18& 45.70& 63.13& 46.74& 70.81& 80.68\\
 SatlasNet& optical& \textbf{38.80}& \textbf{46.37}& 46.97& 46.60& 51.60& 53.74& 48.54& 51.27& 53.88& 45.41& 52.91& 53.62& 65.92& 75.35& 71.72\\
 Scale-MAE& optical& 13.44& 20.32& 25.42& 21.91& 27.17& 33.92& 32.64& 25.52& 34.00& 28.74& 30.13& 36.23& 35.30& 63.69& 73.43\\
 SpectralGPT& optical& 27.37& 39.92& 46.95& 39.67& 43.98& 59.11& 55.66& 59.21& 65.91& 36.52& \underline{52.98}& 63.01& 61.72& 63.72& 64.69\\
 S12-Data2Vec& optical& 34.11& 28.53& \textbf{54.03}& 45.70& 38.19& \textbf{67.55}& 58.73& 36.81& 69.44& 44.19& 41.62& 71.02& \textbf{75.49}& 73.78& \textbf{83.57}\\
 S12-DINO& optical& \underline{38.44}& 31.13& 48.66& 48.74& 42.34& \underline{63.14} & 56.70& 48.11& \textbf{72.75}& \underline{50.19}& 40.39& 62.97& 72.02& 70.26& 76.69\\
S12-MAE& optical& 35.29& 28.07& 45.80& 48.86& 38.32& 58.38& \textbf{62.76}& 37.82& 63.35& 48.23& 43.09& 58.77& \underline{73.54}& 71.27& 79.03\\
S12-MoCo& optical& 36.54& 41.32& 48.58& \underline{49.33} & \underline{53.95} & 58.63& 58.60& 56.37& 60.47& \textbf{51.51}& 53.78& 60.10& 73.16& 78.43& 74.99\\
\midrule
UNet Baseline& optical& 13.88& 36.30& 47.57& 21.10& 43.44& 54.85& 32.42& 47.14& 55.23& 23.06& 42.10& \textbf{68.51}& 42.51& \underline{79.24} & 81.89\\
ViT Baseline& optical& 17.16& 40.79& 51.78& 23.79& 48.26& 60.43& 25.58& 50.09& 61.95& 26.02& 46.84& 59.52& 55.91& 66.62& 74.17\\
 \bottomrule
    \end{tabular}
    }
\end{table*}

%% file: figs_tex/sn7_ai4farms_sample.tex
\begin{figure*}[htbp]
    \centering
    \begin{minipage}{0.60\textwidth}
        \centering
    \begin{subfigure}[b]{0.32\textwidth}
        \centering
       \includegraphics[width=\textwidth]{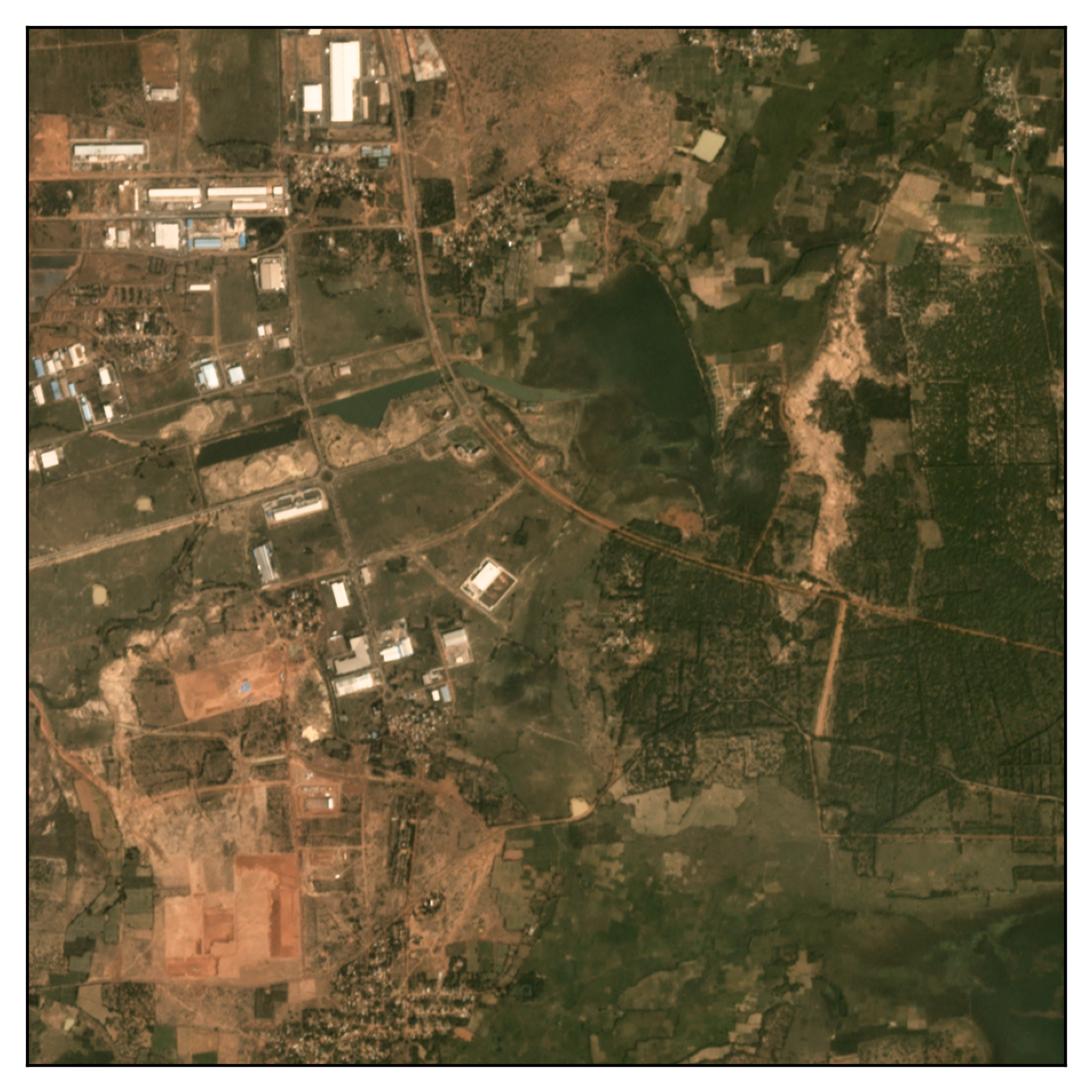} 
        \caption{Pre-change image}
        \label{subfig:prechange_image}
    \end{subfigure}
    \hfill
    \begin{subfigure}[b]{0.32\textwidth}
        \centering
        \includegraphics[width=\textwidth]{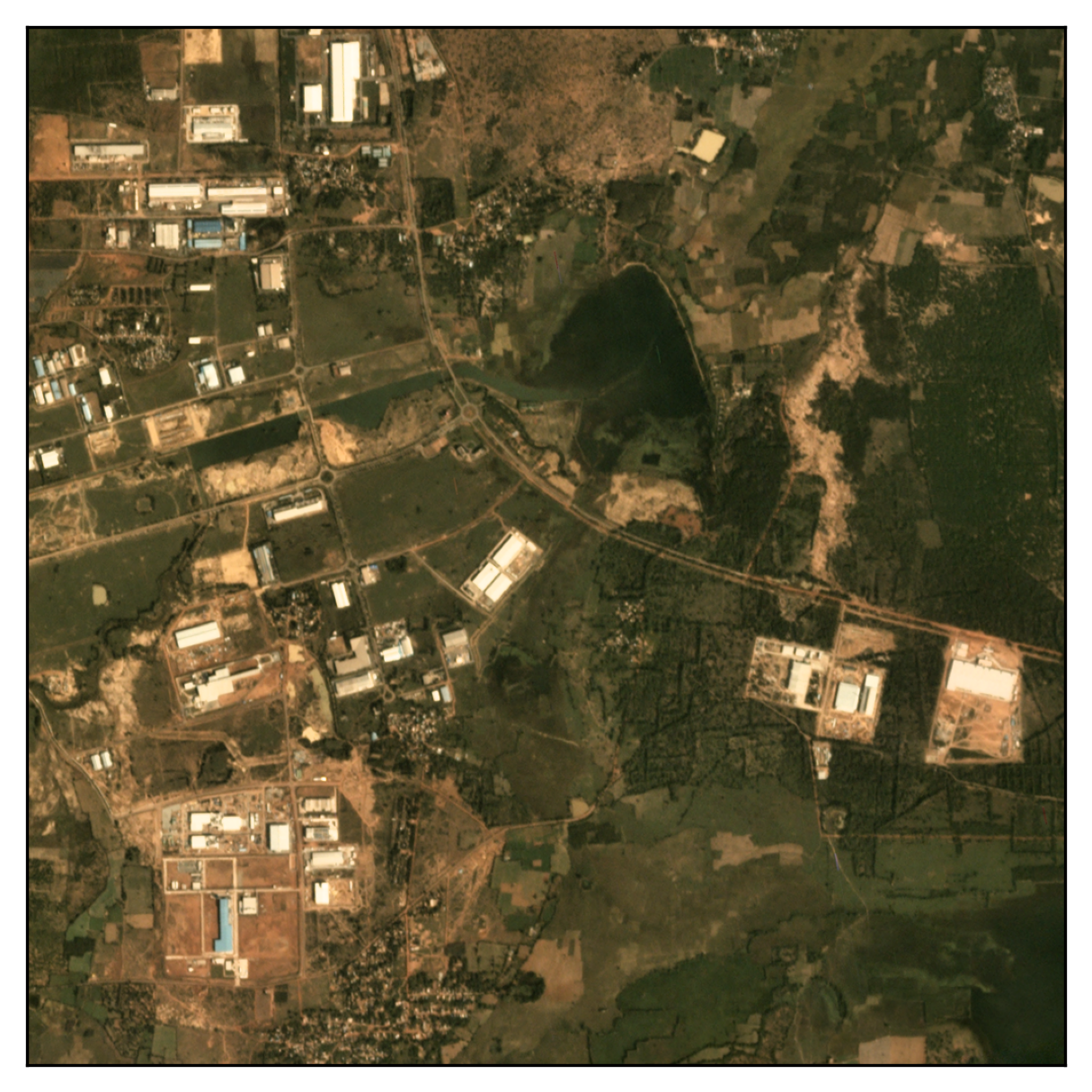} 
        \caption{Post-change image}
        \label{subfig:postcange_image}
    \end{subfigure}
    \hfill
    \begin{subfigure}[b]{0.32\textwidth}
        \centering
       \includegraphics[width=\textwidth]{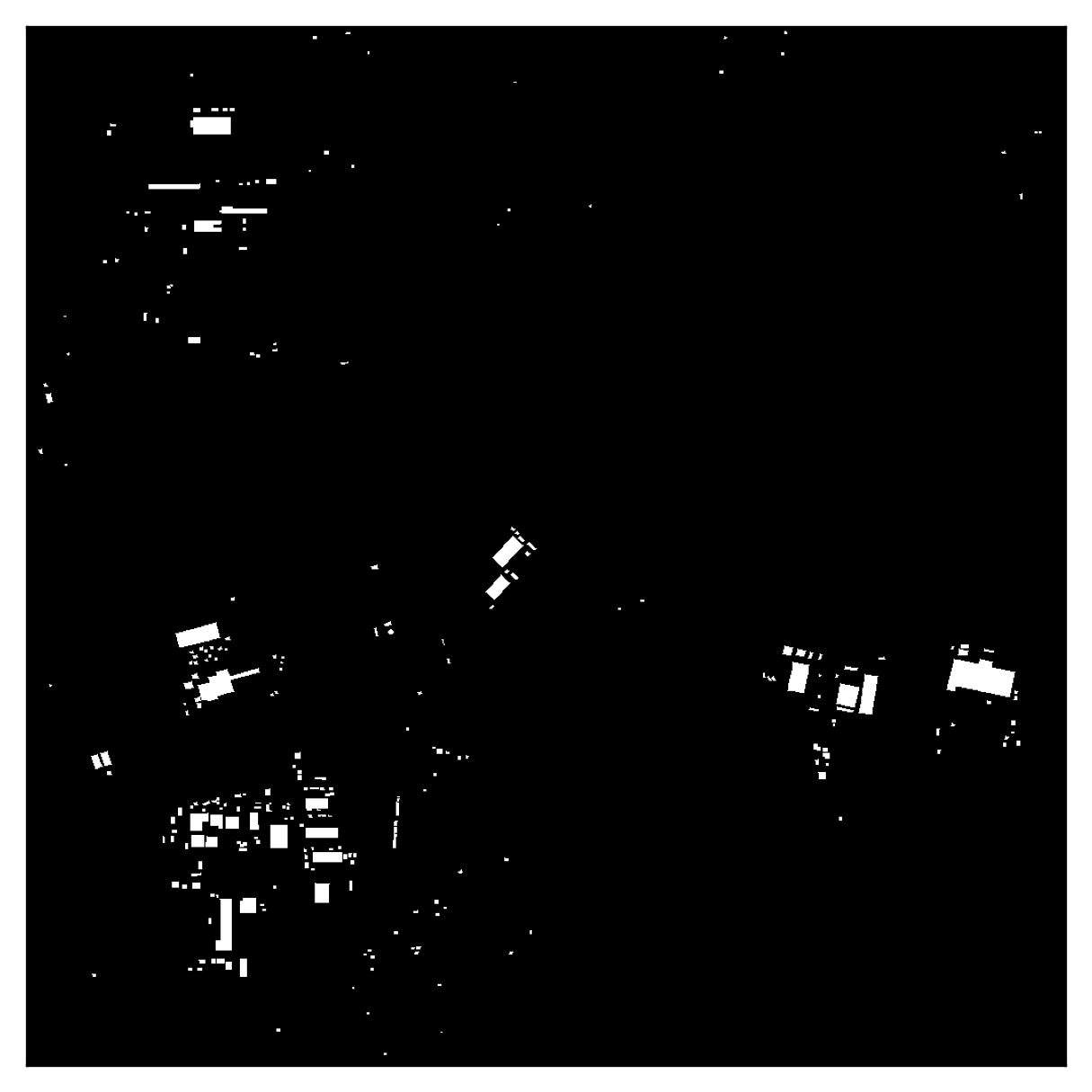} 
        \caption{Building change label}
        \label{subfig:change_label}
    \end{subfigure}
    
    \caption{Example data of \textbf{SpaceNet} (Location: Global. Domain: Urban.). The pre-change and post-change images represent the first and last cloud-free satellite images in the time series. The label visualizes unchanged (black) and changed (white) pixels regarding built-up areas between the two images.}
    \label{fig:sn7-examples}
    \end{minipage}
    \hfill
    \begin{minipage}{0.38\textwidth}
        \centering
        \begin{subfigure}[b]{0.48\textwidth}
        \centering 
        \includegraphics[width=\textwidth]{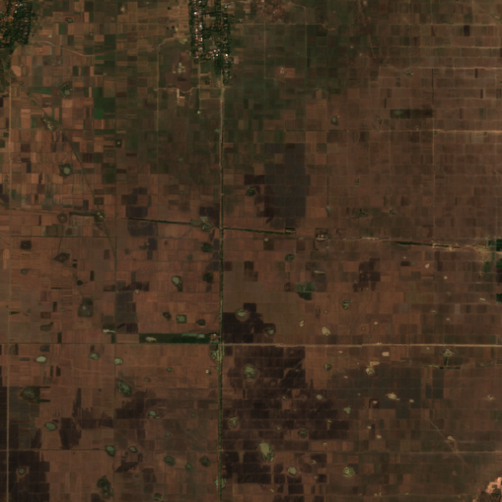} 
        \caption{Sentinel2 image} 
        \label{fig:aisubfig1}
    \end{subfigure}
    \hfill
    \begin{subfigure}[b]{0.48\textwidth}
        \centering
        \includegraphics[width=\textwidth]{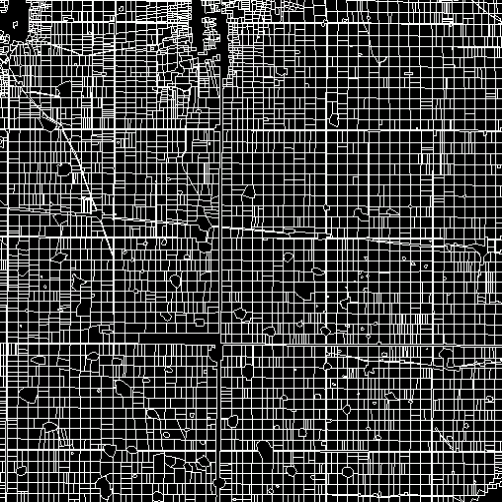} 
        \caption{Small farms label}
        \label{fig:aisubfig2}
    \end{subfigure}
    \caption{Example data of \textbf{AI4SmallFarms} (Location: Southeast Asia. Domain: Agriculture.). The example image is taken from Cambodia and visualized with B4, B3 and B2 bands.}
    \label{fig:ai4sf_example}
    \end{minipage}

    \label{fig:example_ai4smallfarms}
\end{figure*}

%% file: tabs/extensive_sn7_change_strategy.tex
\begin{table*}[htbp]
    \centering
    \caption{Change detection results for the SpaceNet 7 dataset with 100 \% of the data. The results in the last two columns are for the foreground (change) class. The best results are in \textbf{bold} and the second best \underline{underlined}.}
    \begin{tabularx}{\textwidth}{lYYYYYYYYYY}
    \toprule
    & \multicolumn{2}{c}{mIoU $\uparrow$}&  \multicolumn{2}{c}{m-F1 $\uparrow$}&  \multicolumn{2}{c}{m-acc $\uparrow$}&  \multicolumn{2}{c}{ IoU $\uparrow$}& \multicolumn{2}{c}{ F1 $\uparrow$} \\
         \cmidrule(lr){2-3} \cmidrule(lr){4-5} \cmidrule(lr){6-7} \cmidrule(lr){8-9} \cmidrule(lr){10-11}
    Model&  Conc.&  Diff.&  Conc.&  Diff.&  Conc.&  Diff.&  Conc.&  Diff.& Conc.&Diff.\\
    \midrule
    CROMA& 59.28 & 59.85 & 66.04 & 66.77 & 99.14 & 99.28 & 19.41 & 20.41 & 32.52 & 33.91 \\
    DOFA& 61.84 & \underline{61.47} & 69.37 & \underline{68.93} & \textbf{99.45} & \underline{99.35} & 24.24 & \underline{23.60} & 39.02 & \underline{38.19} \\
    GFM-Swin& 60.89 & 60.53 & 68.17 & 67.69 & 99.32 & 99.32 & 22.46 & 21.75 & 36.68 & 35.72 \\
    Prithvi& 56.54 & 58.67 & 62.09 & 65.12 & 99.00 & 99.26 & 14.09 & 18.08 & 24.69 & 30.62 \\
    RemoteCLIP& 57.76 & 58.40 & 63.76 & 64.74 & 99.33 & 99.25 & 16.18 & 17.55 & 27.86 & 29.87 \\
    SatlasNet& 61.88 & 60.63 & 69.44 & 67.83 & 99.38 & 99.31 & 24.38 & 21.96 & 39.20 & 36.01 \\
    Scale-MAE& \textbf{62.96} & \textbf{62.62} & \textbf{70.79} & \textbf{70.39} & \textbf{99.45} & \textbf{99.39} & \textbf{26.47} & \textbf{25.85} & \textbf{41.86} & \textbf{41.09} \\
    SpectralGPT& 58.86 &   59.43   & 65.41 &   66.20   & 99.25 &   99.26   & 18.48 &   19.60   & 31.19 & 32.78\\
    S12-MoCo& 57.64 & 58.28 & 63.71 & 64.63 & 99.08 & 99.12 & 16.20 & 17.44 & 27.89 & 29.70 \\
    S12-DINO& 56.47 & 57.00 & 61.95 & 62.77 & 99.05 & 99.04 & 13.89 & 14.96 & 24.39 & 26.03 \\
    S12-MAE& 57.13 & 57.50 & 62.96 & 63.50 & 99.06 & 99.09 & 15.20 & 15.91 & 26.39 & 27.45 \\
    S12-Data2Vec& 58.23 & 57.89 & 64.50 & 64.04 & 99.23 & 99.15 & 17.22 & 16.62 & 29.39 & 28.51 \\
    \midrule
    UNet Baseline & \underline{62.09} & 51.01 & \underline{69.70} & 54.36 & \underline{99.41} & 96.48 & \underline{24.76} & 5.55 & \underline{39.70} & 10.51 \\
    ViT Baseline & 52.57 & 57.09 & 55.78 & 62.92 & 98.62 & 99.00 & 6.53 & 15.17 & 12.25 & 26.35 \\
    \bottomrule
    \end{tabularx}
    \label{tab:sn7_change_strategy}
\end{table*}

%% file: tabs/extensive_sn7_limited_label.tex
\begin{table*}[htbp]
    \centering
    \caption{Change detection results for the SpaceNet 7 dataset with the concatenation strategy and limited training label scenarios. The best results are in \textbf{bold} and the second best \underline{underlined}.}
    \begin{tabularx}{\textwidth}{lYYYYYYYYY}
    \toprule
         &  \multicolumn{3}{c}{mIoU $\uparrow$}&  \multicolumn{3}{c}{m-F1 $\uparrow$}&  \multicolumn{3}{c}{m-acc $\uparrow$} \\
         \cmidrule(lr){2-4} \cmidrule(lr){5-7} \cmidrule(lr){8-10}
         Model& 10 \% &  50 \% &  100 \% & 10 \% &  50 \% &  100 \% & 10 \% &  50 \% &  100 \% \\
         \midrule
CROMA & 42.15 & 59.31 & 59.28 & 46.74 & 66.07 & 66.04 & 82.90 & 99.20 & 99.14 \\
DOFA & 46.10 & 47.06 & 61.84 & 48.09 & 49.95 & 69.37 & 92.03 & 92.07 & \textbf{99.45} \\
GFM-Swin & 39.48 & 59.83 & 60.89 & 44.58 & 66.77 & 68.17 & 78.31 & 99.24 & 99.32 \\
Prithvi & 36.78 & 49.45 & 56.54 & 42.88 & 52.24 & 62.09 & 72.87 & 95.36 & 99.00 \\
RemoteCLIP & 43.11 & 50.83 & 57.76 & 46.97 & 53.66 & 63.76 & 85.29 & 97.04 & 99.33 \\
 SatlasNet & \textbf{49.78} & \underline{60.74} & 61.88 & \underline{49.89} & \underline{67.97} & 69.44 & \textbf{99.55} & \underline{99.31} & 99.38 \\
 Scale-MAE & \underline{49.68} & \textbf{61.24} & \textbf{62.96} & \textbf{52.10} & \textbf{68.61} & \textbf{70.79} & \underline{96.19} & \textbf{99.36} & \textbf{99.45} \\
 SpectralGPT & 36.31 & 56.28 & 58.86 & 42.74 & 61.82 & 65.41 & 71.68 & 98.76 & 99.25 \\
 S12-MoCo & 49.46 & 56.21 & 57.64 & 50.16 & 61.63 & 63.71 & 98.33 & 98.92 & 99.08 \\
 S12-DINO & 41.15 & 55.14 & 56.47 & 45.71 & 60.00 & 61.95 & 81.51 & 98.81 & 99.05 \\
 S12-MAE & 40.51 & 55.55 & 57.13 & 45.30 & 60.63 & 62.96 & 80.27 & 98.85 & 99.06 \\
 S12-Data2Vec & 40.66 & 56.94 & 58.23 & 45.44 & 62.67 & 64.50 & 80.49 & 99.07 & 99.23 \\
 \midrule
 UNet Baseline & 46.08 & 46.82 & \underline{62.09} & 49.49 & 50.06 & \underline{69.70} & 90.02 & 91.24 & \underline{99.41} \\
 ViT Baseline & 36.01 & 49.21 & 52.57 & 42.44 & 51.45 & 55.78 & 71.22 & 95.85 & 98.62 \\
 \bottomrule
    \end{tabularx}
    \label{tab:sn7_limited_label}
\end{table*}

%% file: tabs/extensive_ai4smallfarms.tex
\begin{table*}[htbp]
    \centering
    \caption{Results for the AI4SmallFarms dataset. Best results in \textbf{bold}, second best \underline{underlined}.}
    \label{tab:ai4smallfarms}
    \begin{tabularx}{\textwidth}{lYYYYYYYYY}
    \toprule
        \multicolumn{1}{c}{}&  \multicolumn{2}{c}{IoU $\uparrow$}&  \multicolumn{2}{c}{F1 $\uparrow$}&  \multicolumn{2}{c}{prec $\uparrow$}&  \multicolumn{2}{c}{rec $\uparrow$}& \multicolumn{1}{c}{acc $\uparrow$}\\
        \cmidrule(lr){2-3} \cmidrule(lr){4-5} \cmidrule(lr){6-7} \cmidrule(lr){8-9}
        Model & mIoU& fg-IoU& m-F1& fg-F1& m-prec& fg-prec& m-rec& fg-rec& m-acc\\
        \midrule
        CROMA& 25.65& \underline{26.54}& 40.82& \underline{41.95}& \underline{56.33}& \underline{86.83}& 56.69& 27.65& 40.84\\
        DOFA& 27.07& 26.18& 42.60& 41.50& 56.11& 82.69& 55.82& \underline{27.70}& 42.62\\
        GFM-Swin& 27.19& 25.44& 42.73& 40.56& 55.09& 79.27& 54.65& 27.25& 42.81\\
        Prithvi& 26.86& 25.79& 42.34& 41.01& 55.50& 81.38& 55.19& 27.41& 42.37\\
        RemoteCLIP& 25.12& 26.24& 40.14& 41.57& 55.76& 86.46& 56.18& 27.36& 40.17\\
        SatlasNet& 25.13& 25.29& 40.17& 40.37& 54.35& 82.27& 54.39& 26.74& 40.17\\
        Scale-MAE& 21.47& 26.43& 35.07& 41.81& 55.30& \textbf{93.76}& \underline{58.06}& 26.91& 35.77\\
        SpectralGPT& 26.75& 25.22& 42.19& 40.27& 54.66& 79.11& 54.30& 27.01& 42.25\\
        S12-Data2Vec& 24.23& 25.25& 39.00& 40.32& 54.06& 83.67& 54.33& 26.56& 39.03\\
        S12-DINO& 25.62& 25.04& 40.79& 40.05& 54.12& 80.35& 53.98& 26.67& 40.80\\
        S12-MAE& 24.69& 24.94& 39.60& 39.93& 53.73& 81.54& 53.78& 26.44& 39.60\\
        S12-MoCo& 25.38& 25.47& 40.48& 40.59& 54.67& 82.62& 54.70& 26.91& 40.48\\
        \midrule
        UNet Baseline & \textbf{46.34}& \textbf{34.07}& \textbf{62.36}& \textbf{50.83}& \textbf{67.82}& 71.61& \textbf{63.38}& \textbf{39.40}& \textbf{65.90}\\
        ViT Baseline& \underline{38.37}& 12.71& \underline{50.31}& 22.56& 50.46& 20.23& 50.54& 25.49& \underline{65.81}\\
    \bottomrule
    \end{tabularx}
\end{table*}

%% file: figs_tex/biomassters_sample.tex
\begin{figure*}[htbp]
    \centering
    \hspace*{\fill} 
        \begin{subfigure}[b]{0.25\textwidth}
        \centering
        \includegraphics[width=\textwidth]{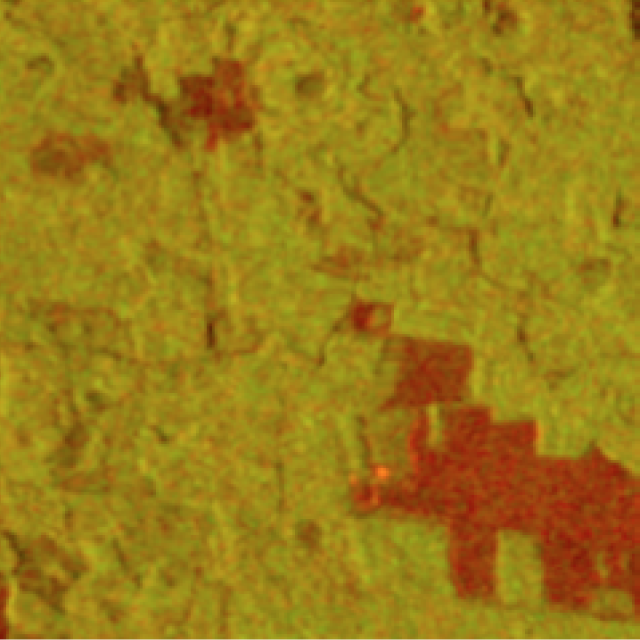} 
        \caption{Sentinel1 image}
        \label{fig:biom-sub4}
    \end{subfigure}
    \hfill
    \begin{subfigure}[b]{0.25\textwidth}
        \centering
        \includegraphics[width=\textwidth]{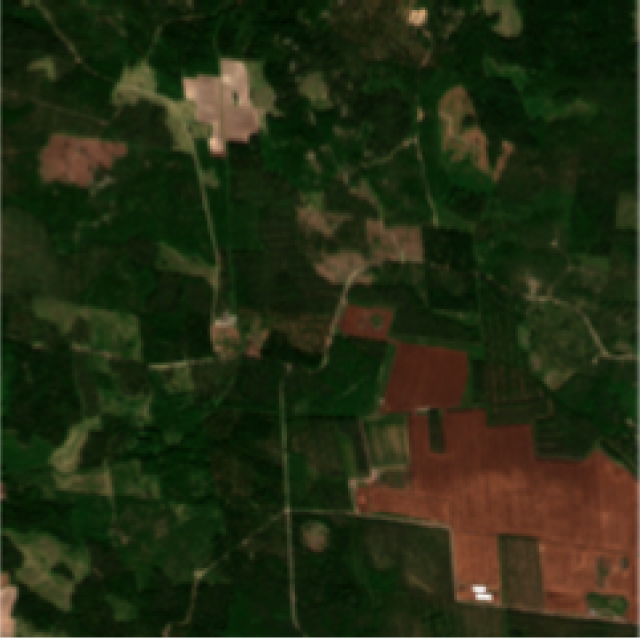} 
        \caption{Sentinel2 image}
        \label{fig:biom-sub5}
    \end{subfigure}
    \hfill
    \begin{subfigure}[b]{0.25\textwidth}
        \centering
        \includegraphics[width=\textwidth]{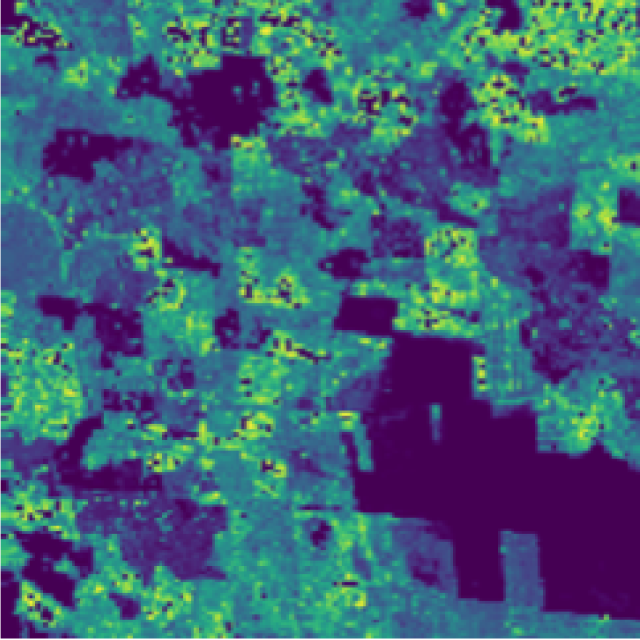} 
        \caption{Aboveground biomass label}
        \label{fig:biom-sub6}
    \end{subfigure}
    \hspace*{\fill} 
    \caption{Example data of \textbf{Biomassters} (Location: Finland. Domain: Forest.). The Sentinel1 images are visualized by taking the VV, VH, VV bands, and the Sentinel-2 images are visualized by taking the B4, B3, and B2 bands and taken from August.}
    \label{AGB_sample}
\end{figure*}

%% file: tabs/extensive_biomass.tex
\begin{table*}[htbp]
\caption{Biomassters results. The evaluation metric is Root Mean Square Error (RMSE $\downarrow$). Best results are highlighted in bold, and second best are underlined. 
SpectralGPT and CROMA are the top-performing models.
Overall multi-temporal with L-TAE is the best configuration.}
\centering
\begin{tabularx}{\textwidth}{lcYYYYYYYYY}
\toprule
             &         & \multicolumn{3}{c}{Uni-temporal} & \multicolumn{6}{c}{Multi-temporal}                                              \\
             &         &           &           &          & \multicolumn{2}{c}{10\%} & \multicolumn{2}{c}{50\%} & \multicolumn{2}{c}{100\%} \\
\cmidrule(lr){3-5} \cmidrule(lr){6-7} \cmidrule(lr){8-9} \cmidrule(lr){10-11}
Model &
  Modality &
  10\% &
  50\% &
  100\% &
  \multicolumn{1}{c}{Linear} &
  \multicolumn{1}{c}{L-TAE} &
  \multicolumn{1}{c}{Linear} &
  \multicolumn{1}{c}{L-TAE} &
  \multicolumn{1}{c}{Linear} &
  \multicolumn{1}{c}{L-TAE} \\ \midrule
CROMA        & joint   & \underline{45.88}     & \underline{43.74}     & \underline{43.07}    & 43.24       & 41.19      & 41.04       & 38.50      & 40.24       & 37.86       \\
CROMA        & optical & 46.99     & 44.88     & 44.16    & 44.11       & \underline{40.25}      & 40.80       & 38.76      & 39.99       & 36.81       \\
CROMA        & sar     & 51.43     & 49.82     & 49.19    & 46.21       & 44.07      & 44.22       & 41.86      & 43.29       & 41.45       \\
DOFA         & joint   & 49.29     & 46.57     & 45.87    & 46.89       & 46.03      & 44.27       & 44.05      & 43.25       & 42.81       \\
GFM-Swin     & optical & 52.98     & 51.34     & 50.46    & 50.24       & 49.30      & 48.55       & 48.19      & 47.96       & 46.83       \\
Prithvi      & optical & 48.93     & 46.76     & 46.05    & 47.58       & 45.48      & 44.04       & 41.03      & 42.90       & 39.99       \\
RemoteCLIP   & optical & 55.14     & 53.58     & 52.81    & 53.22       & 53.32      & 50.65       & 50.09      & 49.60       & 49.79       \\
SatlasNet    & optical & 47.19     & 45.80     & 45.18    & 45.90       & 44.38      & 42.87       & 42.23      & 42.13       & 41.67       \\
Scale-MAE     & optical & 69.53     & 67.27     & 66.55    & 63.19       & 54.16      & 53.94       & 46.74      & 52.24       & 47.15       \\
SpectralGPT  & optical & 46.36     & 44.01     & 43.35    & \underline{42.01}       & \textbf{39.44}      & \underline{39.56}       & \underline{37.34}      & \underline{38.55}       & \underline{36.11}       \\
S12-MoCo     & optical & 47.89     & 46.29     & 45.69    & 44.27       & 44.83      & 42.48       & 41.08      & 41.82       & 40.21       \\
S12-DINO     & optical & 47.65     & 45.98     & 45.39    & 43.91       & 42.74      & 42.13       & 41.47      & 41.54       & 41.23       \\
S12-MAE      & optical & 48.21     & 46.35     & 45.74    & 45.73       & 43.76      & 43.27       & 41.66      & 42.54       & 41.07       \\
S12-MAE      & sar     & 54.41     & 52.44     & 51.69    & 50.43       & 58.62      & 47.88       & 50.88      & 46.87       & 50.09       \\
S12-Data2Vec & optical & 46.38     & 46.20     & 45.48    & 44.86       & 42.02      & 42.76       & 42.82      & 42.02       & 41.91       \\ \midrule
UNet Baseline &
  optical &
  \textbf{44.83} &
  \textbf{41.24} &
  \textbf{40.83} &
  \multicolumn{2}{c}{40.39} &
  \multicolumn{2}{c}{\textbf{36.72}} &
  \multicolumn{2}{c}{\textbf{35.67}} \\
ViT Baseline & optical & 49.13     & 46.22     & 43.77    & 46.19       & 44.89      & 39.96       & 39.56      & 40.94       & 38.55       \\ 
\bottomrule
\end{tabularx}
\label{tab:extensive-biomass}
\end{table*}

%% file: tabs/datasets_per_model.tex
\begin{table*}[htbp]
    \centering
    \caption{Datasets used to evaluate GFMs in their original publications. Every model is evaluated through its protocol.}
    \label{tab:datasets_per_model}
    \setlength\tabcolsep{2pt}
    \scalebox{0.99}{
    \begin{tabularx}{\textwidth}{llYYYYYYYYY}
    \toprule
    Dataset & Task & CROMA & DOFA & GFM & Prithvi & Remote CLIP & Satlas & Scale-MAE  & Spectral GPT & S12-* \\
    \midrule
    BigEarthNet & Cla. & \cmark & & \cmark &  &  & \cmark & & \cmark & \cmark\\
    EuroSAT & Cla. & \cmark & & & & & & \cmark & \cmark & \cmark \\
    So2Sat & Cla. &  & & & & & & & & \cmark\\
    UC Merced & Cla. &  &  & \cmark & & & & \cmark & & \\
    Canadian Cropland & Cla. & \cmark & & & & & & & & \\

    GEO-Bench-class & Cla. &  & \cmark & & & & & & & \\
    
    AiRound & Cla. &  &  &  &  &  &  & \cmark & & \\
    CV-BrCT & Cla. &  &  &  &  &  &  & \cmark & & \\
    MLRSNet & Cla. &  &  &  &  &  &  & \cmark & & \\
    OPTIMAL-31 & Cla. &  &  &  &  &  &  & \cmark & & \\
    WHU-RS19 & Cla. &  &  &  &  &  &  & \cmark & & \\
    RESISC45 & Cla. &  &  &  &  &  & \cmark & \cmark & & \\
    AID & Cla. &  &  &  &  &  & \cmark & & & \\
    fMoW & Cla. &  &  &  &  &  & \cmark & \cmark & & \\
    fMoW-Sentinel & Cla. & \cmark & & & & & & & & \\
    \midrule
    DFC2020 & Seg. & \cmark & & & & & & & & \cmark\\
    iSAID & Seg. &  &  &  &  & \cmark & \cmark & & & \\
    Vaihingen & Seg. &  &  & \cmark &  & \cmark & & & & \\
    Potsdam & Seg. &  &  &  &  & \cmark & & & & \\
    LoveDA & Seg. &  &  &  &  & \cmark & & & & \\
    MARIDA & Seg. & \cmark & & & & & & & & \\
    DW-Expert & Seg. & \cmark & & & & & & & & \\

    GEO-Bench-seg & Seg. &  & \cmark & & & & & & & \\
    
    WHU Aerial & Seg. &  &  & \cmark & & & & & & \\
    Sen1Floods11 & Seg. &  &  &  & \cmark & & & & & \\
    HLS Burn Scars & Seg. &  &  &  & \cmark & & & & & \\

    Mass Roads & Seg. &  &  &  &  &  & \cmark & & & \\
    Mass Buildings & Seg. &  &  &  &  &  & \cmark & & & \\
    HLS Multi-Temp. Crop & Seg. &  &  &  & 
    \cmark & & & & & \\
    SpaceNet-1 & Seg. &  &  &  &  &  &  & \cmark & & \\
    SegMunich & Seg. &  &  &  &  &  &  &  & \cmark & \\
    \midrule
    RSICD & Ret. &  &  &  &  & \cmark & & & & \\
    RSITMD & Ret. &  &  &  &  & \cmark & & & & \\
    UCM & Ret. &  &  &  &  & \cmark & \cmark & & & \\
    \midrule
    DOTA & Det. &  &  &  &  & \cmark & \cmark & & & \\
    DIOR & Det. &  &  &  &  & \cmark & & & & \\
    HRRSD & Det. &  &  &  &  & \cmark & & & & \\
    RSOD & Det. &  &  &  &  & \cmark & & & & \\
    LEVIR & Det. &  &  &  &  & \cmark & & & & \\
    HSRC & Det. &  &  &  &  & \cmark & & & & \\
    Airbus Ships & Det. &  &  &  &  &  & \cmark & & & \\
    \midrule
    OSCD & CD &  &  & \cmark & & & & & \cmark & \\
    DSFIN & CD &  &  & \cmark & & & & & & \\
    \midrule
    SpaceNet-2 & SR &  &  & \cmark & & & & & & \\
    \midrule
    Cloud Gap & Imp. &  &  &  & \cmark & & & & & \\
    \bottomrule
    \end{tabularx}
    }
\end{table*}

%% file: figs_tex/viz1.tex
\begin{figure*}[htbp]
    \centering
    \includegraphics[width=1.0\linewidth]{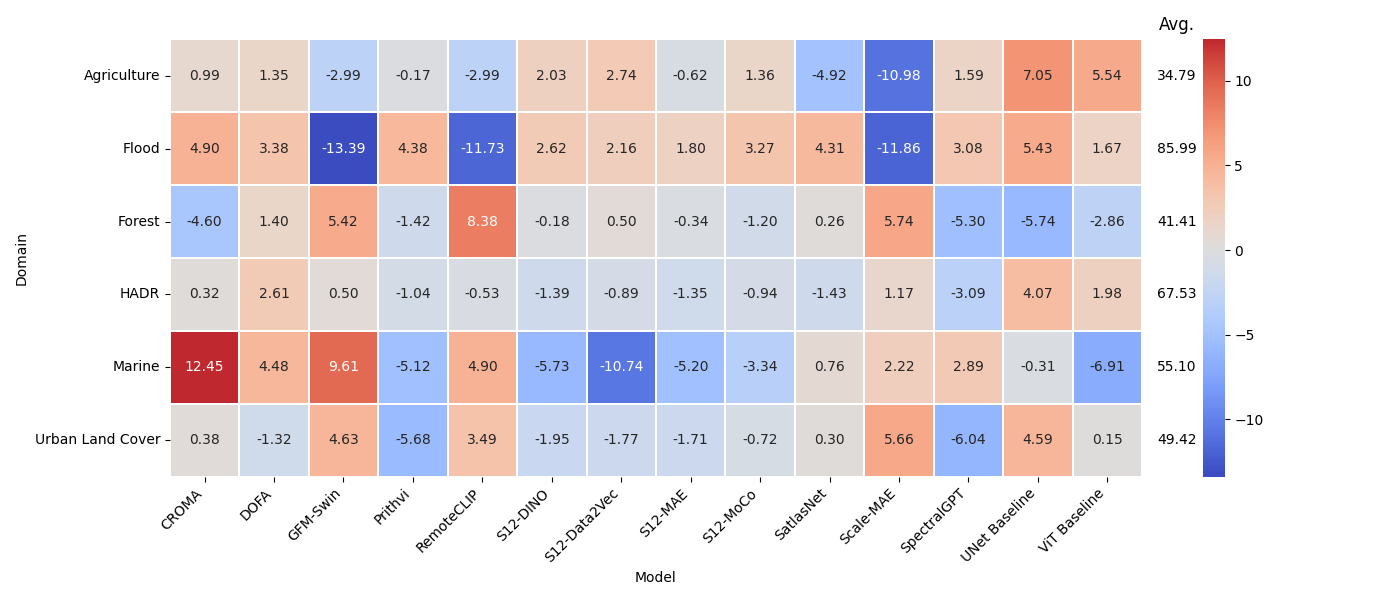}
    \caption{Performance variation across domains for different GFMs. The heatmap illustrates relative performance deltas (higher is better) compared to the domain-specific average. Warmer colors indicate higher-than-average performance, while cooler colors denote underperformance. This visualization highlights the modality-task and domain-specific disparities across models, including performance gaps such as GFM-Swin's relative drop in Flood and Marine tasks, potentially due to distributional shifts in pretraining.}
    \label{fig:app_domain}
\end{figure*}

%% file: figs_tex/viz2.tex
\begin{figure*}[htbp]
    \centering
    \includegraphics[width=1.0\linewidth]{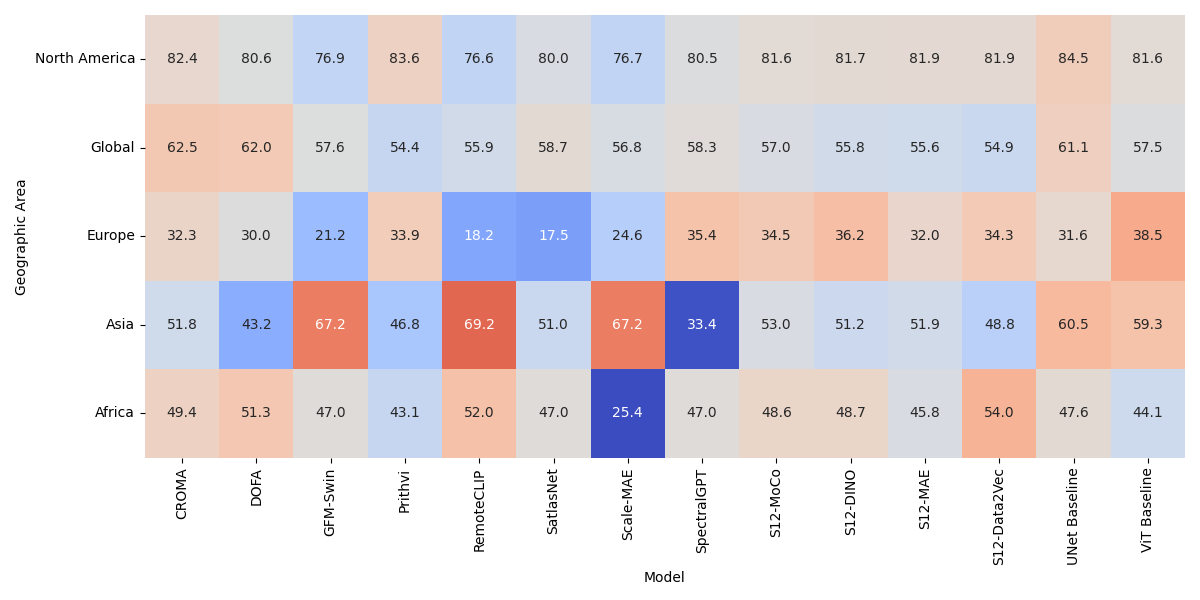}
    \caption{Geographic variation in model performance across foundation models. This heatmap displays average performance scores across five geographic regions. Each cell reflects the model's absolute performance within a specific region, highlighting spatial generalization capabilities. These patterns underscore the importance of geographic context in evaluating GFMs.}
    \label{fig:app_geo}
\end{figure*}

%% file: figs_tex/viz3.tex
\begin{figure*}[htbp]
    \centering
    \includegraphics[width=1.0\linewidth]{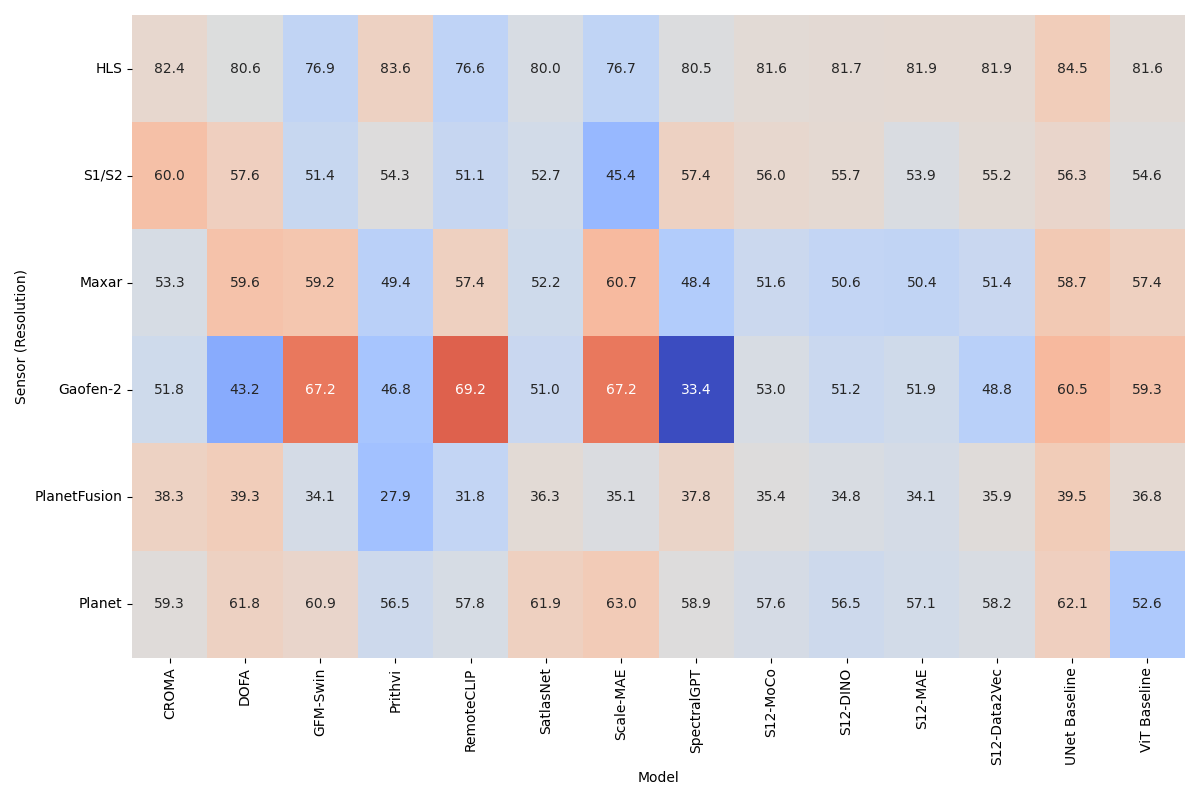}
    \caption{Performance variation of GFMs across different sensors and resolutions. This heatmap illustrates how model performance shifts depending on the source of input imagery, categorized by sensor types and their associated resolutions. The differences highlight the sensitivity of GFMs to input resolution and sensor characteristics.}
    \label{fig:app_res}
\end{figure*}